\documentclass[11pt]{article}
\usepackage[a4paper,margin=0.75in]{geometry}
\usepackage[T1]{fontenc}
\usepackage[utf8]{inputenc}
\usepackage{lmodern}
\usepackage{microtype}
\usepackage{amsmath,amssymb}
\usepackage{booktabs}
\usepackage{adjustbox}
\usepackage{makecell}
\usepackage{placeins}
\usepackage{graphicx}
\usepackage{xcolor}
\usepackage{enumitem}
\usepackage{caption}
\usepackage{tikz}
\usetikzlibrary{arrows.meta,positioning,fit,backgrounds}
\usepackage{pgfplots}
\pgfplotsset{compat=1.18}
\usepgfplotslibrary{groupplots}
\usepackage[numbers,sort&compress]{natbib}
\usepackage[hidelinks]{hyperref}
\usepackage[nameinlink,capitalise]{cleveref}
\usepackage{authblk}

\definecolor{c1}{HTML}{2A78D6}
\definecolor{c2}{HTML}{EB6834}
\definecolor{c3}{HTML}{1BAF7A}
\definecolor{c4}{HTML}{EDA100}

\newcommand{\GhNTestOrgs}{24}
\newcommand{\GhNTrainOrgs}{9}
\newcommand{\GhNValOrgs}{3}
\newcommand{\GhNOrgs}{36}
\newcommand{\GhNDevOrgs}{12}
\newcommand{\GhNEntities}{2{,}496}
\newcommand{\GhNEntityPos}{672}

\newcommand{\GhNPairs}{483{,}912}
\newcommand{\GhNPairPos}{576}

\newcommand{\GhPairPrevalencePct}{0.12}
\newcommand{\GhEntityPrevalencePct}{26.9}
\newcommand{\GhPosPerOrg}{28}
\newcommand{\GhLinkPosHopOne}{48}
\newcommand{\GhLinkPosHopTwo}{192}
\newcommand{\GhLinkPosHopThree}{240}
\newcommand{\GhLinkPosHopFour}{96}
\newcommand{\GhEntityPosHopZero}{96}

\newcommand{\GhLinkMinCell}{32}

\newcommand{\GhExistingEmpMin}{150}
\newcommand{\GhExistingEmpMax}{240}
\newcommand{\GhExistingVendMin}{30}
\newcommand{\GhExistingVendMax}{45}
\newcommand{\GhNewHires}{72}
\newcommand{\GhNewVendors}{32}
\newcommand{\GhNRuns}{6}
\newcommand{\GhRunSpacing}{15}
\newcommand{\GhFraudMult}{2}
\newcommand{\GhLookalikeRatio}{1}
\newcommand{\GhPSuspFraud}{0.7}
\newcommand{\GhPSuspBenign}{0.15}
\newcommand{\GhCamouflage}{0.25}
\newcommand{\GhMissingRate}{0.05}
\newcommand{\GhBgDisclosure}{0.07}
\newcommand{\GhBaseSeed}{9134004}
\newcommand{\GhRecentChangeDays}{3}
\newcommand{\GhTestNodesMin}{1{,}639}
\newcommand{\GhTestNodesMax}{2{,}200}
\newcommand{\GhTestEdgesMin}{3{,}341}
\newcommand{\GhTestEdgesMax}{4{,}640}

\newcommand{\GhNumRelations}{28}
\newcommand{\GhNumEdgeTypes}{14}
\newcommand{\GhFeatureDim}{79}
\newcommand{\GhBootReps}{1{,}000}
\newcommand{\GhBootSeed}{913}
\newcommand{\GhPMin}{0.001}
\newcommand{\GhSeeds}{9131, 9132, 9133}
\newcommand{\GhNSeeds}{3}
\newcommand{\GhFreezeFiles}{60}
\newcommand{\GhFreezeShaShort}{598b8b47161d}
\newcommand{\GhFreezeDate}{2026-10-14}
\newcommand{\GhFreezeVerifiedAtReport}{60}
\newcommand{\GhWallSec}{41{,}521}
\newcommand{\GhWallHours}{11.5}
\newcommand{\GhManifestFiles}{217}
\newcommand{\GhManifestMismatch}{0}
\newcommand{\GhGenSources}{6}
\newcommand{\GhSubgraphMeanKTwo}{5.1}
\newcommand{\GhSubgraphMeanKThree}{49.7}
\newcommand{\GhGraILThreeHours}{9.1}

\newcommand{\GhPyVersion}{3.12.15}
\newcommand{\GhTorchVersion}{2.14.1}
\newcommand{\GhThreads}{6}

\newcommand{\GhNCapHit}{1}
\newcommand{\GhHOneHopTwoMethod}{0.442}
\newcommand{\GhHOneHopTwoRef}{0.650}
\newcommand{\GhHOneHopTwoDiff}{$-$0.208}

\newcommand{\GhHOneHopTwoDiffCI}{[$-$0.262, $-$0.150]}
\newcommand{\GhHOneHopTwoMethodCI}{[0.400, 0.497]}
\newcommand{\GhHOneHopTwoRefCI}{[0.585, 0.716]}

\newcommand{\GhHOneHopThreeMethod}{0.501}
\newcommand{\GhHOneHopThreeRef}{0.081}
\newcommand{\GhHOneHopThreeDiff}{+0.420}

\newcommand{\GhHOneHopThreeDiffCI}{[0.375, 0.469]}
\newcommand{\GhHOneHopThreeMethodCI}{[0.462, 0.549]}
\newcommand{\GhHOneHopThreeRefCI}{[0.066, 0.098]}

\newcommand{\GhHOneHopThreePHolm}{0.005}

\newcommand{\GhHOneHopFourMethod}{0.272}
\newcommand{\GhHOneHopFourRef}{0.000}
\newcommand{\GhHOneHopFourDiff}{+0.272}

\newcommand{\GhHOneHopFourDiffCI}{[0.219, 0.337]}

\newcommand{\GhHOneHopFourPHolm}{0.005}

\newcommand{\GhHTwoMethod}{0.674}
\newcommand{\GhHTwoRef}{0.796}
\newcommand{\GhHTwoDiff}{$-$0.122}

\newcommand{\GhHTwoDiffCI}{[$-$0.151, $-$0.094]}
\newcommand{\GhHTwoMethodCI}{[0.643, 0.705]}
\newcommand{\GhHTwoRefCI}{[0.773, 0.818]}

\newcommand{\GhHThreeMethod}{0.752}
\newcommand{\GhHThreeRef}{0.698}
\newcommand{\GhHThreeDiff}{+0.053}

\newcommand{\GhHThreeDiffCI}{[0.025, 0.086]}
\newcommand{\GhHThreeMethodCI}{[0.719, 0.784]}
\newcommand{\GhHThreeRefCI}{[0.669, 0.728]}

\newcommand{\GhHThreePHolm}{0.006}

\newcommand{\GhNSupported}{3}
\newcommand{\GhNFamily}{5}
\newcommand{\GhNNotSupported}{2}
\newcommand{\GhSOneAMethod}{0.735}
\newcommand{\GhSOneARef}{0.347}
\newcommand{\GhSOneADiff}{+0.388}
\newcommand{\GhSOneADiffCI}{[0.364, 0.412]}

\newcommand{\GhSOneBRef}{0.663}
\newcommand{\GhSOneBDiff}{+0.072}
\newcommand{\GhSOneBDiffCI}{[0.047, 0.099]}

\newcommand{\GhHOneHopTwoDevDiff}{$-$0.034}
\newcommand{\GhHOneHopThreeDevDiff}{+0.405}
\newcommand{\GhHOneHopFourDevDiff}{+0.286}
\newcommand{\GhHTwoDevDiff}{$-$0.061}
\newcommand{\GhHThreeDevDiff}{+0.087}
\newcommand{\GhDevNTrainOrgs}{6}
\newcommand{\GhDevNValOrgs}{2}
\newcommand{\GhDevNOrgs}{4}

\newcommand{\GhDevLinkGraILThreeAUPRC}{0.697}

\newcommand{\GhLinkRulesVTwoRFprTenth}{0.429}

\newcommand{\GhLinkRulesVTwoAUPRCCtlApprover}{0.409}
\newcommand{\GhLinkRulesVTwoAUPRCCtlBankChange}{0.100}
\newcommand{\GhLinkRulesVTwoAUPRCCtlEmpVendor}{0.125}
\newcommand{\GhLinkRulesVTwoAUPRCCtlGhost}{0.474}
\newcommand{\GhLinkRulesVTwoAUPRCCtlSharedAcct}{0.099}

\newcommand{\GhLinkPPRAUROC}{0.993}

\newcommand{\GhLinkPPRAUPRC}{0.093}

\newcommand{\GhLinkPathLRAUPRCHopThree}{0.239}

\newcommand{\GhLinkPathLRAUPRCHopFour}{0.001}

\newcommand{\GhLinkNBFNetAUPRCHopThree}{0.461}

\newcommand{\GhLinkNBFNetAUPRCHopFour}{0.154}

\newcommand{\GhLinkOSFAAUPRCHopThree}{0.394}

\newcommand{\GhLinkOSFAAUPRCHopFour}{0.197}

\newcommand{\GhLinkRDLLinkAUPRC}{0.082}

\newcommand{\GhLinkGraILTwoAUPRCHopThree}{0.392}

\newcommand{\GhLinkGraILTwoAUPRCHopFour}{0.002}

\newcommand{\GhLinkGraILThreeAUPRCSD}{0.071}
\newcommand{\GhLinkGraILThreeRFprTenth}{0.906}

\newcommand{\GhLinkGraILThreeRFprOne}{0.997}

\newcommand{\GhLinkGraILThreeAUPRCCtlEmpVendor}{0.466}
\newcommand{\GhLinkGraILThreeAUPRCCtlGhost}{0.231}

\newcommand{\GhLinkRulesVTwoPlusGraILThreeLRAUPRCHopTwo}{0.730}

\newcommand{\GhLinkRulesVTwoPlusGraILThreeLRAUPRCHopThree}{0.437}

\newcommand{\GhLinkRulesVTwoPlusGraILThreeLRAUPRCHopFour}{0.190}

\newcommand{\GhLinkTwinRulesVTwoAUROCHopFour}{0.780}

\newcommand{\GhLinkTwinRDLLinkAUROC}{0.738}

\newcommand{\GhLinkTwinRDLLinkDiff}{$-$0.058}
\newcommand{\GhLinkTwinRDLLinkDiffCI}{[$-$0.080, $-$0.034]}

\newcommand{\GhLinkTwinGraILThreeAUROCHopFour}{0.803}

\newcommand{\GhLinkTwinRulesVTwoPlusGraILThreeLRAUROC}{0.763}

\newcommand{\GhEntityRulesVTwoRFprOne}{0.104}

\newcommand{\GhEntityRGCNEntAUPRC}{0.776}
\newcommand{\GhEntityRGCNEntAUPRCCI}{[0.744, 0.808]}

\newcommand{\GhEntityRDLEntRFprOne}{0.197}

\newcommand{\GhEntityRDLEntRFprOneSD}{0.060}

\newcommand{\GhEntityRulesVTwoPlusRGCNEntLRAUPRC}{0.801}

\newcommand{\GhEntityRulesVTwoPlusRDLEntLRAUPRC}{0.792}

\newcommand{\GhEntityGraILThreeAUPRC}{0.612}

\newcommand{\GhEntityTwinRulesVTwoAUROC}{0.817}

\newcommand{\GhEntityTwinRDLEntAUROC}{0.821}

\newcommand{\GhEntityTwinRDLEntDiff}{+0.004}
\newcommand{\GhEntityTwinRDLEntDiffCI}{[$-$0.013, 0.023]}

\newcommand{\GhNLinkSaturated}{9}
\newcommand{\GhNLinkMethods}{24}
\newcommand{\GhNEntityMethods}{20}
\newcommand{\GhNLinkTwinBelowRules}{13}
\newcommand{\GhNLinkSingleOthers}{13}
\newcommand{\GhNLinkCombos}{10}
\newcommand{\GhNComboTwinBelowRules}{10}

\newcommand{\GhAudOneNLinkPos}{144}
\newcommand{\GhAudOneNLinkPairs}{103{,}785}

\newcommand{\GhAudOneLinkSOneAllAUPRC}{0.598}

\newcommand{\GhAudOneLinkSTwoMatchAUROC}{0.990}

\newcommand{\GhAudOneLinkSThreeAllAUROC}{0.999}
\newcommand{\GhAudOneLinkSThreeAllAUPRC}{0.685}
\newcommand{\GhAudOneLinkSThreeAllRFprOne}{1.000}

\newcommand{\GhAudOneEntSTwoAllAUROC}{0.888}
\newcommand{\GhAudOneEntSTwoAllAUPRC}{0.838}

\newcommand{\GhAudFiveLinkSTwoMatchAUROC}{0.666}

\newcommand{\GhAudFiveLinkSThreeAllAUROC}{0.999}
\newcommand{\GhAudFiveLinkSThreeAllAUPRC}{0.389}
\newcommand{\GhAudFiveLinkSThreeAllRFprOne}{0.997}

\newcommand{\GhAudFiveEntATTRTwinAUROC}{0.792}

\newcommand{\GhAudOneTellECAUROC}{0.637}
\newcommand{\GhAudOneMutualECPos}{0}
\newcommand{\GhAudOneMutualECNeg}{48}
\newcommand{\GhAudOneMatchedPos}{144}
\newcommand{\GhAudOneMatchedNeg}{156}
\newcommand{\GhAudFiveTwinScreenMaxDev}{0.021}
\newcommand{\GhAudFiveTwinMin}{0.493}
\newcommand{\GhAudFiveTwinMax}{0.514}
\newcommand{\GhPOneNPairPos}{96}
\newcommand{\GhPOneNPairs}{71{,}995}
\newcommand{\GhPOneRulesTunedAUPRC}{0.492}
\newcommand{\GhPOneRulesTunedRFprOneHopThree}{0.205}

\newcommand{\GhPOnePathLRAUPRC}{0.284}
\newcommand{\GhPOnePathLRRFprOneHopThree}{1.000}

\newcommand{\GhPTwoNPairPos}{288}
\newcommand{\GhPTwoNPairs}{215{,}545}
\newcommand{\GhPTwoNTestOrgs}{24}

\newcommand{\GhPTwoRulesTunedAUPRC}{0.502}

\newcommand{\GhPTwoRulesTunedRFprOneHopFour}{0.000}

\newcommand{\GhPTwoGraILThreeAUPRC}{0.973}

\newcommand{\GhPTwoGraILThreeRFprOneHopFour}{1.000}

\newcommand{\GhPTwoNBFNetAUPRC}{0.997}

\newcommand{\GhPTwoNBFNetRFprOneHopFour}{1.000}

\newcommand{\GhPTwoEntRGCNEntAUPRC}{0.989}

\newcommand{\GhLinkPrevPctHopOne}{0.010}

\newcommand{\GhLinkPrevPctHopTwo}{0.040}

\newcommand{\GhLinkPrevPctHopThree}{0.050}

\newcommand{\GhLinkPrevPctHopFour}{0.020}

\newcommand{\GhAudFiveLinkSOneAllAUPRCHopTwo}{0.194}
\newcommand{\GhAudFiveLinkSOneAllAUPRCHopThree}{0.299}
\newcommand{\GhAudFiveLinkSOneAllAUPRCHopFour}{0.052}

\newcommand{\GhAudFiveLinkSThreeAllAUPRCHopTwo}{0.191}
\newcommand{\GhAudFiveLinkSThreeAllAUPRCHopThree}{0.317}
\newcommand{\GhAudFiveLinkSThreeAllAUPRCHopFour}{0.051}
\newcommand{\GhXSanityMaxDev}{$1.5\times10^{-5}$}

\newcommand{\GhXHOneHopFourSeedDiffMin}{+0.130}
\newcommand{\GhXHOneHopFourSeedDiffMax}{+0.378}

\newcommand{\GhXHTwoSeedDiffMin}{$-$0.185}
\newcommand{\GhXHTwoSeedDiffMax}{$-$0.074}

\newcommand{\GhXNSeedHyp}{5}

\newcommand{\GhXReachRulesVTwoHopOne}{0.833}
\newcommand{\GhXReachRulesVTwoHopTwo}{0.854}
\newcommand{\GhXReachRulesVTwoHopThree}{0.180}
\newcommand{\GhXReachRulesVTwoHopFour}{0.000}

\newcommand{\GhXReachPathLRHopThree}{0.336}
\newcommand{\GhXReachPathLRHopFour}{0.001}

\newcommand{\GhXReachGraILThreeHopOne}{0.562}
\newcommand{\GhXReachGraILThreeHopTwo}{0.982}
\newcommand{\GhXReachGraILThreeHopThree}{0.995}
\newcommand{\GhXReachGraILThreeHopFour}{0.989}

\newcommand{\GhXTwRulesVTwoHopThreeWEight}{0.018}

\newcommand{\GhXTwRulesVTwoHopFourWEight}{0.000}

\newcommand{\GhXTwGraILThreeHopThreeWEight}{0.126}

\newcommand{\GhXTwGraILThreeHopFourWEight}{0.050}

\newcommand{\GhXReachGraILThreeHopTwoDiff}{+0.127}
\newcommand{\GhXReachGraILThreeHopTwoDiffCI}{[0.082, 0.177]}

\newcommand{\GhXReachGraILThreeHopThreeDiff}{+0.815}
\newcommand{\GhXReachGraILThreeHopThreeDiffCI}{[0.798, 0.833]}

\newcommand{\GhXReachGraILThreeHopFourDiff}{+0.988}
\newcommand{\GhXReachGraILThreeHopFourDiffCI}{[0.969, 0.999]}

\newcommand{\GhXWMax}{8}
\newcommand{\GhXObsPrevPct}{26.9}

\newcommand{\GhXBaseRulesVTwoPrevOne}{0.089}

\newcommand{\GhXBaseRDLEntPrevOne}{0.119}

\newcommand{\GhXNoSoDDropped}{96}

\newcommand{\GhXNoSoDRDLEntDiff}{+0.052}
\newcommand{\GhXNoSoDRDLEntDiffCI}{[0.021, 0.088]}

\newcommand{\GhXNegTotal}{1{,}824}
\newcommand{\GhXNegTwins}{672}
\newcommand{\GhXNegOtherHard}{552}
\newcommand{\GhXNegOrdinary}{600}
\newcommand{\GhXNegLookalikePct}{67}

\newcommand{\GhXBaseOrdRulesVTwoPrevOne}{0.538}
\newcommand{\GhXBaseOrdRulesVTwoPrevTenth}{0.518}

\newcommand{\GhXBaseOrdRDLEntPrevOne}{0.393}
\newcommand{\GhXBaseOrdRDLEntPrevTenth}{0.269}

\newcommand{\GhXNoSoDTwRDLEntDiff}{+0.073}
\newcommand{\GhXNoSoDTwRDLEntDiffCI}{[0.045, 0.106]}

\newcommand{\GhXConcLinkRulesVTwo}{0.820}
\newcommand{\GhXConcLinkRulesVTwoCI}{[0.792, 0.847]}

\newcommand{\GhXConcLinkGraILThree}{0.728}
\newcommand{\GhXConcLinkGraILThreeCI}{[0.694, 0.762]}
\newcommand{\GhXConcLinkGraILThreeDiff}{$-$0.092}
\newcommand{\GhXConcLinkGraILThreeDiffCI}{[$-$0.119, $-$0.066]}

\newcommand{\GhXConcEntRulesVTwo}{0.830}

\newcommand{\GhXConcEntRDLEnt}{0.831}

\newcommand{\GhXConcNLink}{576}

\newcommand{\GvV}[1]{\ifcsname gv.#1\endcsname\csname gv.#1\endcsname\else\errmessage{undefined addendum value #1}\fi}
\expandafter\def\csname gv.A1.method\endcsname{0.796}
\expandafter\def\csname gv.A1.ref\endcsname{0.667}
\expandafter\def\csname gv.A1.diff\endcsname{+0.129}
\expandafter\def\csname gv.A1.diffci\endcsname{[+0.109, +0.147]}
\expandafter\def\csname gv.A1.p\endcsname{0.001}
\expandafter\def\csname gv.A1.holm\endcsname{0.009}
\expandafter\def\csname gv.A1.supported\endcsname{yes}
\expandafter\def\csname gv.A1.npos\endcsname{576}
\expandafter\def\csname gv.A1.methodci\endcsname{[0.773, 0.818]}
\expandafter\def\csname gv.A1.refci\endcsname{[0.650, 0.682]}
\expandafter\def\csname gv.A2.method\endcsname{0.674}
\expandafter\def\csname gv.A2.ref\endcsname{0.667}
\expandafter\def\csname gv.A2.diff\endcsname{+0.007}
\expandafter\def\csname gv.A2.diffci\endcsname{[$-$0.020, +0.033]}
\expandafter\def\csname gv.A2.p\endcsname{0.317}
\expandafter\def\csname gv.A2.holm\endcsname{1.000}
\expandafter\def\csname gv.A2.supported\endcsname{no}
\expandafter\def\csname gv.A2.npos\endcsname{576}
\expandafter\def\csname gv.A2.methodci\endcsname{[0.643, 0.705]}
\expandafter\def\csname gv.A2.refci\endcsname{[0.650, 0.682]}
\expandafter\def\csname gv.A3.method\endcsname{0.501}
\expandafter\def\csname gv.A3.ref\endcsname{0.692}
\expandafter\def\csname gv.A3.diff\endcsname{$-$0.191}
\expandafter\def\csname gv.A3.diffci\endcsname{[$-$0.225, $-$0.152]}
\expandafter\def\csname gv.A3.p\endcsname{1.000}
\expandafter\def\csname gv.A3.holm\endcsname{1.000}
\expandafter\def\csname gv.A3.supported\endcsname{no}
\expandafter\def\csname gv.A3.npos\endcsname{240}
\expandafter\def\csname gv.A3.methodci\endcsname{[0.462, 0.549]}
\expandafter\def\csname gv.A3.refci\endcsname{[0.631, 0.746]}
\expandafter\def\csname gv.A4.method\endcsname{0.272}
\expandafter\def\csname gv.A4.ref\endcsname{0.646}
\expandafter\def\csname gv.A4.diff\endcsname{$-$0.374}
\expandafter\def\csname gv.A4.diffci\endcsname{[$-$0.443, $-$0.292]}
\expandafter\def\csname gv.A4.p\endcsname{1.000}
\expandafter\def\csname gv.A4.holm\endcsname{1.000}
\expandafter\def\csname gv.A4.supported\endcsname{no}
\expandafter\def\csname gv.A4.npos\endcsname{96}
\expandafter\def\csname gv.A4.methodci\endcsname{[0.220, 0.337]}
\expandafter\def\csname gv.A4.refci\endcsname{[0.536, 0.755]}
\expandafter\def\csname gv.A5.method\endcsname{0.752}
\expandafter\def\csname gv.A5.ref\endcsname{0.861}
\expandafter\def\csname gv.A5.diff\endcsname{$-$0.109}
\expandafter\def\csname gv.A5.diffci\endcsname{[$-$0.127, $-$0.089]}
\expandafter\def\csname gv.A5.p\endcsname{1.000}
\expandafter\def\csname gv.A5.holm\endcsname{1.000}
\expandafter\def\csname gv.A5.supported\endcsname{no}
\expandafter\def\csname gv.A5.npos\endcsname{672}
\expandafter\def\csname gv.A5.methodci\endcsname{[0.719, 0.784]}
\expandafter\def\csname gv.A5.refci\endcsname{[0.832, 0.884]}
\expandafter\def\csname gv.B1.method\endcsname{0.707}
\expandafter\def\csname gv.B1.ref\endcsname{0.504}
\expandafter\def\csname gv.B1.diff\endcsname{+0.202}
\expandafter\def\csname gv.B1.diffci\endcsname{[+0.147, +0.256]}
\expandafter\def\csname gv.B1.p\endcsname{0.001}
\expandafter\def\csname gv.B1.holm\endcsname{0.009}
\expandafter\def\csname gv.B1.supported\endcsname{yes}
\expandafter\def\csname gv.B1.npos\endcsname{336}
\expandafter\def\csname gv.B1.methodci\endcsname{[0.669, 0.754]}
\expandafter\def\csname gv.B1.refci\endcsname{[0.453, 0.559]}
\expandafter\def\csname gv.B2.method\endcsname{0.695}
\expandafter\def\csname gv.B2.ref\endcsname{0.602}
\expandafter\def\csname gv.B2.diff\endcsname{+0.094}
\expandafter\def\csname gv.B2.diffci\endcsname{[+0.045, +0.139]}
\expandafter\def\csname gv.B2.p\endcsname{0.001}
\expandafter\def\csname gv.B2.holm\endcsname{0.009}
\expandafter\def\csname gv.B2.supported\endcsname{yes}
\expandafter\def\csname gv.B2.npos\endcsname{336}
\expandafter\def\csname gv.B2.methodci\endcsname{[0.672, 0.721]}
\expandafter\def\csname gv.B2.refci\endcsname{[0.561, 0.645]}
\expandafter\def\csname gv.C1.method\endcsname{0.441}
\expandafter\def\csname gv.C1.ref\endcsname{0.090}
\expandafter\def\csname gv.C1.diff\endcsname{+0.350}
\expandafter\def\csname gv.C1.diffci\endcsname{[+0.299, +0.395]}
\expandafter\def\csname gv.C1.p\endcsname{0.001}
\expandafter\def\csname gv.C1.holm\endcsname{0.009}
\expandafter\def\csname gv.C1.supported\endcsname{yes}
\expandafter\def\csname gv.C1.npos\endcsname{240}
\expandafter\def\csname gv.C1.methodci\endcsname{[0.396, 0.488]}
\expandafter\def\csname gv.C1.refci\endcsname{[0.074, 0.113]}
\expandafter\def\csname gv.C2.method\endcsname{0.731}
\expandafter\def\csname gv.C2.ref\endcsname{0.684}
\expandafter\def\csname gv.C2.diff\endcsname{+0.047}
\expandafter\def\csname gv.C2.diffci\endcsname{[+0.016, +0.080]}
\expandafter\def\csname gv.C2.p\endcsname{0.003}
\expandafter\def\csname gv.C2.holm\endcsname{0.015}
\expandafter\def\csname gv.C2.supported\endcsname{yes}
\expandafter\def\csname gv.C2.npos\endcsname{672}
\expandafter\def\csname gv.C2.methodci\endcsname{[0.695, 0.766]}
\expandafter\def\csname gv.C2.refci\endcsname{[0.650, 0.715]}
\expandafter\def\csname gv.nsupported\endcsname{5}
\expandafter\def\csname gv.nfamily\endcsname{9}
\expandafter\def\csname gv.nnotsupported\endcsname{4}
\expandafter\def\csname gv.supportedlist\endcsname{A1, B1, B2, C1, C2}
\expandafter\def\csname gv.notsupportedlist\endcsname{A2, A3, A4, A5}
\expandafter\def\csname gv.A.link.rules-v2.auroc\endcsname{0.808}
\expandafter\def\csname gv.A.link.rules-v2.auprc\endcsname{0.347}
\expandafter\def\csname gv.A.link.rules-v2.auprc.ci\endcsname{[0.321, 0.376]}
\expandafter\def\csname gv.A.link.rules-v2.auprc|hop:1\endcsname{0.324}
\expandafter\def\csname gv.A.link.rules-v2.auprc|hop:2\endcsname{0.650}
\expandafter\def\csname gv.A.link.rules-v2.auprc|hop:3\endcsname{0.081}
\expandafter\def\csname gv.A.link.rules-v2.auprc|hop:3.ci\endcsname{[0.066, 0.098]}
\expandafter\def\csname gv.A.link.rules-v2.auprc|hop:4\endcsname{0.000}
\expandafter\def\csname gv.A.link.rules-v2.auprc|hop:4.ci\endcsname{[0.000, 0.000]}
\expandafter\def\csname gv.A.link.rules-v2.twins:auroc\endcsname{0.796}
\expandafter\def\csname gv.A.link.rules-v2.twins:auroc.ci\endcsname{[0.773, 0.818]}
\expandafter\def\csname gv.A.link.rules-v2.twins:auroc|hop:1\endcsname{0.879}
\expandafter\def\csname gv.A.link.rules-v2.twins:auroc|hop:2\endcsname{0.882}
\expandafter\def\csname gv.A.link.rules-v2.twins:auroc|hop:3\endcsname{0.746}
\expandafter\def\csname gv.A.link.rules-v2.twins:auroc|hop:4\endcsname{0.710}
\expandafter\def\csname gv.A.link.rules-generic.auroc\endcsname{0.816}
\expandafter\def\csname gv.A.link.rules-generic.auprc\endcsname{0.186}
\expandafter\def\csname gv.A.link.rules-generic.auprc.ci\endcsname{[0.173, 0.203]}
\expandafter\def\csname gv.A.link.rules-generic.auprc|hop:1\endcsname{0.084}
\expandafter\def\csname gv.A.link.rules-generic.auprc|hop:2\endcsname{0.292}
\expandafter\def\csname gv.A.link.rules-generic.auprc|hop:3\endcsname{0.025}
\expandafter\def\csname gv.A.link.rules-generic.auprc|hop:3.ci\endcsname{[0.022, 0.029]}
\expandafter\def\csname gv.A.link.rules-generic.auprc|hop:4\endcsname{0.000}
\expandafter\def\csname gv.A.link.rules-generic.auprc|hop:4.ci\endcsname{[0.000, 0.000]}
\expandafter\def\csname gv.A.link.rules-generic.twins:auroc\endcsname{0.667}
\expandafter\def\csname gv.A.link.rules-generic.twins:auroc.ci\endcsname{[0.650, 0.682]}
\expandafter\def\csname gv.A.link.rules-generic.twins:auroc|hop:1\endcsname{0.810}
\expandafter\def\csname gv.A.link.rules-generic.twins:auroc|hop:2\endcsname{0.876}
\expandafter\def\csname gv.A.link.rules-generic.twins:auroc|hop:3\endcsname{0.556}
\expandafter\def\csname gv.A.link.rules-generic.twins:auroc|hop:4\endcsname{0.459}
\expandafter\def\csname gv.A.link.path-lr.auroc\endcsname{0.981}
\expandafter\def\csname gv.A.link.path-lr.auprc\endcsname{0.295}
\expandafter\def\csname gv.A.link.path-lr.auprc.ci\endcsname{[0.281, 0.315]}
\expandafter\def\csname gv.A.link.path-lr.auprc|hop:1\endcsname{0.051}
\expandafter\def\csname gv.A.link.path-lr.auprc|hop:2\endcsname{0.177}
\expandafter\def\csname gv.A.link.path-lr.auprc|hop:3\endcsname{0.239}
\expandafter\def\csname gv.A.link.path-lr.auprc|hop:3.ci\endcsname{[0.222, 0.260]}
\expandafter\def\csname gv.A.link.path-lr.auprc|hop:4\endcsname{0.001}
\expandafter\def\csname gv.A.link.path-lr.auprc|hop:4.ci\endcsname{[0.001, 0.002]}
\expandafter\def\csname gv.A.link.path-lr.twins:auroc\endcsname{0.619}
\expandafter\def\csname gv.A.link.path-lr.twins:auroc.ci\endcsname{[0.602, 0.637]}
\expandafter\def\csname gv.A.link.path-lr.twins:auroc|hop:1\endcsname{0.699}
\expandafter\def\csname gv.A.link.path-lr.twins:auroc|hop:2\endcsname{0.671}
\expandafter\def\csname gv.A.link.path-lr.twins:auroc|hop:3\endcsname{0.746}
\expandafter\def\csname gv.A.link.path-lr.twins:auroc|hop:4\endcsname{0.158}
\expandafter\def\csname gv.A.link.cj-count.auroc\endcsname{0.957}
\expandafter\def\csname gv.A.link.cj-count.auprc\endcsname{0.311}
\expandafter\def\csname gv.A.link.cj-count.auprc.ci\endcsname{[0.309, 0.313]}
\expandafter\def\csname gv.A.link.cj-count.auprc|hop:1\endcsname{0.000}
\expandafter\def\csname gv.A.link.cj-count.auprc|hop:2\endcsname{0.282}
\expandafter\def\csname gv.A.link.cj-count.auprc|hop:3\endcsname{0.220}
\expandafter\def\csname gv.A.link.cj-count.auprc|hop:3.ci\endcsname{[0.209, 0.230]}
\expandafter\def\csname gv.A.link.cj-count.auprc|hop:4\endcsname{0.104}
\expandafter\def\csname gv.A.link.cj-count.auprc|hop:4.ci\endcsname{[0.092, 0.117]}
\expandafter\def\csname gv.A.link.cj-count.twins:auroc\endcsname{0.500}
\expandafter\def\csname gv.A.link.cj-count.twins:auroc.ci\endcsname{[0.494, 0.506]}
\expandafter\def\csname gv.A.link.cj-count.twins:auroc|hop:1\endcsname{0.042}
\expandafter\def\csname gv.A.link.cj-count.twins:auroc|hop:2\endcsname{0.838}
\expandafter\def\csname gv.A.link.cj-count.twins:auroc|hop:3\endcsname{0.453}
\expandafter\def\csname gv.A.link.cj-count.twins:auroc|hop:4\endcsname{0.167}
\expandafter\def\csname gv.A.link.cj-gbdt.auroc\endcsname{1.000}
\expandafter\def\csname gv.A.link.cj-gbdt.auprc\endcsname{0.823}
\expandafter\def\csname gv.A.link.cj-gbdt.auprc.ci\endcsname{[0.788, 0.856]}
\expandafter\def\csname gv.A.link.cj-gbdt.auprc|hop:1\endcsname{0.170}
\expandafter\def\csname gv.A.link.cj-gbdt.auprc|hop:2\endcsname{0.603}
\expandafter\def\csname gv.A.link.cj-gbdt.auprc|hop:3\endcsname{0.692}
\expandafter\def\csname gv.A.link.cj-gbdt.auprc|hop:3.ci\endcsname{[0.631, 0.746]}
\expandafter\def\csname gv.A.link.cj-gbdt.auprc|hop:4\endcsname{0.646}
\expandafter\def\csname gv.A.link.cj-gbdt.auprc|hop:4.ci\endcsname{[0.536, 0.755]}
\expandafter\def\csname gv.A.link.cj-gbdt.twins:auroc\endcsname{0.853}
\expandafter\def\csname gv.A.link.cj-gbdt.twins:auroc.ci\endcsname{[0.826, 0.876]}
\expandafter\def\csname gv.A.link.cj-gbdt.twins:auroc|hop:1\endcsname{0.720}
\expandafter\def\csname gv.A.link.cj-gbdt.twins:auroc|hop:2\endcsname{0.845}
\expandafter\def\csname gv.A.link.cj-gbdt.twins:auroc|hop:3\endcsname{0.861}
\expandafter\def\csname gv.A.link.cj-gbdt.twins:auroc|hop:4\endcsname{0.916}
\expandafter\def\csname gv.A.link.grail-k3.auroc\endcsname{0.999}
\expandafter\def\csname gv.A.link.grail-k3.auprc\endcsname{0.663}
\expandafter\def\csname gv.A.link.grail-k3.auprc.ci\endcsname{[0.630, 0.698]}
\expandafter\def\csname gv.A.link.grail-k3.auprc|hop:1\endcsname{0.140}
\expandafter\def\csname gv.A.link.grail-k3.auprc|hop:2\endcsname{0.442}
\expandafter\def\csname gv.A.link.grail-k3.auprc|hop:3\endcsname{0.501}
\expandafter\def\csname gv.A.link.grail-k3.auprc|hop:3.ci\endcsname{[0.462, 0.549]}
\expandafter\def\csname gv.A.link.grail-k3.auprc|hop:4\endcsname{0.272}
\expandafter\def\csname gv.A.link.grail-k3.auprc|hop:4.ci\endcsname{[0.220, 0.337]}
\expandafter\def\csname gv.A.link.grail-k3.twins:auroc\endcsname{0.674}
\expandafter\def\csname gv.A.link.grail-k3.twins:auroc.ci\endcsname{[0.643, 0.705]}
\expandafter\def\csname gv.A.link.grail-k3.twins:auroc|hop:1\endcsname{0.315}
\expandafter\def\csname gv.A.link.grail-k3.twins:auroc|hop:2\endcsname{0.678}
\expandafter\def\csname gv.A.link.grail-k3.twins:auroc|hop:3\endcsname{0.739}
\expandafter\def\csname gv.A.link.grail-k3.twins:auroc|hop:4\endcsname{0.682}
\expandafter\def\csname gv.A.link.nbfnet.auroc\endcsname{0.999}
\expandafter\def\csname gv.A.link.nbfnet.auprc\endcsname{0.673}
\expandafter\def\csname gv.A.link.nbfnet.auprc.ci\endcsname{[0.645, 0.698]}
\expandafter\def\csname gv.A.link.nbfnet.auprc|hop:1\endcsname{0.386}
\expandafter\def\csname gv.A.link.nbfnet.auprc|hop:2\endcsname{0.518}
\expandafter\def\csname gv.A.link.nbfnet.auprc|hop:3\endcsname{0.461}
\expandafter\def\csname gv.A.link.nbfnet.auprc|hop:3.ci\endcsname{[0.420, 0.502]}
\expandafter\def\csname gv.A.link.nbfnet.auprc|hop:4\endcsname{0.154}
\expandafter\def\csname gv.A.link.nbfnet.auprc|hop:4.ci\endcsname{[0.132, 0.188]}
\expandafter\def\csname gv.A.link.nbfnet.twins:auroc\endcsname{0.668}
\expandafter\def\csname gv.A.link.nbfnet.twins:auroc.ci\endcsname{[0.642, 0.691]}
\expandafter\def\csname gv.A.link.nbfnet.twins:auroc|hop:1\endcsname{0.569}
\expandafter\def\csname gv.A.link.nbfnet.twins:auroc|hop:2\endcsname{0.763}
\expandafter\def\csname gv.A.link.nbfnet.twins:auroc|hop:3\endcsname{0.648}
\expandafter\def\csname gv.A.link.nbfnet.twins:auroc|hop:4\endcsname{0.574}
\expandafter\def\csname gv.A.link.osfa-insp.auroc\endcsname{0.999}
\expandafter\def\csname gv.A.link.osfa-insp.auprc\endcsname{0.584}
\expandafter\def\csname gv.A.link.osfa-insp.auprc.ci\endcsname{[0.559, 0.617]}
\expandafter\def\csname gv.A.link.osfa-insp.auprc|hop:1\endcsname{0.045}
\expandafter\def\csname gv.A.link.osfa-insp.auprc|hop:2\endcsname{0.371}
\expandafter\def\csname gv.A.link.osfa-insp.auprc|hop:3\endcsname{0.394}
\expandafter\def\csname gv.A.link.osfa-insp.auprc|hop:3.ci\endcsname{[0.357, 0.444]}
\expandafter\def\csname gv.A.link.osfa-insp.auprc|hop:4\endcsname{0.197}
\expandafter\def\csname gv.A.link.osfa-insp.auprc|hop:4.ci\endcsname{[0.174, 0.229]}
\expandafter\def\csname gv.A.link.osfa-insp.twins:auroc\endcsname{0.635}
\expandafter\def\csname gv.A.link.osfa-insp.twins:auroc.ci\endcsname{[0.611, 0.663]}
\expandafter\def\csname gv.A.link.osfa-insp.twins:auroc|hop:1\endcsname{0.306}
\expandafter\def\csname gv.A.link.osfa-insp.twins:auroc|hop:2\endcsname{0.663}
\expandafter\def\csname gv.A.link.osfa-insp.twins:auroc|hop:3\endcsname{0.664}
\expandafter\def\csname gv.A.link.osfa-insp.twins:auroc|hop:4\endcsname{0.674}
\expandafter\def\csname gv.A.link.rules-v2+grail-k3:lr.auroc\endcsname{1.000}
\expandafter\def\csname gv.A.link.rules-v2+grail-k3:lr.auprc\endcsname{0.735}
\expandafter\def\csname gv.A.link.rules-v2+grail-k3:lr.auprc.ci\endcsname{[0.711, 0.763]}
\expandafter\def\csname gv.A.link.rules-v2+grail-k3:lr.auprc|hop:1\endcsname{0.201}
\expandafter\def\csname gv.A.link.rules-v2+grail-k3:lr.auprc|hop:2\endcsname{0.730}
\expandafter\def\csname gv.A.link.rules-v2+grail-k3:lr.auprc|hop:3\endcsname{0.437}
\expandafter\def\csname gv.A.link.rules-v2+grail-k3:lr.auprc|hop:3.ci\endcsname{[0.412, 0.468]}
\expandafter\def\csname gv.A.link.rules-v2+grail-k3:lr.auprc|hop:4\endcsname{0.190}
\expandafter\def\csname gv.A.link.rules-v2+grail-k3:lr.auprc|hop:4.ci\endcsname{[0.160, 0.224]}
\expandafter\def\csname gv.A.link.rules-v2+grail-k3:lr.twins:auroc\endcsname{0.763}
\expandafter\def\csname gv.A.link.rules-v2+grail-k3:lr.twins:auroc.ci\endcsname{[0.742, 0.783]}
\expandafter\def\csname gv.A.link.rules-v2+grail-k3:lr.twins:auroc|hop:1\endcsname{0.708}
\expandafter\def\csname gv.A.link.rules-v2+grail-k3:lr.twins:auroc|hop:2\endcsname{0.889}
\expandafter\def\csname gv.A.link.rules-v2+grail-k3:lr.twins:auroc|hop:3\endcsname{0.723}
\expandafter\def\csname gv.A.link.rules-v2+grail-k3:lr.twins:auroc|hop:4\endcsname{0.636}
\expandafter\def\csname gv.A.entity.rules-v2.auroc\endcsname{0.872}
\expandafter\def\csname gv.A.entity.rules-v2.auprc\endcsname{0.698}
\expandafter\def\csname gv.A.entity.rules-v2.auprc.ci\endcsname{[0.669, 0.728]}
\expandafter\def\csname gv.A.entity.rules-v2.twins:auroc\endcsname{0.817}
\expandafter\def\csname gv.A.entity.rules-v2.twins:auroc.ci\endcsname{[0.793, 0.839]}
\expandafter\def\csname gv.A.entity.rules-v2.auprc@prev0.01\endcsname{0.089}
\expandafter\def\csname gv.A.entity.rules-v2.auprc@prev0.02\endcsname{0.158}
\expandafter\def\csname gv.A.entity.rules-v2.auprc@prev0.05\endcsname{0.308}
\expandafter\def\csname gv.A.entity.rules-v2.p@budget0.01\endcsname{0.864}
\expandafter\def\csname gv.A.entity.rules-v2.p@budget0.02\endcsname{0.858}
\expandafter\def\csname gv.A.entity.rules-v2.p@budget0.05\endcsname{0.826}
\expandafter\def\csname gv.A.entity.rules-generic.auroc\endcsname{0.780}
\expandafter\def\csname gv.A.entity.rules-generic.auprc\endcsname{0.490}
\expandafter\def\csname gv.A.entity.rules-generic.auprc.ci\endcsname{[0.467, 0.515]}
\expandafter\def\csname gv.A.entity.rules-generic.twins:auroc\endcsname{0.735}
\expandafter\def\csname gv.A.entity.rules-generic.twins:auroc.ci\endcsname{[0.712, 0.756]}
\expandafter\def\csname gv.A.entity.rules-generic.auprc@prev0.01\endcsname{0.028}
\expandafter\def\csname gv.A.entity.rules-generic.auprc@prev0.02\endcsname{0.054}
\expandafter\def\csname gv.A.entity.rules-generic.auprc@prev0.05\endcsname{0.127}
\expandafter\def\csname gv.A.entity.rules-generic.p@budget0.01\endcsname{0.533}
\expandafter\def\csname gv.A.entity.rules-generic.p@budget0.02\endcsname{0.578}
\expandafter\def\csname gv.A.entity.rules-generic.p@budget0.05\endcsname{0.616}
\expandafter\def\csname gv.A.entity.path-lr.auroc\endcsname{0.876}
\expandafter\def\csname gv.A.entity.path-lr.auprc\endcsname{0.725}
\expandafter\def\csname gv.A.entity.path-lr.auprc.ci\endcsname{[0.692, 0.757]}
\expandafter\def\csname gv.A.entity.path-lr.twins:auroc\endcsname{0.815}
\expandafter\def\csname gv.A.entity.path-lr.twins:auroc.ci\endcsname{[0.788, 0.837]}
\expandafter\def\csname gv.A.entity.path-lr.auprc@prev0.01\endcsname{0.092}
\expandafter\def\csname gv.A.entity.path-lr.auprc@prev0.02\endcsname{0.165}
\expandafter\def\csname gv.A.entity.path-lr.auprc@prev0.05\endcsname{0.321}
\expandafter\def\csname gv.A.entity.path-lr.p@budget0.01\endcsname{0.917}
\expandafter\def\csname gv.A.entity.path-lr.p@budget0.02\endcsname{0.889}
\expandafter\def\csname gv.A.entity.path-lr.p@budget0.05\endcsname{0.908}
\expandafter\def\csname gv.A.entity.cj-count.auroc\endcsname{0.628}
\expandafter\def\csname gv.A.entity.cj-count.auprc\endcsname{0.325}
\expandafter\def\csname gv.A.entity.cj-count.auprc.ci\endcsname{[0.323, 0.327]}
\expandafter\def\csname gv.A.entity.cj-count.twins:auroc\endcsname{0.500}
\expandafter\def\csname gv.A.entity.cj-count.twins:auroc.ci\endcsname{[0.495, 0.504]}
\expandafter\def\csname gv.A.entity.cj-count.auprc@prev0.01\endcsname{0.013}
\expandafter\def\csname gv.A.entity.cj-count.auprc@prev0.02\endcsname{0.027}
\expandafter\def\csname gv.A.entity.cj-count.auprc@prev0.05\endcsname{0.065}
\expandafter\def\csname gv.A.entity.cj-count.p@budget0.01\endcsname{0.100}
\expandafter\def\csname gv.A.entity.cj-count.p@budget0.02\endcsname{0.117}
\expandafter\def\csname gv.A.entity.cj-count.p@budget0.05\endcsname{0.146}
\expandafter\def\csname gv.A.entity.cj-gbdt-ent.auroc\endcsname{0.952}
\expandafter\def\csname gv.A.entity.cj-gbdt-ent.auprc\endcsname{0.861}
\expandafter\def\csname gv.A.entity.cj-gbdt-ent.auprc.ci\endcsname{[0.832, 0.884]}
\expandafter\def\csname gv.A.entity.cj-gbdt-ent.twins:auroc\endcsname{0.881}
\expandafter\def\csname gv.A.entity.cj-gbdt-ent.twins:auroc.ci\endcsname{[0.860, 0.898]}
\expandafter\def\csname gv.A.entity.cj-gbdt-ent.auprc@prev0.01\endcsname{0.210}
\expandafter\def\csname gv.A.entity.cj-gbdt-ent.auprc@prev0.02\endcsname{0.329}
\expandafter\def\csname gv.A.entity.cj-gbdt-ent.auprc@prev0.05\endcsname{0.526}
\expandafter\def\csname gv.A.entity.cj-gbdt-ent.p@budget0.01\endcsname{0.958}
\expandafter\def\csname gv.A.entity.cj-gbdt-ent.p@budget0.02\endcsname{0.958}
\expandafter\def\csname gv.A.entity.cj-gbdt-ent.p@budget0.05\endcsname{0.950}
\expandafter\def\csname gv.A.entity.rgcn-ent.auroc\endcsname{0.900}
\expandafter\def\csname gv.A.entity.rgcn-ent.auprc\endcsname{0.776}
\expandafter\def\csname gv.A.entity.rgcn-ent.auprc.ci\endcsname{[0.744, 0.808]}
\expandafter\def\csname gv.A.entity.rgcn-ent.twins:auroc\endcsname{0.821}
\expandafter\def\csname gv.A.entity.rgcn-ent.twins:auroc.ci\endcsname{[0.796, 0.846]}
\expandafter\def\csname gv.A.entity.rgcn-ent.auprc@prev0.01\endcsname{0.147}
\expandafter\def\csname gv.A.entity.rgcn-ent.auprc@prev0.02\endcsname{0.237}
\expandafter\def\csname gv.A.entity.rgcn-ent.auprc@prev0.05\endcsname{0.405}
\expandafter\def\csname gv.A.entity.rgcn-ent.p@budget0.01\endcsname{0.903}
\expandafter\def\csname gv.A.entity.rgcn-ent.p@budget0.02\endcsname{0.931}
\expandafter\def\csname gv.A.entity.rgcn-ent.p@budget0.05\endcsname{0.922}
\expandafter\def\csname gv.A.entity.rdl-ent.auroc\endcsname{0.890}
\expandafter\def\csname gv.A.entity.rdl-ent.auprc\endcsname{0.752}
\expandafter\def\csname gv.A.entity.rdl-ent.auprc.ci\endcsname{[0.719, 0.784]}
\expandafter\def\csname gv.A.entity.rdl-ent.twins:auroc\endcsname{0.821}
\expandafter\def\csname gv.A.entity.rdl-ent.twins:auroc.ci\endcsname{[0.797, 0.843]}
\expandafter\def\csname gv.A.entity.rdl-ent.auprc@prev0.01\endcsname{0.119}
\expandafter\def\csname gv.A.entity.rdl-ent.auprc@prev0.02\endcsname{0.198}
\expandafter\def\csname gv.A.entity.rdl-ent.auprc@prev0.05\endcsname{0.361}
\expandafter\def\csname gv.A.entity.rdl-ent.p@budget0.01\endcsname{0.889}
\expandafter\def\csname gv.A.entity.rdl-ent.p@budget0.02\endcsname{0.896}
\expandafter\def\csname gv.A.entity.rdl-ent.p@budget0.05\endcsname{0.881}
\expandafter\def\csname gv.A.entity.rules-v2+rdl-ent:lr.auroc\endcsname{0.914}
\expandafter\def\csname gv.A.entity.rules-v2+rdl-ent:lr.auprc\endcsname{0.792}
\expandafter\def\csname gv.A.entity.rules-v2+rdl-ent:lr.auprc.ci\endcsname{[0.763, 0.822]}
\expandafter\def\csname gv.A.entity.rules-v2+rdl-ent:lr.twins:auroc\endcsname{0.844}
\expandafter\def\csname gv.A.entity.rules-v2+rdl-ent:lr.twins:auroc.ci\endcsname{[0.822, 0.866]}
\expandafter\def\csname gv.A.entity.rules-v2+rdl-ent:lr.auprc@prev0.01\endcsname{0.132}
\expandafter\def\csname gv.A.entity.rules-v2+rdl-ent:lr.auprc@prev0.02\endcsname{0.225}
\expandafter\def\csname gv.A.entity.rules-v2+rdl-ent:lr.auprc@prev0.05\endcsname{0.407}
\expandafter\def\csname gv.A.entity.rules-v2+rdl-ent:lr.p@budget0.01\endcsname{0.917}
\expandafter\def\csname gv.A.entity.rules-v2+rdl-ent:lr.p@budget0.02\endcsname{0.889}
\expandafter\def\csname gv.A.entity.rules-v2+rdl-ent:lr.p@budget0.05\endcsname{0.894}
\expandafter\def\csname gv.A.vocab\endcsname{24}
\expandafter\def\csname gv.A.linkfeatures\endcsname{188}
\expandafter\def\csname gv.A.gbdt.link.hp\endcsname{learning rate 0.05, 63 leaves, min.\ leaf size 20, $\ell_2$ 1}
\expandafter\def\csname gv.A.gbdt.link.val\endcsname{0.863}
\expandafter\def\csname gv.A.gbdt.link.ngrid\endcsname{8}
\expandafter\def\csname gv.A.gbdt.ent.hp\endcsname{learning rate 0.05, 7 leaves, min.\ leaf size 10, $\ell_2$ 0}
\expandafter\def\csname gv.A.gbdt.ent.val\endcsname{0.881}
\expandafter\def\csname gv.A.gbdt.ent.ngrid\endcsname{16}
\expandafter\def\csname gv.B.nent\endcsname{16,800}
\expandafter\def\csname gv.B.npos\endcsname{336}
\expandafter\def\csname gv.B.ntwins\endcsname{336}
\expandafter\def\csname gv.B.prevpct\endcsname{2.0}
\expandafter\def\csname gv.B.rules-v2.auroc\endcsname{0.900}
\expandafter\def\csname gv.B.rules-v2.auprc\endcsname{0.504}
\expandafter\def\csname gv.B.rules-v2.auprc.ci\endcsname{[0.453, 0.559]}
\expandafter\def\csname gv.B.rules-v2.twins:auroc\endcsname{0.792}
\expandafter\def\csname gv.B.rules-v2.twins:auroc.ci\endcsname{[0.767, 0.818]}
\expandafter\def\csname gv.B.rules-v2.p@budget0.01\endcsname{0.800}
\expandafter\def\csname gv.B.rules-v2.r@budget0.01\endcsname{0.400}
\expandafter\def\csname gv.B.rules-v2.p@budget0.02\endcsname{0.602}
\expandafter\def\csname gv.B.rules-v2.p@budget0.02.ci\endcsname{[0.561, 0.645]}
\expandafter\def\csname gv.B.rules-v2.r@budget0.02\endcsname{0.602}
\expandafter\def\csname gv.B.rules-v2.p@budget0.05\endcsname{0.267}
\expandafter\def\csname gv.B.rules-v2.r@budget0.05\endcsname{0.668}
\expandafter\def\csname gv.B.rules-v2.auprc@prev0.01\endcsname{0.405}
\expandafter\def\csname gv.B.rules-v2.auprc@prev0.02\endcsname{0.504}
\expandafter\def\csname gv.B.rules-v2.auprc@prev0.05\endcsname{0.615}
\expandafter\def\csname gv.B.rules-generic.auroc\endcsname{0.855}
\expandafter\def\csname gv.B.rules-generic.auprc\endcsname{0.155}
\expandafter\def\csname gv.B.rules-generic.auprc.ci\endcsname{[0.123, 0.198]}
\expandafter\def\csname gv.B.rules-generic.twins:auroc\endcsname{0.728}
\expandafter\def\csname gv.B.rules-generic.twins:auroc.ci\endcsname{[0.699, 0.759]}
\expandafter\def\csname gv.B.rules-generic.p@budget0.01\endcsname{0.343}
\expandafter\def\csname gv.B.rules-generic.r@budget0.01\endcsname{0.171}
\expandafter\def\csname gv.B.rules-generic.p@budget0.02\endcsname{0.239}
\expandafter\def\csname gv.B.rules-generic.p@budget0.02.ci\endcsname{[0.203, 0.276]}
\expandafter\def\csname gv.B.rules-generic.r@budget0.02\endcsname{0.239}
\expandafter\def\csname gv.B.rules-generic.p@budget0.05\endcsname{0.174}
\expandafter\def\csname gv.B.rules-generic.r@budget0.05\endcsname{0.434}
\expandafter\def\csname gv.B.rules-generic.auprc@prev0.01\endcsname{0.091}
\expandafter\def\csname gv.B.rules-generic.auprc@prev0.02\endcsname{0.155}
\expandafter\def\csname gv.B.rules-generic.auprc@prev0.05\endcsname{0.285}
\expandafter\def\csname gv.B.cj-count.auroc\endcsname{0.871}
\expandafter\def\csname gv.B.cj-count.auprc\endcsname{0.211}
\expandafter\def\csname gv.B.cj-count.auprc.ci\endcsname{[0.208, 0.215]}
\expandafter\def\csname gv.B.cj-count.twins:auroc\endcsname{0.500}
\expandafter\def\csname gv.B.cj-count.twins:auroc.ci\endcsname{[0.493, 0.507]}
\expandafter\def\csname gv.B.cj-count.p@budget0.01\endcsname{0.098}
\expandafter\def\csname gv.B.cj-count.r@budget0.01\endcsname{0.049}
\expandafter\def\csname gv.B.cj-count.p@budget0.02\endcsname{0.184}
\expandafter\def\csname gv.B.cj-count.p@budget0.02.ci\endcsname{[0.171, 0.197]}
\expandafter\def\csname gv.B.cj-count.r@budget0.02\endcsname{0.184}
\expandafter\def\csname gv.B.cj-count.p@budget0.05\endcsname{0.314}
\expandafter\def\csname gv.B.cj-count.r@budget0.05\endcsname{0.786}
\expandafter\def\csname gv.B.cj-count.auprc@prev0.01\endcsname{0.121}
\expandafter\def\csname gv.B.cj-count.auprc@prev0.02\endcsname{0.211}
\expandafter\def\csname gv.B.cj-count.auprc@prev0.05\endcsname{0.382}
\expandafter\def\csname gv.B.cj-gbdt-ent.auroc\endcsname{0.995}
\expandafter\def\csname gv.B.cj-gbdt-ent.auprc\endcsname{0.798}
\expandafter\def\csname gv.B.cj-gbdt-ent.auprc.ci\endcsname{[0.761, 0.838]}
\expandafter\def\csname gv.B.cj-gbdt-ent.twins:auroc\endcsname{0.828}
\expandafter\def\csname gv.B.cj-gbdt-ent.twins:auroc.ci\endcsname{[0.798, 0.858]}
\expandafter\def\csname gv.B.cj-gbdt-ent.p@budget0.01\endcsname{0.851}
\expandafter\def\csname gv.B.cj-gbdt-ent.r@budget0.01\endcsname{0.426}
\expandafter\def\csname gv.B.cj-gbdt-ent.p@budget0.02\endcsname{0.735}
\expandafter\def\csname gv.B.cj-gbdt-ent.p@budget0.02.ci\endcsname{[0.702, 0.768]}
\expandafter\def\csname gv.B.cj-gbdt-ent.r@budget0.02\endcsname{0.735}
\expandafter\def\csname gv.B.cj-gbdt-ent.p@budget0.05\endcsname{0.398}
\expandafter\def\csname gv.B.cj-gbdt-ent.r@budget0.05\endcsname{0.994}
\expandafter\def\csname gv.B.cj-gbdt-ent.auprc@prev0.01\endcsname{0.681}
\expandafter\def\csname gv.B.cj-gbdt-ent.auprc@prev0.02\endcsname{0.798}
\expandafter\def\csname gv.B.cj-gbdt-ent.auprc@prev0.05\endcsname{0.903}
\expandafter\def\csname gv.B.rgcn-ent.auroc\endcsname{0.974}
\expandafter\def\csname gv.B.rgcn-ent.auroc.sd\endcsname{0.010}
\expandafter\def\csname gv.B.rgcn-ent.auprc\endcsname{0.686}
\expandafter\def\csname gv.B.rgcn-ent.auprc.ci\endcsname{[0.651, 0.726]}
\expandafter\def\csname gv.B.rgcn-ent.auprc.sd\endcsname{0.017}
\expandafter\def\csname gv.B.rgcn-ent.twins:auroc\endcsname{0.749}
\expandafter\def\csname gv.B.rgcn-ent.twins:auroc.ci\endcsname{[0.719, 0.780]}
\expandafter\def\csname gv.B.rgcn-ent.twins:auroc.sd\endcsname{0.014}
\expandafter\def\csname gv.B.rgcn-ent.p@budget0.01\endcsname{0.780}
\expandafter\def\csname gv.B.rgcn-ent.p@budget0.01.sd\endcsname{0.012}
\expandafter\def\csname gv.B.rgcn-ent.r@budget0.01\endcsname{0.390}
\expandafter\def\csname gv.B.rgcn-ent.r@budget0.01.sd\endcsname{0.006}
\expandafter\def\csname gv.B.rgcn-ent.p@budget0.02\endcsname{0.670}
\expandafter\def\csname gv.B.rgcn-ent.p@budget0.02.ci\endcsname{[0.639, 0.700]}
\expandafter\def\csname gv.B.rgcn-ent.p@budget0.02.sd\endcsname{0.017}
\expandafter\def\csname gv.B.rgcn-ent.r@budget0.02\endcsname{0.670}
\expandafter\def\csname gv.B.rgcn-ent.r@budget0.02.sd\endcsname{0.017}
\expandafter\def\csname gv.B.rgcn-ent.p@budget0.05\endcsname{0.358}
\expandafter\def\csname gv.B.rgcn-ent.p@budget0.05.sd\endcsname{0.014}
\expandafter\def\csname gv.B.rgcn-ent.r@budget0.05\endcsname{0.894}
\expandafter\def\csname gv.B.rgcn-ent.r@budget0.05.sd\endcsname{0.034}
\expandafter\def\csname gv.B.rgcn-ent.auprc@prev0.01\endcsname{0.561}
\expandafter\def\csname gv.B.rgcn-ent.auprc@prev0.01.sd\endcsname{0.015}
\expandafter\def\csname gv.B.rgcn-ent.auprc@prev0.02\endcsname{0.686}
\expandafter\def\csname gv.B.rgcn-ent.auprc@prev0.02.sd\endcsname{0.017}
\expandafter\def\csname gv.B.rgcn-ent.auprc@prev0.05\endcsname{0.815}
\expandafter\def\csname gv.B.rgcn-ent.auprc@prev0.05.sd\endcsname{0.021}
\expandafter\def\csname gv.B.rdl-ent.auroc\endcsname{0.959}
\expandafter\def\csname gv.B.rdl-ent.auroc.sd\endcsname{0.002}
\expandafter\def\csname gv.B.rdl-ent.auprc\endcsname{0.707}
\expandafter\def\csname gv.B.rdl-ent.auprc.ci\endcsname{[0.669, 0.754]}
\expandafter\def\csname gv.B.rdl-ent.auprc.sd\endcsname{0.018}
\expandafter\def\csname gv.B.rdl-ent.twins:auroc\endcsname{0.772}
\expandafter\def\csname gv.B.rdl-ent.twins:auroc.ci\endcsname{[0.744, 0.802]}
\expandafter\def\csname gv.B.rdl-ent.twins:auroc.sd\endcsname{0.020}
\expandafter\def\csname gv.B.rdl-ent.p@budget0.01\endcsname{0.808}
\expandafter\def\csname gv.B.rdl-ent.p@budget0.01.sd\endcsname{0.003}
\expandafter\def\csname gv.B.rdl-ent.r@budget0.01\endcsname{0.404}
\expandafter\def\csname gv.B.rdl-ent.r@budget0.01.sd\endcsname{0.002}
\expandafter\def\csname gv.B.rdl-ent.p@budget0.02\endcsname{0.695}
\expandafter\def\csname gv.B.rdl-ent.p@budget0.02.ci\endcsname{[0.672, 0.721]}
\expandafter\def\csname gv.B.rdl-ent.p@budget0.02.sd\endcsname{0.024}
\expandafter\def\csname gv.B.rdl-ent.r@budget0.02\endcsname{0.695}
\expandafter\def\csname gv.B.rdl-ent.r@budget0.02.sd\endcsname{0.024}
\expandafter\def\csname gv.B.rdl-ent.p@budget0.05\endcsname{0.360}
\expandafter\def\csname gv.B.rdl-ent.p@budget0.05.sd\endcsname{0.005}
\expandafter\def\csname gv.B.rdl-ent.r@budget0.05\endcsname{0.900}
\expandafter\def\csname gv.B.rdl-ent.r@budget0.05.sd\endcsname{0.012}
\expandafter\def\csname gv.B.rdl-ent.auprc@prev0.01\endcsname{0.586}
\expandafter\def\csname gv.B.rdl-ent.auprc@prev0.01.sd\endcsname{0.022}
\expandafter\def\csname gv.B.rdl-ent.auprc@prev0.02\endcsname{0.707}
\expandafter\def\csname gv.B.rdl-ent.auprc@prev0.02.sd\endcsname{0.018}
\expandafter\def\csname gv.B.rdl-ent.auprc@prev0.05\endcsname{0.824}
\expandafter\def\csname gv.B.rdl-ent.auprc@prev0.05.sd\endcsname{0.012}
\expandafter\def\csname gv.B.rgcn-ent.bestvalmean\endcsname{0.731}
\expandafter\def\csname gv.B.rdl-ent.bestvalmean\endcsname{0.778}
\expandafter\def\csname gv.B.gbdt.hp\endcsname{learning rate 0.05, 7 leaves, min.\ leaf size 10, $\ell_2$ 1}
\expandafter\def\csname gv.C.clean.link.rules-v2.auprc\endcsname{0.347}
\expandafter\def\csname gv.C.clean.link.rules-v2.auprc.ci\endcsname{[0.321, 0.376]}
\expandafter\def\csname gv.C.clean.link.rules-v2.auprc|hop:1\endcsname{0.324}
\expandafter\def\csname gv.C.clean.link.rules-v2.auprc|hop:2\endcsname{0.650}
\expandafter\def\csname gv.C.clean.link.rules-v2.auprc|hop:3\endcsname{0.081}
\expandafter\def\csname gv.C.clean.link.rules-v2.auprc|hop:3.ci\endcsname{[0.066, 0.098]}
\expandafter\def\csname gv.C.clean.link.rules-v2.auprc|hop:4\endcsname{0.000}
\expandafter\def\csname gv.C.clean.link.rules-v2.twins:auroc\endcsname{0.796}
\expandafter\def\csname gv.C.clean.link.rules-generic.auprc\endcsname{0.186}
\expandafter\def\csname gv.C.clean.link.rules-generic.auprc.ci\endcsname{[0.173, 0.203]}
\expandafter\def\csname gv.C.clean.link.rules-generic.auprc|hop:1\endcsname{0.084}
\expandafter\def\csname gv.C.clean.link.rules-generic.auprc|hop:2\endcsname{0.292}
\expandafter\def\csname gv.C.clean.link.rules-generic.auprc|hop:3\endcsname{0.025}
\expandafter\def\csname gv.C.clean.link.rules-generic.auprc|hop:3.ci\endcsname{[0.022, 0.029]}
\expandafter\def\csname gv.C.clean.link.rules-generic.auprc|hop:4\endcsname{0.000}
\expandafter\def\csname gv.C.clean.link.rules-generic.twins:auroc\endcsname{0.667}
\expandafter\def\csname gv.C.clean.link.cj-count.auprc\endcsname{0.311}
\expandafter\def\csname gv.C.clean.link.cj-count.auprc.ci\endcsname{[0.309, 0.313]}
\expandafter\def\csname gv.C.clean.link.cj-count.auprc|hop:1\endcsname{0.000}
\expandafter\def\csname gv.C.clean.link.cj-count.auprc|hop:2\endcsname{0.282}
\expandafter\def\csname gv.C.clean.link.cj-count.auprc|hop:3\endcsname{0.220}
\expandafter\def\csname gv.C.clean.link.cj-count.auprc|hop:3.ci\endcsname{[0.209, 0.230]}
\expandafter\def\csname gv.C.clean.link.cj-count.auprc|hop:4\endcsname{0.104}
\expandafter\def\csname gv.C.clean.link.cj-count.twins:auroc\endcsname{0.500}
\expandafter\def\csname gv.C.clean.link.cj-gbdt.auprc\endcsname{0.823}
\expandafter\def\csname gv.C.clean.link.cj-gbdt.auprc.ci\endcsname{[0.788, 0.856]}
\expandafter\def\csname gv.C.clean.link.cj-gbdt.auprc|hop:1\endcsname{0.170}
\expandafter\def\csname gv.C.clean.link.cj-gbdt.auprc|hop:2\endcsname{0.603}
\expandafter\def\csname gv.C.clean.link.cj-gbdt.auprc|hop:3\endcsname{0.692}
\expandafter\def\csname gv.C.clean.link.cj-gbdt.auprc|hop:3.ci\endcsname{[0.631, 0.746]}
\expandafter\def\csname gv.C.clean.link.cj-gbdt.auprc|hop:4\endcsname{0.646}
\expandafter\def\csname gv.C.clean.link.cj-gbdt.twins:auroc\endcsname{0.853}
\expandafter\def\csname gv.C.clean.link.nbfnet.auprc\endcsname{0.688}
\expandafter\def\csname gv.C.clean.link.nbfnet.auprc.ci\endcsname{[0.660, 0.712]}
\expandafter\def\csname gv.C.clean.link.nbfnet.auprc|hop:1\endcsname{0.402}
\expandafter\def\csname gv.C.clean.link.nbfnet.auprc|hop:2\endcsname{0.514}
\expandafter\def\csname gv.C.clean.link.nbfnet.auprc|hop:3\endcsname{0.502}
\expandafter\def\csname gv.C.clean.link.nbfnet.auprc|hop:3.ci\endcsname{[0.460, 0.547]}
\expandafter\def\csname gv.C.clean.link.nbfnet.auprc|hop:4\endcsname{0.172}
\expandafter\def\csname gv.C.clean.link.nbfnet.twins:auroc\endcsname{0.693}
\expandafter\def\csname gv.C.clean.entity.rules-v2.auprc\endcsname{0.698}
\expandafter\def\csname gv.C.clean.entity.rules-v2.auprc.ci\endcsname{[0.669, 0.728]}
\expandafter\def\csname gv.C.clean.entity.rules-v2.twins:auroc\endcsname{0.817}
\expandafter\def\csname gv.C.clean.entity.rules-generic.auprc\endcsname{0.490}
\expandafter\def\csname gv.C.clean.entity.rules-generic.auprc.ci\endcsname{[0.467, 0.515]}
\expandafter\def\csname gv.C.clean.entity.rules-generic.twins:auroc\endcsname{0.735}
\expandafter\def\csname gv.C.clean.entity.cj-count.auprc\endcsname{0.325}
\expandafter\def\csname gv.C.clean.entity.cj-count.auprc.ci\endcsname{[0.323, 0.327]}
\expandafter\def\csname gv.C.clean.entity.cj-count.twins:auroc\endcsname{0.500}
\expandafter\def\csname gv.C.clean.entity.cj-gbdt-ent.auprc\endcsname{0.861}
\expandafter\def\csname gv.C.clean.entity.cj-gbdt-ent.auprc.ci\endcsname{[0.832, 0.884]}
\expandafter\def\csname gv.C.clean.entity.cj-gbdt-ent.twins:auroc\endcsname{0.881}
\expandafter\def\csname gv.C.clean.entity.rdl-ent.auprc\endcsname{0.752}
\expandafter\def\csname gv.C.clean.entity.rdl-ent.auprc.ci\endcsname{[0.719, 0.784]}
\expandafter\def\csname gv.C.clean.entity.rdl-ent.twins:auroc\endcsname{0.821}
\expandafter\def\csname gv.C.idsplit-10.link.rules-v2.auprc\endcsname{0.305}
\expandafter\def\csname gv.C.idsplit-10.link.rules-v2.auprc|hop:1\endcsname{0.366}
\expandafter\def\csname gv.C.idsplit-10.link.rules-v2.auprc|hop:2\endcsname{0.524}
\expandafter\def\csname gv.C.idsplit-10.link.rules-v2.auprc|hop:3\endcsname{0.084}
\expandafter\def\csname gv.C.idsplit-10.link.rules-v2.auprc|hop:4\endcsname{0.000}
\expandafter\def\csname gv.C.idsplit-10.link.rules-v2.twins:auroc\endcsname{0.784}
\expandafter\def\csname gv.C.idsplit-10.link.rules-generic.auprc\endcsname{0.165}
\expandafter\def\csname gv.C.idsplit-10.link.rules-generic.auprc|hop:1\endcsname{0.101}
\expandafter\def\csname gv.C.idsplit-10.link.rules-generic.auprc|hop:2\endcsname{0.234}
\expandafter\def\csname gv.C.idsplit-10.link.rules-generic.auprc|hop:3\endcsname{0.026}
\expandafter\def\csname gv.C.idsplit-10.link.rules-generic.auprc|hop:4\endcsname{0.000}
\expandafter\def\csname gv.C.idsplit-10.link.rules-generic.twins:auroc\endcsname{0.680}
\expandafter\def\csname gv.C.idsplit-10.link.cj-count.auprc\endcsname{0.251}
\expandafter\def\csname gv.C.idsplit-10.link.cj-count.auprc|hop:1\endcsname{0.000}
\expandafter\def\csname gv.C.idsplit-10.link.cj-count.auprc|hop:2\endcsname{0.216}
\expandafter\def\csname gv.C.idsplit-10.link.cj-count.auprc|hop:3\endcsname{0.198}
\expandafter\def\csname gv.C.idsplit-10.link.cj-count.auprc|hop:4\endcsname{0.066}
\expandafter\def\csname gv.C.idsplit-10.link.cj-count.twins:auroc\endcsname{0.512}
\expandafter\def\csname gv.C.idsplit-10.link.cj-gbdt.auprc\endcsname{0.716}
\expandafter\def\csname gv.C.idsplit-10.link.cj-gbdt.auprc|hop:1\endcsname{0.197}
\expandafter\def\csname gv.C.idsplit-10.link.cj-gbdt.auprc|hop:2\endcsname{0.514}
\expandafter\def\csname gv.C.idsplit-10.link.cj-gbdt.auprc|hop:3\endcsname{0.638}
\expandafter\def\csname gv.C.idsplit-10.link.cj-gbdt.auprc|hop:4\endcsname{0.468}
\expandafter\def\csname gv.C.idsplit-10.link.cj-gbdt.twins:auroc\endcsname{0.771}
\expandafter\def\csname gv.C.idsplit-10.link.nbfnet.auprc\endcsname{0.605}
\expandafter\def\csname gv.C.idsplit-10.link.nbfnet.auprc|hop:1\endcsname{0.423}
\expandafter\def\csname gv.C.idsplit-10.link.nbfnet.auprc|hop:2\endcsname{0.417}
\expandafter\def\csname gv.C.idsplit-10.link.nbfnet.auprc|hop:3\endcsname{0.478}
\expandafter\def\csname gv.C.idsplit-10.link.nbfnet.auprc|hop:4\endcsname{0.122}
\expandafter\def\csname gv.C.idsplit-10.link.nbfnet.twins:auroc\endcsname{0.653}
\expandafter\def\csname gv.C.idsplit-10.entity.rules-v2.auprc\endcsname{0.674}
\expandafter\def\csname gv.C.idsplit-10.entity.rules-v2.twins:auroc\endcsname{0.804}
\expandafter\def\csname gv.C.idsplit-10.entity.rules-generic.auprc\endcsname{0.480}
\expandafter\def\csname gv.C.idsplit-10.entity.rules-generic.twins:auroc\endcsname{0.731}
\expandafter\def\csname gv.C.idsplit-10.entity.cj-count.auprc\endcsname{0.310}
\expandafter\def\csname gv.C.idsplit-10.entity.cj-count.twins:auroc\endcsname{0.509}
\expandafter\def\csname gv.C.idsplit-10.entity.cj-gbdt-ent.auprc\endcsname{0.800}
\expandafter\def\csname gv.C.idsplit-10.entity.cj-gbdt-ent.twins:auroc\endcsname{0.807}
\expandafter\def\csname gv.C.idsplit-10.entity.rdl-ent.auprc\endcsname{0.735}
\expandafter\def\csname gv.C.idsplit-10.entity.rdl-ent.twins:auroc\endcsname{0.812}
\expandafter\def\csname gv.C.idsplit-20.link.rules-v2.auprc\endcsname{0.292}
\expandafter\def\csname gv.C.idsplit-20.link.rules-v2.auprc.ci\endcsname{[0.263, 0.322]}
\expandafter\def\csname gv.C.idsplit-20.link.rules-v2.auprc|hop:1\endcsname{0.417}
\expandafter\def\csname gv.C.idsplit-20.link.rules-v2.auprc|hop:2\endcsname{0.478}
\expandafter\def\csname gv.C.idsplit-20.link.rules-v2.auprc|hop:3\endcsname{0.090}
\expandafter\def\csname gv.C.idsplit-20.link.rules-v2.auprc|hop:3.ci\endcsname{[0.074, 0.113]}
\expandafter\def\csname gv.C.idsplit-20.link.rules-v2.auprc|hop:4\endcsname{0.000}
\expandafter\def\csname gv.C.idsplit-20.link.rules-v2.twins:auroc\endcsname{0.795}
\expandafter\def\csname gv.C.idsplit-20.link.rules-generic.auprc\endcsname{0.161}
\expandafter\def\csname gv.C.idsplit-20.link.rules-generic.auprc.ci\endcsname{[0.144, 0.179]}
\expandafter\def\csname gv.C.idsplit-20.link.rules-generic.auprc|hop:1\endcsname{0.113}
\expandafter\def\csname gv.C.idsplit-20.link.rules-generic.auprc|hop:2\endcsname{0.219}
\expandafter\def\csname gv.C.idsplit-20.link.rules-generic.auprc|hop:3\endcsname{0.027}
\expandafter\def\csname gv.C.idsplit-20.link.rules-generic.auprc|hop:3.ci\endcsname{[0.023, 0.031]}
\expandafter\def\csname gv.C.idsplit-20.link.rules-generic.auprc|hop:4\endcsname{0.000}
\expandafter\def\csname gv.C.idsplit-20.link.rules-generic.twins:auroc\endcsname{0.697}
\expandafter\def\csname gv.C.idsplit-20.link.cj-count.auprc\endcsname{0.219}
\expandafter\def\csname gv.C.idsplit-20.link.cj-count.auprc.ci\endcsname{[0.199, 0.240]}
\expandafter\def\csname gv.C.idsplit-20.link.cj-count.auprc|hop:1\endcsname{0.000}
\expandafter\def\csname gv.C.idsplit-20.link.cj-count.auprc|hop:2\endcsname{0.202}
\expandafter\def\csname gv.C.idsplit-20.link.cj-count.auprc|hop:3\endcsname{0.173}
\expandafter\def\csname gv.C.idsplit-20.link.cj-count.auprc|hop:3.ci\endcsname{[0.149, 0.197]}
\expandafter\def\csname gv.C.idsplit-20.link.cj-count.auprc|hop:4\endcsname{0.042}
\expandafter\def\csname gv.C.idsplit-20.link.cj-count.twins:auroc\endcsname{0.512}
\expandafter\def\csname gv.C.idsplit-20.link.cj-gbdt.auprc\endcsname{0.664}
\expandafter\def\csname gv.C.idsplit-20.link.cj-gbdt.auprc.ci\endcsname{[0.615, 0.710]}
\expandafter\def\csname gv.C.idsplit-20.link.cj-gbdt.auprc|hop:1\endcsname{0.214}
\expandafter\def\csname gv.C.idsplit-20.link.cj-gbdt.auprc|hop:2\endcsname{0.506}
\expandafter\def\csname gv.C.idsplit-20.link.cj-gbdt.auprc|hop:3\endcsname{0.608}
\expandafter\def\csname gv.C.idsplit-20.link.cj-gbdt.auprc|hop:3.ci\endcsname{[0.553, 0.665]}
\expandafter\def\csname gv.C.idsplit-20.link.cj-gbdt.auprc|hop:4\endcsname{0.333}
\expandafter\def\csname gv.C.idsplit-20.link.cj-gbdt.twins:auroc\endcsname{0.740}
\expandafter\def\csname gv.C.idsplit-20.link.nbfnet.auprc\endcsname{0.553}
\expandafter\def\csname gv.C.idsplit-20.link.nbfnet.auprc.ci\endcsname{[0.512, 0.596]}
\expandafter\def\csname gv.C.idsplit-20.link.nbfnet.auprc|hop:1\endcsname{0.439}
\expandafter\def\csname gv.C.idsplit-20.link.nbfnet.auprc|hop:2\endcsname{0.398}
\expandafter\def\csname gv.C.idsplit-20.link.nbfnet.auprc|hop:3\endcsname{0.441}
\expandafter\def\csname gv.C.idsplit-20.link.nbfnet.auprc|hop:3.ci\endcsname{[0.396, 0.488]}
\expandafter\def\csname gv.C.idsplit-20.link.nbfnet.auprc|hop:4\endcsname{0.083}
\expandafter\def\csname gv.C.idsplit-20.link.nbfnet.twins:auroc\endcsname{0.630}
\expandafter\def\csname gv.C.idsplit-20.entity.rules-v2.auprc\endcsname{0.684}
\expandafter\def\csname gv.C.idsplit-20.entity.rules-v2.auprc.ci\endcsname{[0.650, 0.715]}
\expandafter\def\csname gv.C.idsplit-20.entity.rules-v2.twins:auroc\endcsname{0.814}
\expandafter\def\csname gv.C.idsplit-20.entity.rules-generic.auprc\endcsname{0.494}
\expandafter\def\csname gv.C.idsplit-20.entity.rules-generic.auprc.ci\endcsname{[0.467, 0.523]}
\expandafter\def\csname gv.C.idsplit-20.entity.rules-generic.twins:auroc\endcsname{0.745}
\expandafter\def\csname gv.C.idsplit-20.entity.cj-count.auprc\endcsname{0.305}
\expandafter\def\csname gv.C.idsplit-20.entity.cj-count.auprc.ci\endcsname{[0.297, 0.314]}
\expandafter\def\csname gv.C.idsplit-20.entity.cj-count.twins:auroc\endcsname{0.510}
\expandafter\def\csname gv.C.idsplit-20.entity.cj-gbdt-ent.auprc\endcsname{0.778}
\expandafter\def\csname gv.C.idsplit-20.entity.cj-gbdt-ent.auprc.ci\endcsname{[0.744, 0.808]}
\expandafter\def\csname gv.C.idsplit-20.entity.cj-gbdt-ent.twins:auroc\endcsname{0.785}
\expandafter\def\csname gv.C.idsplit-20.entity.rdl-ent.auprc\endcsname{0.731}
\expandafter\def\csname gv.C.idsplit-20.entity.rdl-ent.auprc.ci\endcsname{[0.695, 0.766]}
\expandafter\def\csname gv.C.idsplit-20.entity.rdl-ent.twins:auroc\endcsname{0.810}
\expandafter\def\csname gv.C.idsplit-40.link.rules-v2.auprc\endcsname{0.201}
\expandafter\def\csname gv.C.idsplit-40.link.rules-v2.auprc|hop:1\endcsname{0.431}
\expandafter\def\csname gv.C.idsplit-40.link.rules-v2.auprc|hop:2\endcsname{0.228}
\expandafter\def\csname gv.C.idsplit-40.link.rules-v2.auprc|hop:3\endcsname{0.086}
\expandafter\def\csname gv.C.idsplit-40.link.rules-v2.auprc|hop:4\endcsname{0.000}
\expandafter\def\csname gv.C.idsplit-40.link.rules-v2.twins:auroc\endcsname{0.760}
\expandafter\def\csname gv.C.idsplit-40.link.rules-generic.auprc\endcsname{0.122}
\expandafter\def\csname gv.C.idsplit-40.link.rules-generic.auprc|hop:1\endcsname{0.160}
\expandafter\def\csname gv.C.idsplit-40.link.rules-generic.auprc|hop:2\endcsname{0.115}
\expandafter\def\csname gv.C.idsplit-40.link.rules-generic.auprc|hop:3\endcsname{0.032}
\expandafter\def\csname gv.C.idsplit-40.link.rules-generic.auprc|hop:4\endcsname{0.000}
\expandafter\def\csname gv.C.idsplit-40.link.rules-generic.twins:auroc\endcsname{0.708}
\expandafter\def\csname gv.C.idsplit-40.link.cj-count.auprc\endcsname{0.123}
\expandafter\def\csname gv.C.idsplit-40.link.cj-count.auprc|hop:1\endcsname{0.000}
\expandafter\def\csname gv.C.idsplit-40.link.cj-count.auprc|hop:2\endcsname{0.096}
\expandafter\def\csname gv.C.idsplit-40.link.cj-count.auprc|hop:3\endcsname{0.117}
\expandafter\def\csname gv.C.idsplit-40.link.cj-count.auprc|hop:4\endcsname{0.012}
\expandafter\def\csname gv.C.idsplit-40.link.cj-count.twins:auroc\endcsname{0.505}
\expandafter\def\csname gv.C.idsplit-40.link.cj-gbdt.auprc\endcsname{0.474}
\expandafter\def\csname gv.C.idsplit-40.link.cj-gbdt.auprc|hop:1\endcsname{0.245}
\expandafter\def\csname gv.C.idsplit-40.link.cj-gbdt.auprc|hop:2\endcsname{0.315}
\expandafter\def\csname gv.C.idsplit-40.link.cj-gbdt.auprc|hop:3\endcsname{0.453}
\expandafter\def\csname gv.C.idsplit-40.link.cj-gbdt.auprc|hop:4\endcsname{0.116}
\expandafter\def\csname gv.C.idsplit-40.link.cj-gbdt.twins:auroc\endcsname{0.645}
\expandafter\def\csname gv.C.idsplit-40.link.nbfnet.auprc\endcsname{0.399}
\expandafter\def\csname gv.C.idsplit-40.link.nbfnet.auprc|hop:1\endcsname{0.466}
\expandafter\def\csname gv.C.idsplit-40.link.nbfnet.auprc|hop:2\endcsname{0.221}
\expandafter\def\csname gv.C.idsplit-40.link.nbfnet.auprc|hop:3\endcsname{0.334}
\expandafter\def\csname gv.C.idsplit-40.link.nbfnet.auprc|hop:4\endcsname{0.026}
\expandafter\def\csname gv.C.idsplit-40.link.nbfnet.twins:auroc\endcsname{0.580}
\expandafter\def\csname gv.C.idsplit-40.entity.rules-v2.auprc\endcsname{0.626}
\expandafter\def\csname gv.C.idsplit-40.entity.rules-v2.twins:auroc\endcsname{0.785}
\expandafter\def\csname gv.C.idsplit-40.entity.rules-generic.auprc\endcsname{0.471}
\expandafter\def\csname gv.C.idsplit-40.entity.rules-generic.twins:auroc\endcsname{0.737}
\expandafter\def\csname gv.C.idsplit-40.entity.cj-count.auprc\endcsname{0.279}
\expandafter\def\csname gv.C.idsplit-40.entity.cj-count.twins:auroc\endcsname{0.504}
\expandafter\def\csname gv.C.idsplit-40.entity.cj-gbdt-ent.auprc\endcsname{0.683}
\expandafter\def\csname gv.C.idsplit-40.entity.cj-gbdt-ent.twins:auroc\endcsname{0.726}
\expandafter\def\csname gv.C.idsplit-40.entity.rdl-ent.auprc\endcsname{0.712}
\expandafter\def\csname gv.C.idsplit-40.entity.rdl-ent.twins:auroc\endcsname{0.811}
\expandafter\def\csname gv.C.recdrop-10.link.rules-v2.auprc\endcsname{0.347}
\expandafter\def\csname gv.C.recdrop-10.link.rules-v2.auprc|hop:1\endcsname{0.326}
\expandafter\def\csname gv.C.recdrop-10.link.rules-v2.auprc|hop:2\endcsname{0.650}
\expandafter\def\csname gv.C.recdrop-10.link.rules-v2.auprc|hop:3\endcsname{0.077}
\expandafter\def\csname gv.C.recdrop-10.link.rules-v2.auprc|hop:4\endcsname{0.000}
\expandafter\def\csname gv.C.recdrop-10.link.rules-v2.twins:auroc\endcsname{0.791}
\expandafter\def\csname gv.C.recdrop-10.link.rules-generic.auprc\endcsname{0.185}
\expandafter\def\csname gv.C.recdrop-10.link.rules-generic.auprc|hop:1\endcsname{0.084}
\expandafter\def\csname gv.C.recdrop-10.link.rules-generic.auprc|hop:2\endcsname{0.293}
\expandafter\def\csname gv.C.recdrop-10.link.rules-generic.auprc|hop:3\endcsname{0.025}
\expandafter\def\csname gv.C.recdrop-10.link.rules-generic.auprc|hop:4\endcsname{0.000}
\expandafter\def\csname gv.C.recdrop-10.link.rules-generic.twins:auroc\endcsname{0.664}
\expandafter\def\csname gv.C.recdrop-10.link.cj-count.auprc\endcsname{0.279}
\expandafter\def\csname gv.C.recdrop-10.link.cj-count.auprc|hop:1\endcsname{0.000}
\expandafter\def\csname gv.C.recdrop-10.link.cj-count.auprc|hop:2\endcsname{0.299}
\expandafter\def\csname gv.C.recdrop-10.link.cj-count.auprc|hop:3\endcsname{0.171}
\expandafter\def\csname gv.C.recdrop-10.link.cj-count.auprc|hop:4\endcsname{0.079}
\expandafter\def\csname gv.C.recdrop-10.link.cj-count.twins:auroc\endcsname{0.492}
\expandafter\def\csname gv.C.recdrop-10.link.cj-gbdt.auprc\endcsname{0.750}
\expandafter\def\csname gv.C.recdrop-10.link.cj-gbdt.auprc|hop:1\endcsname{0.175}
\expandafter\def\csname gv.C.recdrop-10.link.cj-gbdt.auprc|hop:2\endcsname{0.608}
\expandafter\def\csname gv.C.recdrop-10.link.cj-gbdt.auprc|hop:3\endcsname{0.577}
\expandafter\def\csname gv.C.recdrop-10.link.cj-gbdt.auprc|hop:4\endcsname{0.547}
\expandafter\def\csname gv.C.recdrop-10.link.cj-gbdt.twins:auroc\endcsname{0.795}
\expandafter\def\csname gv.C.recdrop-10.link.nbfnet.auprc\endcsname{0.610}
\expandafter\def\csname gv.C.recdrop-10.link.nbfnet.auprc|hop:1\endcsname{0.377}
\expandafter\def\csname gv.C.recdrop-10.link.nbfnet.auprc|hop:2\endcsname{0.490}
\expandafter\def\csname gv.C.recdrop-10.link.nbfnet.auprc|hop:3\endcsname{0.385}
\expandafter\def\csname gv.C.recdrop-10.link.nbfnet.auprc|hop:4\endcsname{0.129}
\expandafter\def\csname gv.C.recdrop-10.link.nbfnet.twins:auroc\endcsname{0.657}
\expandafter\def\csname gv.C.recdrop-10.entity.rules-v2.auprc\endcsname{0.688}
\expandafter\def\csname gv.C.recdrop-10.entity.rules-v2.twins:auroc\endcsname{0.812}
\expandafter\def\csname gv.C.recdrop-10.entity.rules-generic.auprc\endcsname{0.484}
\expandafter\def\csname gv.C.recdrop-10.entity.rules-generic.twins:auroc\endcsname{0.729}
\expandafter\def\csname gv.C.recdrop-10.entity.cj-count.auprc\endcsname{0.320}
\expandafter\def\csname gv.C.recdrop-10.entity.cj-count.twins:auroc\endcsname{0.493}
\expandafter\def\csname gv.C.recdrop-10.entity.cj-gbdt-ent.auprc\endcsname{0.821}
\expandafter\def\csname gv.C.recdrop-10.entity.cj-gbdt-ent.twins:auroc\endcsname{0.835}
\expandafter\def\csname gv.C.recdrop-10.entity.rdl-ent.auprc\endcsname{0.737}
\expandafter\def\csname gv.C.recdrop-10.entity.rdl-ent.twins:auroc\endcsname{0.811}
\expandafter\def\csname gv.C.recdrop-20.link.rules-v2.auprc\endcsname{0.344}
\expandafter\def\csname gv.C.recdrop-20.link.rules-v2.auprc|hop:1\endcsname{0.317}
\expandafter\def\csname gv.C.recdrop-20.link.rules-v2.auprc|hop:2\endcsname{0.646}
\expandafter\def\csname gv.C.recdrop-20.link.rules-v2.auprc|hop:3\endcsname{0.073}
\expandafter\def\csname gv.C.recdrop-20.link.rules-v2.auprc|hop:4\endcsname{0.000}
\expandafter\def\csname gv.C.recdrop-20.link.rules-v2.twins:auroc\endcsname{0.791}
\expandafter\def\csname gv.C.recdrop-20.link.rules-generic.auprc\endcsname{0.184}
\expandafter\def\csname gv.C.recdrop-20.link.rules-generic.auprc|hop:1\endcsname{0.084}
\expandafter\def\csname gv.C.recdrop-20.link.rules-generic.auprc|hop:2\endcsname{0.292}
\expandafter\def\csname gv.C.recdrop-20.link.rules-generic.auprc|hop:3\endcsname{0.024}
\expandafter\def\csname gv.C.recdrop-20.link.rules-generic.auprc|hop:4\endcsname{0.000}
\expandafter\def\csname gv.C.recdrop-20.link.rules-generic.twins:auroc\endcsname{0.663}
\expandafter\def\csname gv.C.recdrop-20.link.cj-count.auprc\endcsname{0.262}
\expandafter\def\csname gv.C.recdrop-20.link.cj-count.auprc|hop:1\endcsname{0.000}
\expandafter\def\csname gv.C.recdrop-20.link.cj-count.auprc|hop:2\endcsname{0.312}
\expandafter\def\csname gv.C.recdrop-20.link.cj-count.auprc|hop:3\endcsname{0.134}
\expandafter\def\csname gv.C.recdrop-20.link.cj-count.auprc|hop:4\endcsname{0.079}
\expandafter\def\csname gv.C.recdrop-20.link.cj-count.twins:auroc\endcsname{0.504}
\expandafter\def\csname gv.C.recdrop-20.link.cj-gbdt.auprc\endcsname{0.701}
\expandafter\def\csname gv.C.recdrop-20.link.cj-gbdt.auprc|hop:1\endcsname{0.185}
\expandafter\def\csname gv.C.recdrop-20.link.cj-gbdt.auprc|hop:2\endcsname{0.624}
\expandafter\def\csname gv.C.recdrop-20.link.cj-gbdt.auprc|hop:3\endcsname{0.480}
\expandafter\def\csname gv.C.recdrop-20.link.cj-gbdt.auprc|hop:4\endcsname{0.490}
\expandafter\def\csname gv.C.recdrop-20.link.cj-gbdt.twins:auroc\endcsname{0.770}
\expandafter\def\csname gv.C.recdrop-20.link.nbfnet.auprc\endcsname{0.556}
\expandafter\def\csname gv.C.recdrop-20.link.nbfnet.auprc|hop:1\endcsname{0.334}
\expandafter\def\csname gv.C.recdrop-20.link.nbfnet.auprc|hop:2\endcsname{0.466}
\expandafter\def\csname gv.C.recdrop-20.link.nbfnet.auprc|hop:3\endcsname{0.304}
\expandafter\def\csname gv.C.recdrop-20.link.nbfnet.auprc|hop:4\endcsname{0.117}
\expandafter\def\csname gv.C.recdrop-20.link.nbfnet.twins:auroc\endcsname{0.651}
\expandafter\def\csname gv.C.recdrop-20.entity.rules-v2.auprc\endcsname{0.685}
\expandafter\def\csname gv.C.recdrop-20.entity.rules-v2.twins:auroc\endcsname{0.812}
\expandafter\def\csname gv.C.recdrop-20.entity.rules-generic.auprc\endcsname{0.482}
\expandafter\def\csname gv.C.recdrop-20.entity.rules-generic.twins:auroc\endcsname{0.728}
\expandafter\def\csname gv.C.recdrop-20.entity.cj-count.auprc\endcsname{0.320}
\expandafter\def\csname gv.C.recdrop-20.entity.cj-count.twins:auroc\endcsname{0.504}
\expandafter\def\csname gv.C.recdrop-20.entity.cj-gbdt-ent.auprc\endcsname{0.808}
\expandafter\def\csname gv.C.recdrop-20.entity.cj-gbdt-ent.twins:auroc\endcsname{0.825}
\expandafter\def\csname gv.C.recdrop-20.entity.rdl-ent.auprc\endcsname{0.731}
\expandafter\def\csname gv.C.recdrop-20.entity.rdl-ent.twins:auroc\endcsname{0.813}
\expandafter\def\csname gv.C.recdrop-40.link.rules-v2.auprc\endcsname{0.338}
\expandafter\def\csname gv.C.recdrop-40.link.rules-v2.auprc|hop:1\endcsname{0.308}
\expandafter\def\csname gv.C.recdrop-40.link.rules-v2.auprc|hop:2\endcsname{0.633}
\expandafter\def\csname gv.C.recdrop-40.link.rules-v2.auprc|hop:3\endcsname{0.066}
\expandafter\def\csname gv.C.recdrop-40.link.rules-v2.auprc|hop:4\endcsname{0.000}
\expandafter\def\csname gv.C.recdrop-40.link.rules-v2.twins:auroc\endcsname{0.780}
\expandafter\def\csname gv.C.recdrop-40.link.rules-generic.auprc\endcsname{0.184}
\expandafter\def\csname gv.C.recdrop-40.link.rules-generic.auprc|hop:1\endcsname{0.084}
\expandafter\def\csname gv.C.recdrop-40.link.rules-generic.auprc|hop:2\endcsname{0.292}
\expandafter\def\csname gv.C.recdrop-40.link.rules-generic.auprc|hop:3\endcsname{0.024}
\expandafter\def\csname gv.C.recdrop-40.link.rules-generic.auprc|hop:4\endcsname{0.000}
\expandafter\def\csname gv.C.recdrop-40.link.rules-generic.twins:auroc\endcsname{0.655}
\expandafter\def\csname gv.C.recdrop-40.link.cj-count.auprc\endcsname{0.208}
\expandafter\def\csname gv.C.recdrop-40.link.cj-count.auprc|hop:1\endcsname{0.000}
\expandafter\def\csname gv.C.recdrop-40.link.cj-count.auprc|hop:2\endcsname{0.334}
\expandafter\def\csname gv.C.recdrop-40.link.cj-count.auprc|hop:3\endcsname{0.071}
\expandafter\def\csname gv.C.recdrop-40.link.cj-count.auprc|hop:4\endcsname{0.030}
\expandafter\def\csname gv.C.recdrop-40.link.cj-count.twins:auroc\endcsname{0.496}
\expandafter\def\csname gv.C.recdrop-40.link.cj-gbdt.auprc\endcsname{0.576}
\expandafter\def\csname gv.C.recdrop-40.link.cj-gbdt.auprc|hop:1\endcsname{0.186}
\expandafter\def\csname gv.C.recdrop-40.link.cj-gbdt.auprc|hop:2\endcsname{0.628}
\expandafter\def\csname gv.C.recdrop-40.link.cj-gbdt.auprc|hop:3\endcsname{0.340}
\expandafter\def\csname gv.C.recdrop-40.link.cj-gbdt.auprc|hop:4\endcsname{0.204}
\expandafter\def\csname gv.C.recdrop-40.link.cj-gbdt.twins:auroc\endcsname{0.688}
\expandafter\def\csname gv.C.recdrop-40.link.nbfnet.auprc\endcsname{0.447}
\expandafter\def\csname gv.C.recdrop-40.link.nbfnet.auprc|hop:1\endcsname{0.329}
\expandafter\def\csname gv.C.recdrop-40.link.nbfnet.auprc|hop:2\endcsname{0.456}
\expandafter\def\csname gv.C.recdrop-40.link.nbfnet.auprc|hop:3\endcsname{0.179}
\expandafter\def\csname gv.C.recdrop-40.link.nbfnet.auprc|hop:4\endcsname{0.042}
\expandafter\def\csname gv.C.recdrop-40.link.nbfnet.twins:auroc\endcsname{0.602}
\expandafter\def\csname gv.C.recdrop-40.entity.rules-v2.auprc\endcsname{0.655}
\expandafter\def\csname gv.C.recdrop-40.entity.rules-v2.twins:auroc\endcsname{0.799}
\expandafter\def\csname gv.C.recdrop-40.entity.rules-generic.auprc\endcsname{0.466}
\expandafter\def\csname gv.C.recdrop-40.entity.rules-generic.twins:auroc\endcsname{0.716}
\expandafter\def\csname gv.C.recdrop-40.entity.cj-count.auprc\endcsname{0.310}
\expandafter\def\csname gv.C.recdrop-40.entity.cj-count.twins:auroc\endcsname{0.496}
\expandafter\def\csname gv.C.recdrop-40.entity.cj-gbdt-ent.auprc\endcsname{0.736}
\expandafter\def\csname gv.C.recdrop-40.entity.cj-gbdt-ent.twins:auroc\endcsname{0.762}
\expandafter\def\csname gv.C.recdrop-40.entity.rdl-ent.auprc\endcsname{0.705}
\expandafter\def\csname gv.C.recdrop-40.entity.rdl-ent.twins:auroc\endcsname{0.801}
\expandafter\def\csname gv.C.repro.nbfnet.maxdiff\endcsname{17}
\expandafter\def\csname gv.C.repro.nbfnet.frozen\endcsname{0.673}
\expandafter\def\csname gv.C.repro.nbfnet.retrained\endcsname{0.688}
\expandafter\def\csname gv.C.repro.rdl-ent.maxdiff\endcsname{0}
\expandafter\def\csname gv.C.repro.rdl-ent.frozen\endcsname{0.752}
\expandafter\def\csname gv.C.repro.rdl-ent.retrained\endcsname{0.752}
\expandafter\def\csname gv.run.wall\endcsname{4,658}
\expandafter\def\csname gv.run.wallh\endcsname{1.3}
\expandafter\def\csname gv.run.freezev2\endcsname{12}
\expandafter\def\csname gv.run.freezeorig\endcsname{60}
\expandafter\def\csname gv.freeze.sha\endcsname{ef29f64b37209738}
\expandafter\def\csname gv.boot\endcsname{1000}

\newcommand{\paperTitle}{Payroll Fraud Detection in Synthetic HR/ERP Graphs: An Inductive Comparison of Rules, Path Features, and Graph Neural Networks}
\newcommand{\paperKeywords}{payroll fraud detection; ghost employees; occupational fraud; employee--vendor collusion; HR and ERP systems; internal audit; segregation of duties; inductive link prediction; knowledge graph completion; graph neural networks; gradient boosting; audit analytics; synthetic fraud benchmark; data leakage; preregistration}
\newcommand{\paperKeywordsPDF}{payroll fraud detection, ghost employees, occupational fraud, employee-vendor collusion, HR and ERP systems, internal audit, segregation of duties, inductive link prediction, knowledge graph completion, graph neural networks, GraIL, NBFNet, relational deep learning, gradient boosting, chained-join audit queries, base rate, record linkage, synthetic fraud benchmark, data leakage, preregistration}

\hypersetup{
  pdftitle={\paperTitle},
  pdfauthor={Xiaobo Chen, Genliang Zhu, Ziyuan Wang, Chu Wang},
  pdfsubject={Inductive payroll-fraud screening in synthetic HR/ERP graphs: rules, path features and graph neural networks},
  pdfkeywords={\paperKeywordsPDF}
}

\title{\paperTitle}
\author[1]{Xiaobo Chen}
\author[2,3]{Genliang Zhu}
\author[4]{Ziyuan Wang}
\author[2,5]{Chu Wang}
\affil[1]{University of Massachusetts Amherst}
\affil[2]{Accentrust}
\affil[3]{Georgia Institute of Technology}
\affil[4]{Duke University}
\affil[5]{University of Illinois Urbana-Champaign}
\date{October 2026}

\begin{document}
\maketitle

\begin{abstract}
\noindent
Payroll-fraud screening requires finding hidden employee--payee links while distinguishing suspicious relationships from legitimate ones.
We formulate this problem as inductive link prediction and entity ranking on synthetic human-resources and enterprise resource planning (HR/ERP) graphs, with unseen organizations and a matched benign twin for every planted case.
An original study on \GhNTestOrgs{} held-out organizations and a prospectively frozen extension compare analyst rules, explicit path features and graph neural networks (GNNs).
A tuned gradient-boosted model on chained-join features achieved the highest overall link average precision (AP; \GvV{A.link.cj-gbdt.auprc}), above the GraIL-style GNN (\GvV{A.link.grail-k3.auprc}); its 4-hop AP was \GvV{A4.ref} versus \GvV{A4.method}.
Thus, the observed multi-hop gain is not specific to GNNs.
On a new synthetic cohort with \GvV{B.prevpct}\% positive entities, chained-join boosting again led (AP \GvV{B.cj-gbdt-ent.auprc}); the relational entity GNN also exceeded record-aware rules (\GvV{B1.method} versus \GvV{B1.ref}).
At a 2\% review budget, their precision was \GvV{B2.method} versus \GvV{B2.ref}.
Under 40\% failed identifier joins, the entity GNN ranked above chained-join boosting (AP \GvV{C.idsplit-40.entity.rdl-ent.auprc} versus \GvV{C.idsplit-40.entity.cj-gbdt-ent.auprc}, descriptively).
Structure-only auditing exposed a shortcut in the initial generator; matched twins reduced the audited twin-pool AUROC to \GhAudFiveTwinMin{}--\GhAudFiveTwinMax{}.
These results separate multi-hop candidate discovery, record-based discrimination and review prioritization within an explicit synthetic evaluation design.
Code and data are available from the authors on request.
\end{abstract}

\noindent\textbf{Keywords:} \paperKeywords.

\section{Introduction}
\label{sec:intro}

Payroll assurance involves both relational discovery and assessment of the records supporting a payment. An employee and a vendor can share an account, a payee can use an account held by an employee's emergency contact, and an approver can have an undisclosed relationship with a new hire. The same graph shape can describe a legitimate joint account, a disclosed related-party transaction or an authorized family-member payee. Internal-control frameworks and fraud-risk assessment therefore consider declarations, independent approvals, activity records and verified account changes alongside relational matches~\citep{coso2013internal,iaasb2025isa240,wells2017corporate,vona2017fraudanalytics}.

These workflows pose three distinct questions. \emph{Candidate discovery} identifies which existing employee could be linked to a new employee or vendor. \emph{Record discrimination} distinguishes a suspicious candidate from a legitimate relationship of similar shape. \emph{Review prioritization} allocates a limited number of analyst checks across new entities. Identifying a four-hop chain does not establish that a payment is fraudulent; a direct-match rule can interpret a control record accurately while missing indirect candidates.

We study these questions as inductive link prediction and entity ranking on synthetic HR/ERP knowledge graphs. Every test organization and its entities are unseen in training. Every new entity--employee pair is scored without candidate filtering. The design includes seven planted patterns and a structurally matched benign twin for every planted case. Positive labels are controlled study assignments; the entity-level suite also includes a separately identified segregation-of-duties (SoD) control deficiency. This distinction does not equate a control exception with employee misconduct.

\paragraph{Why compare rules, path features and GNNs together?}
Inductive graph methods such as GraIL~\citep{teru2020grail} and NBFNet~\citep{zhu2021nbfnet} can score relations involving unseen entities. An analyst can also follow identifier chains with relational joins and use their outputs as explicit features. Both deserve comparison with record-aware rules. Prior graph-anomaly benchmarking has shown that tree ensembles with neighborhood aggregation can outperform GNNs~\citep{tang2023gadbench}. Here the question is how these approaches compare for organization-held-out payroll relations with matched benign topology, not whether boosting or graph learning is universally superior.

\paragraph{Main findings.}
The strongest overall link model in the extended comparison was a tuned GBDT on chained-join features: average precision (AP, reported as AUPRC in the result tables) was \GvV{A.link.cj-gbdt.auprc}, versus \GvV{A.link.grail-k3.auprc} for GraIL-style. At three and four hops it achieved \GvV{A3.ref} and \GvV{A4.ref}, versus \GvV{A3.method} and \GvV{A4.method}. These results establish the value of multi-hop features in the evaluated setting without attributing it exclusively to GNNs. Boosting also achieved twin-pool AUROC \GvV{A.link.cj-gbdt.twins:auroc}, above record-aware rules' \GvV{A.link.rules-v2.twins:auroc}.

On a new synthetic cohort with \GvV{B.prevpct}\% positive entities, boosting again led (AP \GvV{B.cj-gbdt-ent.auprc}). The RDL-style entity GNN exceeded record-aware rules (\GvV{B1.method} versus \GvV{B1.ref}), including precision at a 2\% per-organization review budget (\GvV{B2.method} versus \GvV{B2.ref}). Under 40\% failed joins, that GNN ranked above boosting (AP \GvV{C.idsplit-40.entity.rdl-ent.auprc} versus \GvV{C.idsplit-40.entity.cj-gbdt-ent.auprc}, descriptively). These are comparisons of implemented training procedures: boosting was validation-tuned, whereas GNN hyperparameters stayed fixed from development.

\paragraph{Topology and records as separate measurement targets.}
An attribute-free audit separated the initial generator's planted cases from hard negatives with AUROC \GhAudOneLinkSTwoMatchAUROC{}. Benign twins generated by the same structural procedure reduced audited structure-only twin-pool AUROC to \GhAudFiveTwinMin{}--\GhAudFiveTwinMax{}. This checks a particular shortcut channel and creates a separate record-discrimination view; it does not certify the absence of every possible shortcut.

\paragraph{Evidence stages.}
The original protocol was internally frozen before its \GhNTestOrgs{} test organizations were evaluated. The extension fixed new methods and tests before new computations; its comparison and noise arms reuse those organizations, while its lower-prevalence arm uses a new cohort from the same generator. \Cref{tab:study-lineage} distinguishes these roles. Original hypotheses, score-only diagnostics and extension results remain separately identifiable, with all outcomes retained.

\paragraph{Contributions.}
\begin{enumerate}[leftmargin=*,itemsep=2pt]
  \item \textbf{A relation-level audit task.} Organization-held-out payroll and employee--vendor link prediction and entity ranking, with evidence-hop and audit-test breakdowns (\cref{sec:problem}).
  \item \textbf{Matched controls and shortcut diagnostics.} Paired suspicious and benign structures, with attribute-free audits separating topology-based discovery from record discrimination (\cref{sec:generator,sec:leakage}).
  \item \textbf{An explicit multi-hop comparison.} Record-aware and generic rules, chained-join queries, tuned path-feature boosting and GNN-style models on shared pools (\cref{sec:methods,sec:addendum,sec:results}).
  \item \textbf{Review and robustness evidence.} A lower-prevalence cohort, fixed review budgets, failed joins and missing records identify where rankings persist or change (\cref{sec:addendum}).
\end{enumerate}

\paragraph{Data and use.}
All records are synthetic, use opaque identifiers and contain no real employee or customer data. The study evaluates behavior under the stated generator and protocols; operational use calls for population-specific validation and human review (\cref{sec:limitations}). Code, synthetic datasets and run records are available from the authors on request.

\section{Related Work}
\label{sec:related}

The closest prior work falls into five groups: payroll-fraud practice and audit analytics, inductive knowledge-graph completion, graph-based fraud and anomaly detection, machine learning on accounting and ERP data, and leakage in machine-learning evaluation. To our knowledge, based on an arXiv search and a Crossref search of journals and books (\cref{app:litsearch}), the search identified a distinct opportunity for an organization-held-out payroll benchmark combining relation-level labels, evidence-hop comparisons and structurally matched benign controls. The empirical contribution below is this combination and its measured consequences, rather than a claim to invent inductive link prediction or multi-hop auditing.

\paragraph{Payroll fraud and audit analytics.} Occupational-fraud surveys by the Association of Certified Fraud Examiners classify payroll schemes (ghost employees, falsified wages, commission schemes) as a distinct category of asset misappropriation~\citep{acfe2024rttn}, and practitioner handbooks describe the red flags we plant: a payee with no time records, a bank account or address shared with another employee, an employee who adds and pays the same record~\citep{wells2017corporate}. Fraud data analytics turns such red flags into scenario-based tests over core business systems~\citep{vona2017fraudanalytics}; digit analysis such as Benford's law screens transaction amounts~\citep{nigrini2012benford}; continuous auditing runs control tests automatically and repeatedly~\citep{alles2008continuous}; and process mining reconstructs and checks the actual approval paths of business processes~\citep{jans2013processmining}. Role-based access control and SoD rules in ERP systems are documented for practice~\citep{rahim2023oracle}. Many such tests are direct matches and threshold rules; our record-aware rule baseline (Rules v2, \cref{sec:methods}) represents this family in its one-edge form, although a practitioner's join query can also follow a short chain across tables. Recent machine-learning work on payroll fraud uses tabular classifiers: a scoping review of United States public-sector payroll systems finds the evidence in payroll specifically thinner than in banking, tax or audit settings~\citep{debrah2026payrollreview}; a preprint surveys supervised and unsupervised AI for payroll fraud in ERP systems~\citep{saha2025payrollerp}; and a study of home-health workforces cross-checks identity, scheduling, visit-verification and payroll data on a synthetic dataset and reports gradient boosting above rule thresholds, with network-style collusion indicators such as shared accounts and addresses~\citep{sultana2025hris}. That learned models beat rule thresholds on such direct indicators there, whereas our record-aware rules beat the GNNs on direct (2-hop) matches, is plausibly explained by the baselines: their rules are thresholds, ours are generator-aware conjunctions over control records. None of these works models multi-hop relations or unseen entities explicitly.

\paragraph{Inductive knowledge-graph completion and link prediction.} Classical link-prediction heuristics score a pair from its shared neighbourhood or its walks: common neighbours and the Adamic--Adar index~\citep{adamic2003friends,libennowell2007}, the Katz index~\citep{katz1953} and personalized PageRank~\citep{haveliwala2002ppr}. Relational GNNs such as R-GCN~\citep{schlichtkrull2018rgcn} learn node embeddings over typed relations, but node-wise embeddings cannot distinguish pairs whose endpoints are automorphic; labelling-trick methods such as SEAL~\citep{zhang2018seal} fix this by marking the two target nodes, and \citet{zhang2021labeling} give the theory. GraIL~\citep{teru2020grail} applies the labelling idea to inductive relation prediction: it extracts the enclosing subgraph between head and tail, labels nodes by their distances to both, and scores the pair with an R-GCN, so it never needs an embedding for an entity seen at training time. NBFNet~\citep{zhu2021nbfnet} generalizes path-based methods with a Bellman--Ford message-passing scheme conditioned on the query entity. RED-GNN~\citep{zhang2022redgnn} reasons over relational digraphs; NodePiece~\citep{galkin2021nodepiece} represents entities compositionally from anchors and relations; InGram~\citep{lee2023ingram} extends inductiveness to unseen relations; ULTRA~\citep{galkin2024ultra} aims at a single model that transfers across knowledge graphs. GLAR, proposed in One Subgraph for All~\citep{xie2024onesubgraph}, encodes a single opening subgraph around the query entity and reuses it to score all candidate tails, which reduces the per-pair extraction cost of GraIL-style methods. Inductive benchmarks~\citep{galkin2022ilpc} and re-evaluations of GNN link prediction~\citep{li2023heart} stress, as we do, that evaluation protocol and negative sampling can change conclusions. Our setting is the strictly inductive one these methods target: every test entity, every test link and every test organization is unseen in training. We implement GraIL-style and NBFNet-style models and a model \emph{inspired by} the opening-subgraph idea; none is a reproduction of the original code (\cref{sec:methods}). Inductive link prediction has been applied to banking fraud~\citep{bukhori2023inductive}: new transactions are linked into an existing homogeneous customer graph and scored as fraudulent or not. Our setting differs in three ways: induction is across organizations (no test entity or organization is seen in training), the prediction target is the colluding employee rather than a transaction label, and results are reported by evidence-chain length.

\paragraph{Graph-based fraud and anomaly detection.} Graph-fraud benchmarking includes extensive work on node classification, camouflage and class imbalance. CARE-GNN~\citep{dou2020caregnn} and PC-GNN~\citep{liu2021pcgnn} address camouflaged fraudsters and class imbalance; BWGNN~\citep{tang2022bwgnn} and the GADBench benchmark~\citep{tang2023gadbench} study spectral properties of anomalies and benchmark supervised graph anomaly detection; \citet{weber2019elliptic} and \citet{elmougy2023ellipticpp} released Bitcoin transaction graphs for anti-money-laundering; \citet{cheng2024gnnfraudreview} review more than 100 studies of GNNs for financial fraud. AMLworld~\citep{altman2023amlworld} uses simulation to make financial-crime ground truth explicit; here synthetic generation similarly supplies controlled labels. The closest design is TravelFraudBench~\citep{sajja2026travelfraudbench}, a configurable generator of fraud rings in heterogeneous travel-platform graphs that compares graph and non-graph methods. We differ in three ways: our task is \emph{link discovery} (which existing employee colludes with this new entity) rather than node classification; it is strictly inductive across organizations; and the planted patterns are mapped to named audit tests. Large language models have also been combined with graph fraud detectors~\citep{huang2025mled}; we do not evaluate an LLM arm.

\paragraph{Accounting, ERP and audit data.} Deep autoencoders have been used to flag anomalous journal entries~\citep{schreyer2017erp}, and self-supervised multi-view representations of accounting data have been learned for audit tasks~\citep{schreyer2021multiview}. \citet{tritscher2022openerp} generated open ERP system data with occupational-fraud scenarios designed with auditing experts; their data are process and transaction records rather than an entity--relation graph with link-level ground truth. Recent arXiv work proposes ERP test-data provisioning for financial control testing~\citep{samudrala2026seqfct}, a leakage-safe ensemble benchmark for financial risk in ERP systems~\citep{mishra2026erpriskbench}, and LLMs that flag irregularities in double-entry ledgers~\citep{kadir2025auditcopilot}. Relational deep learning~\citep{fey2023rdl} and RelBench~\citep{robinson2024relbench} treat an enterprise relational database as a heterogeneous graph of rows linked by foreign keys and train GNNs end to end; an HR/ERP system is such a database, so we include an RDL-style baseline at entity and link level (\cref{sec:methods}). Linking an emergency contact, an account holder and a vendor representative to the same person is a record-linkage problem~\citep{fellegi1969recordlinkage}, which our generator assumes solved (\cref{sec:limitations}).

\paragraph{Leakage, shortcuts, minimal pairs and prospective protocol.} \citet{kaufman2012leakage} formalized leakage in data mining, and \citet{kapoor2023leakage} documented how leakage undermines ML-based science. Shortcut learning~\citep{geirhos2020shortcut}---models exploiting a cue that predicts the label in the benchmark but not in the target task---is the failure mode we found in our first generator. Our matched benign twins are the graph analogue of contrast sets~\citep{gardner2020contrast}: minimally different inputs with different labels that test a model's local decision boundary. We follow the prospective protocol logic of \citet{nosek2018prereg}, use AUPRC as the primary metric because only \GhPairPrevalencePct\% of candidate pairs are positive~\citep{davis2006prroc,saito2015prc}, a cluster bootstrap over organizations~\citep{efron1979bootstrap,field2007clusterbootstrap}, and Holm's step-down correction~\citep{holm1979}.

\section{Problem Setting}
\label{sec:problem}

We study two inductive tasks on one HR/ERP knowledge graph per organization: score every (new entity, existing employee) pair for hidden collusion, and score every new entity for fraud. Results are broken down by the length of the planted evidence chain (hop).

We define an \emph{HR/ERP knowledge graph} of one organization as a typed, directed multigraph $G=(V,E,R)$ whose nodes are employees, vendors, other persons (emergency contacts, vendor representatives, account holders), bank accounts, addresses and phones, and whose relation types $R$ record pay accounts, account holders, addresses, phones, emergency contacts, vendor representatives, conflict-of-interest declarations, reporting lines (\texttt{reports\_to}) and hiring approvals (\texttt{hired\_by}). Payroll runs and access roles are exposed as node attributes (runs paid, roles held). The graph models see \GhNumRelations{} relation types (\GhNumEdgeTypes{} edge types and their inverses). Persons are already resolved: the same emergency contact, account holder or vendor representative is one node wherever it appears (\cref{sec:limitations}).

\paragraph{Unseen entities.} Each organization has an \emph{existing workforce} (employees and vendors present before the observation window) and a set of \emph{new} entities: new hires and new vendors whose own records and edges are observed but who appear in no training organization. We call a new entity $u$ and an existing employee $c$.

\paragraph{Tasks.} We define two tasks.
\begin{itemize}[leftmargin=*,itemsep=2pt]
  \item \textbf{Link level.} For every pair $(u,c)$ of a new entity and an existing employee in an organization, score whether $u$ and $c$ are connected by a hidden fraudulent relation (\texttt{colludes\_with} or \texttt{same\_beneficiary}). Every pair is scored; there is no candidate filtering. The test pool has \GhNPairs{} pairs, of which \GhNPairPos{} are positive (prevalence \GhPairPrevalencePct\%).
  \item \textbf{Entity level.} Score every new entity for whether it is fraudulent. The test pool has \GhNEntities{} entities, of which \GhNEntityPos{} are positive (\GhEntityPrevalencePct\%, a design parameter; \cref{sec:addendum} evaluates a lower-prevalence cohort). The \GhEntityPosHopZero{} SoD conflicts among them are control deficiencies rather than frauds; we keep them because the prospective protocol does, and report H3 without them as an exploratory check.
\end{itemize}

\paragraph{Hop.} We define the \emph{hop} of a planted fraud as the number of relations in its planted evidence chain between $u$ and the colluder $c$, excluding the direct manager or approver edge that every new hire has. Hop 0 is the entity-only SoD conflict; hop 1 is a ghost employee whose manager is also the hiring approver; hops 2--4 are chains through accounts, addresses, phones and persons (\cref{tab:patterns}). In auditors' terms, hop 2 is a \emph{direct match} ($u$ and $c$ share an account, address or phone node, two edges apart), and hops 3 and 4 are \emph{indirect} matches through one or two intermediate persons or identifiers. We call a rule \emph{one-step} if it tests one graph edge between identifiers (or the manager/approver edge); such rules can name $c$ directly only at hop $\le 2$ or through a manager/approver edge. In audit-query terms, some 3-hop chains (for example ``the account holder is $C$'s emergency contact'') are a single join across two tables, so chained join queries could reach them; the extension evaluates a chained-join query and boosting on path features (\cref{sec:addendum}).

\paragraph{Inductive split.} Organizations, not pairs, are split: \GhNDevOrgs{} development (dev) organizations and \GhNTestOrgs{} test organizations. Learned models train on \GhNTrainOrgs{} dev organizations and early-stop on the other \GhNValOrgs{}. Every test entity, every test link and every test organization is therefore unseen in training.

\paragraph{Metrics and pools.} The primary metric is link-level AUPRC (step-wise average precision with tied scores forming one threshold). We define the \emph{hop-$h$ pool} as all hop-$h$ positives together with all negatives; positives of other hops are excluded to retain a common negative reference for within-hop method comparisons. This is a diagnostic pool, not a full deployment pool. Its chance AUPRC is its prevalence, which is \GhLinkPrevPctHopOne\%, \GhLinkPrevPctHopTwo\%, \GhLinkPrevPctHopThree\% and \GhLinkPrevPctHopFour\% for $h=1,\dots,4$; because prevalence differs, AUPRC values should be compared across methods within a hop, not across hops. We also report AUROC (ties count half), recall at a fixed false-positive rate (thresholds between tie groups only), precision at $k$, and per-pattern and per-test cells. We define the \emph{twin pool} as the planted positives together with their structurally matched benign twins only (\cref{sec:generator}); its AUROC measures whether a method reads the control records rather than the topology. We define the \emph{hard-negative pool} of the exploratory phases as the planted positives together with the hand-built hard negatives of generator v1 (\cref{sec:leakage}).

\begin{table}[t]\centering\small
\caption{Each of the seven planted fraud patterns is caught by one of six audit tests, each named after the control it checks, and is planted at one, two or three different hop lengths ($C$ = colluding existing employee; EC = emergency contact). The two employee--vendor patterns share one test.}
\label{tab:patterns}
\footnotesize
\begin{tabular}{@{}>{\raggedright\arraybackslash}p{2.3cm}>{\raggedright\arraybackslash}p{2.5cm}>{\raggedright\arraybackslash}p{1.3cm}>{\raggedright\arraybackslash}p{1.05cm}>{\raggedright\arraybackslash}p{6.75cm}@{}}
\toprule
pattern & audit test (control checked) & entity & hops & planted evidence chain \\
\midrule
ghost employee & ghost-employee test (activity vs pay) & new hire & 1, 3 & paid in $\ge2$ runs with no activity, reports to $C$; h1: $C$ is also the hiring approver; h3: pay account held by $C$'s EC \\
shared payroll account & shared-account test (account uniqueness) & new hire & 2, 3, 4 & h2: same account as $C$; h3: account holder is $C$'s EC; h4: holder lives at $C$'s address \\
employee--vendor account & employee--vendor test (conflict-of-interest declaration) & new vendor & 2, 3, 4 & h2: vendor account equals $C$'s; h3: holder is $C$'s EC; h4: holder shares $C$'s phone \\
employee--vendor contact & employee--vendor test (conflict-of-interest declaration) & new vendor & 2, 3, 4 & h2: vendor address or phone equals $C$'s; h3: representative lives at $C$'s address; h4: representative shares a phone with $C$'s EC \\
approver link & approver--payee test (four-eyes approval) & new hire & 2, 3 & hired by $C$; h2: shares $C$'s address or account; h3: a person at the hire's address is $C$'s EC \\
account change before run & bank-change test (call-back verification) & hire or vendor & 2, 3 & bank details changed at most \GhRecentChangeDays{} days before a run; h2: to $C$'s account; h3: to an account held by $C$'s EC \\
SoD conflict & SoD test (maintain data $\times$ run payroll) & new hire & 0 & holds both \emph{maintain employee data} and \emph{run payroll} without an approved exception (entity level only; a control deficiency) \\
\bottomrule
\end{tabular}
\end{table}

\section{A Leakage-Controlled Synthetic HR/ERP Generator}
\label{sec:generator}

The confirmatory study uses generator v2, which plants seven fraud patterns at 0--4 hops and gives each a benign twin with identical topology. Its central design decision is that \emph{the label depends on control records, not on topology}: every planted fraud shape is built a second time as a benign twin with the same topology, and only the records attached to the entities say which of the two is fraud. Generator v1, used in the exploratory phases, lacked this property; \cref{sec:leakage} shows the consequence.

\subsection{Organizations and background structure}

Each synthetic organization has \GhExistingEmpMin--\GhExistingEmpMax{} existing employees and \GhExistingVendMin--\GhExistingVendMax{} existing vendors, \GhNRuns{} payroll runs \GhRunSpacing{} days apart, HR maintainers and payroll runners (one of whom holds both roles with an approved exception), and \GhNewHires{} new hires and \GhNewVendors{} new vendors that form the unseen query entities. Entities are opaque random identifiers; there are no names. Benign structure that resembles fraud is generated on purpose: couples who share an address, list each other as emergency contacts and sometimes hold a joint account; siblings who share an emergency contact; joint accounts held by a non-employee spouse; office main lines shared by many employees; multi-tenant commercial buildings shared by vendors; declared employee--vendor relationships; benign recent bank-detail changes; untracked attendance and leave. The v1 hand-built hard negatives (for example, a new hire who is the spouse of an existing employee) are kept in v2. Test organizations have \GhTestNodesMin--\GhTestNodesMax{} nodes and \GhTestEdgesMin--\GhTestEdgesMax{} edges.

\subsection{Planted fraud}

Each organization receives \GhPosPerOrg{} planted frauds (fraud multiplier \GhFraudMult{}), one per fraudulent entity, covering the seven patterns of \cref{tab:patterns}. Hop counts rotate across organizations so that every pattern--hop cell is populated; in the \GhNTestOrgs{} test organizations the link-level cells have at least \GhLinkMinCell{} positives each, and the hop totals are \GhLinkPosHopOne{} (h1), \GhLinkPosHopTwo{} (h2), \GhLinkPosHopThree{} (h3) and \GhLinkPosHopFour{} (h4); there are \GhEntityPosHopZero{} SoD conflicts at entity level (\cref{tab:design-counts} in the appendix). The ground truth stores, for every instance, the pattern, hop, entity, colluder, relation and the planted evidence edges; methods receive only the observed graph.

\subsection{Matched benign twins}

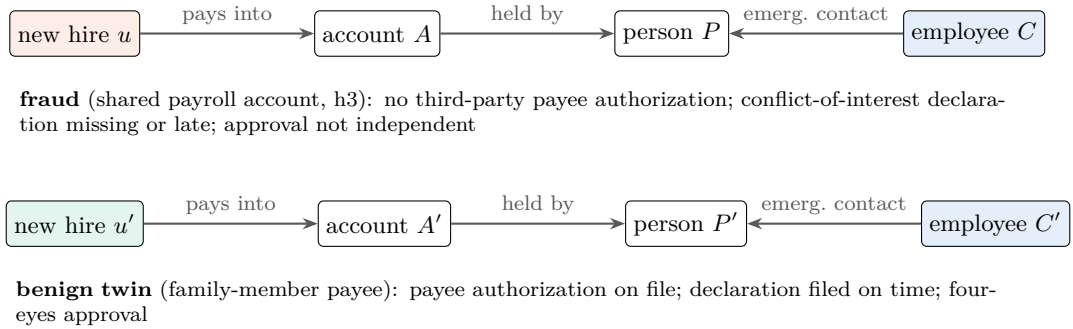
\begin{figure}[t]\centering
\begin{tikzpicture}[
  ent/.style={draw, rounded corners=2pt, minimum height=6mm, inner sep=3pt, font=\footnotesize, fill=white},
  rel/.style={-{Stealth[length=2mm]}, thick, gray!70!black},
  lab/.style={font=\scriptsize, midway, fill=white, inner sep=1pt},
  rec/.style={font=\scriptsize, align=left, text width=13cm}]
  % fraud
  \node[ent, fill=c2!12] (u1) {new hire $u$};
  \node[ent, right=2.3cm of u1] (a1) {account $A$};
  \node[ent, right=2.3cm of a1] (p1) {person $P$};
  \node[ent, right=2.3cm of p1, fill=c1!12] (c1n) {employee $C$};
  \draw[rel] (u1) -- node[lab, above=2pt] {pays into} (a1);
  \draw[rel] (a1) -- node[lab, above=2pt] {held by} (p1);
  \draw[rel] (c1n) -- node[lab, above=2pt] {emerg.\ contact} (p1);
  \node[rec, below=3mm of u1.south west, anchor=north west] (r1) {\textbf{fraud} (shared payroll account, h3): no third-party payee authorization; conflict-of-interest declaration missing or late; approval not independent};
  % twin
  \node[ent, fill=c3!12, below=1.9cm of u1] (u2) {new hire $u'$};
  \node[ent, right=2.3cm of u2] (a2) {account $A'$};
  \node[ent, right=2.3cm of a2] (p2) {person $P'$};
  \node[ent, right=2.3cm of p2, fill=c1!12] (c2n) {employee $C'$};
  \draw[rel] (u2) -- node[lab, above=2pt] {pays into} (a2);
  \draw[rel] (a2) -- node[lab, above=2pt] {held by} (p2);
  \draw[rel] (c2n) -- node[lab, above=2pt] {emerg.\ contact} (p2);
  \node[rec, below=3mm of u2.south west, anchor=north west] (r2) {\textbf{benign twin} (family-member payee): payee authorization on file; declaration filed on time; four-eyes approval};
\end{tikzpicture}
\caption{A planted 3-hop fraud and its matched benign twin have identical topology; only the control records attached to the entities differ. A model that separates the two must read the records.}
\label{fig:twin}
\end{figure}

For every planted fraud the generator calls the \emph{same} structural code a second time with \texttt{fraud=false} on another new entity and another existing employee (look-alike ratio \GhLookalikeRatio). The twin reuses the fraud's variant, so it has the same evidence edge types, the same hop chain, the same intermediary types and the same degree side-effects---for example, if the fraud gives the colluder an extra emergency contact, so does the twin. Each twin has a benign story: relatives with a declared joint account, a family-member payee who is the related employee's emergency contact, a declared related vendor, a referral hire approved by a related employee, a verified change to a family account shortly before a run, a documented SoD exception. \Cref{fig:twin} shows one pair.

\subsection{Control records and the label}

Each planted shape has two or three evidence \emph{channels}, stored as observed employee or vendor attributes: conflict-of-interest declaration on file and its timing (a declaration filed long after the start date counts as late); third-party payee or joint-account authorization; call-back verification of a bank-detail change; four-eyes (independent) approval; badge and system-login activity; time-sheets; and the SoD exception record. For a fraud, each channel is suspicious with probability \GhPSuspFraud, with at least one suspicious channel. We define a \emph{camouflaged} fraud as one with exactly one suspicious channel and legitimate-looking records otherwise; a fraction \GhCamouflage{} of frauds is camouflaged. For a twin, each channel is suspicious with probability \GhPSuspBenign{} (missing paperwork, late filing, legitimately low activity). Background entities carry the same attributes at background rates (for example, a conflict-of-interest declaration rate of \GhBgDisclosure). Every record of every employee or vendor is independently missing with probability \GhMissingRate. No label noise is applied. All rates are modelling choices made to keep the record signal noisy but learnable; none is estimated from real cases.

The record distributions overlap by design: a camouflaged case and a benign twin can carry the same observed controls. The twin comparison therefore tests discrimination from noisy records, separately from whether a connecting chain exists.

\subsection{Data, seeds and the superseded export}

The confirmatory dataset (\texttt{hrgraph-004}) has \GhNOrgs{} organizations---\GhNDevOrgs{} dev and \GhNTestOrgs{} test---generated from base seed \GhBaseSeed{}. A manifest records the SHA-256 of every organization file and of the generator sources; the confirmatory run re-checked all \GhManifestFiles{} data files and \GhGenSources{} generator sources before training (\GhManifestMismatch{} mismatches). Two data-handling deviations are disclosed in \cref{sec:deviations}: a seed collision that ruled out the first suggested seed, and a superseded first v2 export (\texttt{hrgraph-003}).

\section{Leakage Audit: Near-Perfect Scores, Structural Tells and Generator v2}
\label{sec:leakage}

An attribute-free audit identified a structural shortcut in the initial generator; matched twins provide a control for that channel. Before the confirmatory study, two exploratory phases on generator v1 produced GNN results that looked like a strong positive finding. A leakage audit showed they were mostly a property of the generator. This section reports that sequence because it changed the study design and because the audit is reusable. All numbers in this section are exploratory and come from dev organizations or from v1 data; none enters the confirmatory test.

\subsection{Exploratory results and the audit trigger}

In phase 1 (generator v1, \GhPOneNPairs{} test pairs, \GhPOneNPairPos{} positive) the tuned rules had the best link AUPRC (\GhPOneRulesTunedAUPRC, against \GhPOnePathLRAUPRC{} for path features + LR), but they recovered only \GhPOneRulesTunedRFprOneHopThree{} of 3-hop positives at 1\% FPR, against \GhPOnePathLRRFprOneHopThree{} for the path model; no phase-1 method solved 4 hops at low FPR. Phase 2 scaled the corpus to \GhPTwoNTestOrgs{} test organizations (\GhPTwoNPairs{} pairs, \GhPTwoNPairPos{} positive) and added the GNNs. The result (\cref{tab:phase2}) looked decisive: NBFNet-style link AUPRC \GhPTwoNBFNetAUPRC{} and GraIL-style $k=3$ \GhPTwoGraILThreeAUPRC, against \GhPTwoRulesTunedAUPRC{} for the tuned rules, with 4-hop recall at 1\% FPR of \GhPTwoNBFNetRFprOneHopFour{} and \GhPTwoGraILThreeRFprOneHopFour{} where the rules had \GhPTwoRulesTunedRFprOneHopFour. At entity level the R-GCN reached AUPRC \GhPTwoEntRGCNEntAUPRC.

% generated by xiaobo-experiment/render-results-draft.py; do not edit
\begin{table}[t]\centering\footnotesize\setlength{\tabcolsep}{3.5pt}
\caption{Exploratory phase 2 on generator v1 (24 test orgs, 215{,}545 pairs): near-ceiling GNN scores that the leakage audit later traced to structural tells. Not confirmatory; reported to document the trap.}
\label{tab:phase2}
\begin{adjustbox}{max width=\linewidth}
\begin{tabular}{lrlrrrr}
\toprule
method & AUROC & AUPRC & AUPRC h3 & AUPRC h4 & R@1\%FPR h3 & R@1\%FPR h4 \\
\midrule
Rules tuned & 0.826 & 0.502 & 0.201 & 0.000 & 0.200 & 0.000 \\
Personalized PageRank & 0.991 & 0.080 & 0.033 & 0.005 & 0.742 & 0.062 \\
Path features + LR & 0.984 & 0.312 & 0.271 & 0.002 & 1.000 & 0.021 \\
RDL-style (link)$^\dagger$ & 0.892 & 0.065$\pm$0.073 & 0.037 & 0.023 & 0.439 & 0.194 \\
GraIL-style $k{=}2$$^\dagger$ & 0.983 & 0.805$\pm$0.015 & 0.975 & 0.003 & 1.000 & 0.194 \\
GraIL-style $k{=}3$ & 0.999 & 0.973$\pm$0.010 & 0.974 & 0.928 & 1.000 & 1.000 \\
NBFNet-style$^\dagger$ & 1.000 & 0.997$\pm$0.002 & 0.993 & 0.902 & 1.000 & 1.000 \\
Opening-subgraph (OSFA-insp.)$^\dagger$ & 0.998 & 0.956$\pm$0.013 & 0.928 & 0.949 & 0.981 & 1.000 \\
\bottomrule
\end{tabular}
\end{adjustbox}
\end{table}

The high scores motivated an attribute-free audit: the initial generator planted cases through structural code paths that benign entities did not share. Phase 1 had identified one candidate cue, an \emph{extra} emergency contact of the colluder. We tested those structural signals before the original prospective protocol was finalized.

\subsection{The audit: structure-only baselines}

The audit (\texttt{leakage-audit.mjs}, dev organizations only) asks how far one gets \emph{without any node or edge attributes}---topology and relation types only. It fits four structure-only baselines with leave-one-organization-out logistic regression:
\begin{itemize}[leftmargin=*,itemsep=1pt]
  \item \textbf{A0}: an unsupervised ``a path of length $\le4$ exists'' score;
  \item \textbf{A1}: a bag of typed simple paths between $u$ and $c$ (length $\le4$);
  \item \textbf{A2}: relation-type degree profiles of $u$, of $c$ and of $u$'s neighbours, with \emph{no} $u$--$c$ path feature;
  \item \textbf{A3}: A1 and A2 together.
\end{itemize}
It evaluates them on four pools: all pairs; connected pairs; the hard-negative pool (positives against the hand-built hard negatives); and, for v2, the twin pool.

\paragraph{Finding 1: topology alone solved v1.} On v1 dev data (\GhAudOneNLinkPairs{} pairs, \GhAudOneNLinkPos{} positive), A3 reached link AUROC \GhAudOneLinkSThreeAllAUROC, AUPRC \GhAudOneLinkSThreeAllAUPRC{} and recall \GhAudOneLinkSThreeAllRFprOne{} at 1\% FPR; A1 alone reached AUPRC \GhAudOneLinkSOneAllAUPRC. For comparison, the tuned rules had test AUPRC \GhPTwoRulesTunedAUPRC{} in phase 2.

\paragraph{Finding 2: degree profiles alone separated fraud from the hard negatives.} On the hard-negative pool, A2---which sees no $u$--$c$ path at all---reached AUROC \GhAudOneLinkSTwoMatchAUROC. The tells came from v1's construction. A colluder gained an extra emergency contact (single-feature AUROC of the colluder's emergency-contact out-degree: \GhAudOneTellECAUROC{} over all pairs). Hard negatives carried the opposite tells: the mutual-emergency-contact path occurred in \GhAudOneMutualECPos{} of \GhAudOneMatchedPos{} positives and in \GhAudOneMutualECNeg{} of \GhAudOneMatchedNeg{} hard negatives. At entity level A2 alone reached AUROC \GhAudOneEntSTwoAllAUROC{} and AUPRC \GhAudOneEntSTwoAllAUPRC.

\paragraph{Conclusion of the audit.} In v1, ``this multi-hop structure exists around this pair'' was close to equivalent to ``this pair is fraud''. The near-ceiling phase-2 GNN scores therefore did not show multi-hop \emph{reasoning about fraud}; they showed that the models could detect the generator's planting code.

\subsection{Generator v2 and the twin check}

Generator v2 (\cref{sec:generator}) removed the tells by construction: every fraud has a twin produced by the same structural code, and the label lives in the control records. We re-ran the same audit on v2 dev data (\cref{tab:leakage}).

% generated by xiaobo-experiment/render-results-draft.py; do not edit
\begin{table}[t]\centering\footnotesize\setlength{\tabcolsep}{3.5pt}
\caption{Leakage audit on the 12 dev organizations: with generator v1, topology alone (no attributes) separated fraud from the hand-built hard negatives (hard-negative-pool AUROC up to 0.990); with generator v2 every structure-only baseline is at chance on the twin pool. Upper block: link level; lower block: entity level. A0--A3 are the attribute-free audit baselines; ATTR (attributes only) is a non-structural reference.}
\label{tab:leakage}
\begin{adjustbox}{max width=\linewidth}
\begin{tabular}{lrrrrrrr}
\toprule
baseline & \makecell[l]{all pairs\\ {}AUPRC v1} & v2 & \makecell[l]{all pairs\\ {}recall@1\% v1} & v2 & \makecell[l]{hard-neg.\\ {}AUROC v1} & v2 & \makecell[l]{twin pool\\ {}AUROC v2} \\
\midrule
link: A0 path exists & 0.018 & 0.020 & 0.007 & 0.007 & 0.146 & 0.384 & 0.509 \\
link: A1 path bag & 0.598 & 0.374 & 0.944 & 0.976 & 0.788 & 0.599 & 0.506 \\
link: A2 degree profile & 0.009 & 0.004 & 0.076 & 0.031 & 0.990 & 0.666 & 0.494 \\
link: A3 = A1 + A2 & 0.685 & 0.389 & 1.000 & 0.997 & 0.935 & 0.644 & 0.504 \\
\midrule
entity: A0 (max over $c$) & 0.299 & 0.302 & 0.000 & 0.000 & 0.425 & 0.474 & 0.513 \\
entity: A1 (max over $c$) & 0.634 & 0.410 & 0.167 & 0.027 & 0.767 & 0.619 & 0.514 \\
entity: A2 degree profile & 0.838 & 0.454 & 0.673 & 0.033 & 0.920 & 0.685 & 0.493 \\
entity: A3 (max over $c$) & 0.760 & 0.427 & 0.196 & 0.018 & 0.876 & 0.654 & 0.508 \\
entity: ATTR (attributes only) & 0.805 & 0.681 & 0.571 & 0.170 & 0.880 & 0.845 & 0.792 \\
\bottomrule
\end{tabular}
\end{adjustbox}
\end{table}

\paragraph{Finding 3: structure-only baselines are at chance on the twin pool.} Every structure-only baseline has twin-pool AUROC between \GhAudFiveTwinMin{} and \GhAudFiveTwinMax{} at both levels, and no single structural feature in the univariate screen deviates from 0.5 by more than \GhAudFiveTwinScreenMaxDev{} on the twin pool. The attributes-only reference (ATTR) reaches entity twin-pool AUROC \GhAudFiveEntATTRTwinAUROC: the signal is in the records, as intended.

\paragraph{Finding 4: structural candidate generation and the twin-pool diagnostic.} The generator connects each planted positive pair, and only a small fraction of all pairs is. On all pairs, A3 still reaches link AUROC \GhAudFiveLinkSThreeAllAUROC{} and recall \GhAudFiveLinkSThreeAllRFprOne{} at 1\% FPR on v2, but its AUPRC drops to \GhAudFiveLinkSThreeAllAUPRC{} (v1: \GhAudOneLinkSThreeAllAUPRC): the ranking puts each fraud next to its twin. Per hop, A3 reaches dev AUPRC \GhAudFiveLinkSThreeAllAUPRCHopTwo, \GhAudFiveLinkSThreeAllAUPRCHopThree{} and \GhAudFiveLinkSThreeAllAUPRCHopFour{} at hops 2, 3 and 4 (A1: \GhAudFiveLinkSOneAllAUPRCHopTwo, \GhAudFiveLinkSOneAllAUPRCHopThree, \GhAudFiveLinkSOneAllAUPRCHopFour). These are dev values from leave-one-organization-out fits, not comparable one-to-one with the test values of \cref{sec:results}, and the audit baselines were not run on the test split. With one twin per fraud, an idealized structure-only detector that ranks each hop-$h$ fraud and its twin together above every other pair, and cannot order the two, has hop-$h$ AUPRC of about 0.5; this is the ceiling that reach alone can attain on our pools. The hard-negative pool stays above chance on v2 (A2 AUROC \GhAudFiveLinkSTwoMatchAUROC) only because the v1 hard negatives keep their own benign structure, which points \emph{away} from fraud.

\paragraph{Consequence for the protocol.} Two decisions in the prospective protocol follow directly. First, link recall at 1\% FPR is saturated by structure alone on v2 and was declared non-informative before the test. Second, the twin-pool AUROC became a confirmatory hypothesis (H2): it is the only link-level view in which a model cannot win by finding the candidate, and must read the records.

\subsection{A checklist for synthetic graph fraud benchmarks}

We propose four checks that would have caught our v1 problem before any model was trained:
\begin{enumerate}[leftmargin=*,itemsep=1pt]
  \item \textbf{Structure-only ceiling.} Fit attribute-free baselines (path bags, degree profiles); if they approach the best model, the benchmark measures structure, not the target concept.
  \item \textbf{Matched twins.} Generate each positive's benign twin with the same code path and report the twin-pool AUROC of every method; structure-only baselines must be at chance there.
  \item \textbf{Univariate screens.} List the single features with the most extreme AUROC on each pool; planting side-effects show up as degree features.
  \item \textbf{Saturation audit of metrics.} Identify metrics that structure-only baselines already saturate and declare them non-informative before the test.
\end{enumerate}

\section{Methods, Study Stages and Analysis Protocols}
\label{sec:methods}

The original study compares \GhNLinkMethods{} link-level and \GhNEntityMethods{} entity-level scorers: four rule sets, four graph heuristics, a path-feature model, seven GNN configurations from five families, and rule--GNN combinations. The extension adds generic rules, a chained-join query and tuned path-feature boosting. Selection uses development organizations; original hyperparameters and the extension grid were fixed before their respective formal computations. \Cref{tab:study-lineage} distinguishes the evidence stages. The original five-test and extension nine-test families retain their recorded organization-bootstrap convention and Holm adjustments.

\begin{table}[htbp]\centering\small
\caption{Study lineage. Prospective freezes are internal protocol records, not public registry registration. Reused test organizations remain identified; new seeds from the same generator are not an independent external data source.}
\label{tab:study-lineage}
\begin{tabular}{@{}>{\raggedright\arraybackslash}p{2cm}>{\raggedright\arraybackslash}p{4cm}>{\raggedright\arraybackslash}p{4.2cm}>{\raggedright\arraybackslash}p{4.2cm}@{}}
\toprule
stage & data and prior visibility & role & selection and inference \\
\midrule
Original & hrgraph-004; no prior test output & \GhNTestOrgs{} held-out organizations & Dev-fixed models; five recorded tests \\
Diagnostics & Original written scores and truth & Seed, twin, prevalence and SoD analyses & Exploratory, no training \\
Extension A & hrgraph-004; original results visible & Multi-hop/non-GNN and generic-rule comparisons & Dev-selected GBDT; reused test \\
Extension B & New hrgraph-005; no prior cohort output & \GvV{B.prevpct}\% positive-rate ranking and review budget & New cohort, same generator; retraining \\
Extension C & hrgraph-004; original results visible & Failed joins and missing records & Clean-dev training; reused test organizations \\
\bottomrule
\end{tabular}
\end{table}

\subsection{Analyst rules}

\paragraph{Rules naive and rules tuned.} Eight direct, one-step rules: shared account, account-holder mismatch, shared address, shared phone, recent bank-detail change, SoD conflict, paid without attendance, and a match with the hiring approver. The tuned variant adds the exemptions an analyst would apply: declared household (mutual emergency contact), declared conflict of interest, office main line, approved SoD exception, untracked attendance or leave, and an account holder who is the entity's own contact. The entity score is the number of flags; the link score counts direct identifier matches between $u$ and $c$, plus manager-and-no-attendance and approver-and-approver-link, plus a small multiple of the entity flag count.

\paragraph{Rules severity.} The tuned rules with hand-set severity weights (beneficiary and control flags weigh more than contact overlaps, which weigh more than a recent change). These weights were added after a phase-1 debug run showed heavy ties among flag counts; they are not fitted, but they are not blind either (\cref{sec:deviations}).

\paragraph{Rules v2 (record-aware).} The primary baseline. On top of the tuned rules it reads the v2 control records: an identifier match is exempt if a conflict-of-interest declaration was filed on time and the approval was not single-handed; a holder mismatch is exempt if a payee authorization is on file; a recent change is exempt if it was call-back verified; ``paid without activity'' also fires on low badge \emph{and} login activity; and a new flag fires when approval was not independent. The weights are hand-set. Rules v2 is a generator-informed reference whose exemptions use the control-record schema (\cref{sec:deviations}); its performance is not a bound on other analysts. It remains one-step: it can name a colluder only through a direct identifier match or a manager/approver edge.

\subsection{Graph heuristics and path features}

Common neighbours, Adamic--Adar, truncated Katz and personalized PageRank (PPR) are computed on an undirected linkage view of the graph; entity scores are the maximum over candidates. \emph{Path features + LR} counts typed simple paths of length $\le3$ between $u$ and $c$ (edge-type and direction sequences), combined with the entity's attribute flags (including the v2 records) and attribute $\times$ path-type crosses, in an L2-regularized logistic regression fitted on all \GhNDevOrgs{} dev organizations; its entity model is a second logistic regression on the attributes and the maximum and count of positive link logits.

\subsection{Graph neural networks}

All GNNs use the same \GhFeatureDim-dimensional node features: type one-hot, log degrees (total and per relation), the phase-1 attribute flags, run and activity counts, account-change and role indicators, and \emph{the v2 control records}. No feature is derived from the label (unit-tested). Hyperparameters were fixed in phase 2, before generator v2 existed, and were not tuned (\cref{tab:hparams}).

\begin{itemize}[leftmargin=*,itemsep=2pt]
  \item \textbf{R-GCN (entity)}: a two-layer relational GCN~\citep{schlichtkrull2018rgcn} entity classifier with basis decomposition.
  \item \textbf{RDL-style (entity / link)}: a relational-deep-learning-style baseline~\citep{fey2023rdl,robinson2024relbench}. The graph is treated as relational tables (node tables and foreign-key tables); per-table input encoders are followed by two layers of heterogeneous message passing with GraphSAGE-style~\citep{hamilton2017graphsage} convolutions. Embeddings are node-wise, with no labelling trick. The entity head is an MLP on $h_u$; the link head is an MLP on $[h_u, h_c, h_u\odot h_c]$.
  \item \textbf{GraIL-style ($k=2$, $k=3$)}: for each candidate pair, the enclosing subgraph of radius $k$ is the intersection of the two radius-$k$ neighbourhoods, with distances recomputed inside the subgraph and nodes farther than $k$ pruned~\citep{teru2020grail}. Nodes carry double-radius distance labels \emph{concatenated with the node features} (original GraIL uses labels only), and a three-layer R-GCN with a [mean-pool, $h_u$, $h_c$] readout scores the pair. On dev, enclosing subgraphs had \GhSubgraphMeanKTwo{} ($k=2$) and \GhSubgraphMeanKThree{} ($k=3$) nodes on average. A 4-hop chain fits inside a $k$-subgraph only when $h\le k+1$, so $k=2$ cannot see h4 by construction.
  \item \textbf{NBFNet-style}: six layers of generalized Bellman--Ford message passing from a boundary condition at the query entity, with relation-conditioned (DistMult) messages and a [sum, min] aggregator~\citep{zhu2021nbfnet}; observed node features are injected in every update (a deviation from the original); candidates are scored from $[h_c, x_c]$.
  \item \textbf{Opening-subgraph (OSFA-insp., inspired by One Subgraph for All)}: inspired only by the abstract-level idea of GLAR~\citep{xie2024onesubgraph}: one shared opening subgraph, the radius-4 ball around the query entity, is encoded once by a four-layer R-GCN with distance-from-query labels and reused to score all candidates; candidates outside the ball get a learned embedding. Nothing else from that paper is implemented; it is not a reproduction.
\end{itemize}
Entity scores from link models are the maximum over all candidates.

\paragraph{Training.} GNNs train on the first \GhNTrainOrgs{} dev organizations with class-balanced binary cross-entropy and Adam, and keep the epoch with the best validation AUPRC on the remaining \GhNValOrgs. Each GNN is trained with \GhNSeeds{} seeds (\GhSeeds); we define the \emph{seed-mean metric} as the metric computed separately on each seed's scores and then averaged over the seeds, and all reported GNN values are seed-mean metrics, with the seed standard deviation (SD) reported separately. Training ran on CPU only.

\paragraph{Combinations.} Rules v2 is combined with each GNN in two ways, fitted only on the \GhNValOrgs{} validation organizations (the same ones used for early stopping, so the combination weights are not independent of model selection): \texttt{lr} is a two-feature class-balanced logistic regression on the two scores; \texttt{max} takes the maximum of the two methods' validation-negative-calibrated surprisals. The \texttt{max} combination saturates at the validation pool size and produces ties at the top; it was kept unchanged from phase 2 and is reported only in the appendix tables.

\subsection{Original prospective protocol and statistics}

The original prospective protocol was internally frozen on \GhFreezeDate{} as \texttt{freeze-phase3-001}: the prospective protocol text, all code, the data manifest and the design document, \GhFreezeFiles{} files with SHA-256 checksums (checksum file digest \texttt{\GhFreezeShaShort\ldots}). No file of any test organization had been opened by an analysis script at that point. The frozen run script verified every checksum and every data file before training, ran each model once, and wrote to a create-only directory; the report script re-verified the freeze (\GhFreezeVerifiedAtReport{} files) before reading any score.

\paragraph{Pre-specified models.} Chosen on the dev rehearsal before the freeze: GraIL-style $k=3$ is the primary link model (highest dev link AUPRC, \GhDevLinkGraILThreeAUPRC{}) and RDL-style (entity) the primary entity model (highest single-model dev entity AUPRC and twin AUROC); Rules v2 is the primary baseline. R-GCN (entity), NBFNet-style and the opening-subgraph model were run under the same protocol but are not pre-specified; their results are descriptive.

\paragraph{Hypotheses.} One family of five one-sided tests:
\begin{itemize}[leftmargin=*,itemsep=2pt]
  \item \textbf{H1-h2, H1-h3, H1-h4 (multi-hop reach).} GraIL-style $k=3$ minus Rules v2 link AUPRC $>0$, separately for hop-$h$ positives against all negatives, $h=2,3,4$.
  \item \textbf{H2 (record reading; leakage-free).} GraIL-style $k=3$ minus Rules v2 link AUROC $>0$ on the twin pool.
  \item \textbf{H3 (entity).} RDL-style (entity) minus Rules v2 entity AUPRC $>0$ over all unseen entities.
\end{itemize}

\paragraph{Statistics and recorded decision rule.} The statistic is the difference of seed-mean metrics on the pooled test organizations. A paired organization-cluster bootstrap~\citep{efron1979bootstrap,field2007clusterbootstrap} uses \GhBootReps{} replicates of the \GhNTestOrgs{} test organizations (seed \GhBootSeed), shared across methods and pools. Each GNN replicate averages separately computed seed metrics. The protocol reports $(1+\#\{d_b\le0\})/(1+B)$ as a one-sided bootstrap-tail statistic, applies Holm over five tests, and records support when the point difference is positive and the adjusted statistic is $\le0.025$. Its minimum reported value \GhPMin{} is Monte Carlo resolution, not a bound on a population $p$-value. This uncentered tail convention is approximate, not an exact null-randomization test; adjustment alone does not establish finite-sample calibration. We preserve the recorded tests and their decisions. Percentile intervals condition on these organizations and fitted seeds; seed variability is reported separately.

\paragraph{Secondary and non-informative metrics.} Three secondary comparisons are descriptive and outside the family (all non-family CIs in this paper are unadjusted): S1 the rules + GraIL-style \texttt{lr} combination against each of its inputs on overall link AUPRC, S2 GraIL-style against Rules v2 on overall link AUPRC, and S3 RDL-style against Rules v2 on the entity twin pool. Link-level recall at 1\% FPR was declared non-informative in the prospective protocol, because structure-only baselines saturate it on v2 dev data (\cref{sec:leakage}); it is reported, but no claim rests on it.

\paragraph{Dev expectations.} Before the freeze, a dev rehearsal (train on \GhDevNTrainOrgs{} dev organizations, validate on \GhDevNValOrgs, evaluate on the remaining \GhDevNOrgs) recorded the expected directions: H1-h2 not supported (dev difference \GhHOneHopTwoDevDiff), H1-h3 and H1-h4 supported (\GhHOneHopThreeDevDiff, \GhHOneHopFourDevDiff), H2 not supported (\GhHTwoDevDiff), H3 supported (\GhHThreeDevDiff). These were recorded so that the expected negative results would be reported as such.

\section{Cross-Method Comparison and Robustness Studies}
\label{sec:addendum}

The extension adds generic rules, explicit multi-hop references, a lower-prevalence synthetic cohort and identity-error conditions. It was designed after original results were available, with its code, data manifest and nine-test decision family internally frozen before new formal computations (SHA-256 prefix \texttt{\GvV{freeze.sha}}). The original five decisions remain unchanged. A/C reuse the already-evaluated \texttt{hrgraph-004} organizations; B evaluates a new \texttt{hrgraph-005} cohort from the same generator (\cref{tab:study-lineage}). The run verified \GvV{run.freezev2} extension and \GvV{run.freezeorig} original files and completed in \GvV{run.wallh}\,h on CPU.

\subsection{New arms and data}
\label{sec:addendum-methods}

\paragraph{Generic rules (not tuned to the generator).} Ten standard payroll and vendor red flags, each counted once: shared bank account; third-party payee without authorization on file; shared residential address or non-office phone with an existing employee; unverified bank change within 5 days of a run; SoD conflict without documented exception; paid with zero attendance or with low badge and login activity; approver conflict (shared identifier or non-independent approval); identifier match without a related-party disclosure on file; and any missing control record. Unlike Rules v2 there are no weights and no exemption logic that combines disclosure timing with four-eyes approval. The authors who wrote it know the generator, so it is untuned, not blind.

\paragraph{Chained-join query and chained-join GBDT.} We enumerate every simple path of length 1--4 from a new entity to an existing employee through identifier and person records (bank accounts, addresses, phones, persons), typed by relation and node class. The \emph{chained-join query} is a deterministic audit query: shortest identifier chain first, then the number of chains, with commercial addresses and office switchboard lines ignored. The \emph{chained-join GBDT} is a \emph{tuned} gradient-boosted tree model~\citep{pedregosa2011sklearn} on hand-crafted features: log counts of \GvV{A.vocab} path types, counts per length, the shortest length, and the same observed record attributes that the GNNs receive for both endpoints (\GvV{A.linkfeatures} features). It is trained on the 9 training dev organizations, and its hyperparameters were chosen by a pre-specified grid (\GvV{A.gbdt.link.ngrid} link and \GvV{A.gbdt.ent.ngrid} entity settings) on the 3 validation dev organizations (link: \GvV{A.gbdt.link.hp}). An entity-level variant aggregates the path features over all candidates.

\paragraph{A 2\% positive-rate synthetic cohort (\texttt{hrgraph-005}).} The same generator code with larger organizations (600--900 existing employees), 700 new entities per organization and one copy of each planted pattern gives 14 frauds per organization, \GvV{B.prevpct}\% of new entities; every fraud keeps one matched twin. The 24 test organizations hold \GvV{B.nent} new entities, \GvV{B.npos} of them fraudulent. All learned models (R-GCN and RDL-style entity GNNs with the frozen hyperparameters and seeds, and the chained-join GBDT) were retrained on its 12 dev organizations. This corpus is evaluated at entity level only: its \textasciitilde19 million candidate pairs were not scored. We report precision and recall when each organization reviews a fixed share of its new entities (1, 2 or 5\%; 2\% equals the number of frauds per organization).

\paragraph{Identity-resolution noise.} On the observed test graphs, each identifier link (to a bank account, address or phone) is re-pointed with probability $p\in\{0.1,0.2,0.4\}$ to a duplicate record, so that the join fails (``split''). In a second family of conditions, each emergency-contact, account-holder, representative and conflict-of-interest record is missing with probability $q\in\{0.1,0.2,0.4\}$ (``drop''). Models are trained on clean dev organizations. GraIL was not retrained (\textasciitilde11.5\,h per run), so the NBFNet-style model is the link GNN of this stage. Retrained with the frozen code, seeds and hyperparameters, the RDL-style model reproduced its frozen test scores exactly, while the NBFNet-style model did not reproduce bitwise under the changed thread count (clean link AUPRC \GvV{C.repro.nbfnet.retrained} against \GvV{C.repro.nbfnet.frozen} frozen; disclosed).

\begin{table}[t]
\centering
\caption{Extension: the nine pre-specified tests (one Holm family, one-sided $\alpha=0.025$). Stage A reuses the \texttt{hrgraph-004} test organizations (new arms against the frozen GNN scores); stage B uses the new 2\%-prevalence corpus \texttt{hrgraph-005}; stage C uses the \texttt{hrgraph-004} test organizations with 20\% of identifier links split (failed joins). Difference = method $-$ reference with 95\% paired org-bootstrap CI; $p$ one-sided. Designed after the original results were examined; the original five tests are unchanged.}
\label{tab:v2-hyp}
\small
\begin{adjustbox}{max width=\textwidth}
\begin{tabular}{llrrlrrl}
\toprule
id & test (method $-$ reference, metric) & method & ref. & difference [95\% CI] & $p$ & Holm $p$ & supported \\
\midrule
A1 & Rules v2 $-$ generic rules, link twin AUROC & 0.796 & 0.667 & +0.129 [+0.109, +0.147] & 0.001 & 0.009 & yes \\
A2 & GraIL $k{=}3$ $-$ generic rules, link twin AUROC & 0.674 & 0.667 & +0.007 [$-$0.020, +0.033] & 0.317 & 1.000 & no \\
A3 & GraIL $k{=}3$ $-$ chained-join GBDT, link AUPRC h3 & 0.501 & 0.692 & $-$0.191 [$-$0.225, $-$0.152] & 1.000 & 1.000 & no \\
A4 & GraIL $k{=}3$ $-$ chained-join GBDT, link AUPRC h4 & 0.272 & 0.646 & $-$0.374 [$-$0.443, $-$0.292] & 1.000 & 1.000 & no \\
A5 & RDL-style $-$ chained-join GBDT, entity AUPRC & 0.752 & 0.861 & $-$0.109 [$-$0.127, $-$0.089] & 1.000 & 1.000 & no \\
B1 & RDL-style $-$ Rules v2, entity AUPRC (2\% prevalence) & 0.707 & 0.504 & +0.202 [+0.147, +0.256] & 0.001 & 0.009 & yes \\
B2 & RDL-style $-$ Rules v2, precision at 2\% budget & 0.695 & 0.602 & +0.094 [+0.045, +0.139] & 0.001 & 0.009 & yes \\
C1 & NBFNet-style $-$ Rules v2, link AUPRC h3 (idsplit 0.2) & 0.441 & 0.090 & +0.350 [+0.299, +0.395] & 0.001 & 0.009 & yes \\
C2 & RDL-style $-$ Rules v2, entity AUPRC (idsplit 0.2) & 0.731 & 0.684 & +0.047 [+0.016, +0.080] & 0.003 & 0.015 & yes \\
\bottomrule
\end{tabular}
\end{adjustbox}
\end{table}

\subsection{Results}

\GvV{nsupported} of the \GvV{nfamily} addendum hypotheses were supported (\GvV{supportedlist}) and \GvV{nnotsupported} were not (\GvV{notsupportedlist}); see \cref{tab:v2-hyp}.

\paragraph{The multi-hop advantage is not specific to GNNs (A3--A5 not supported).} The tuned chained-join GBDT was the strongest model in the whole study (\cref{tab:v2-a-link,tab:v2-a-ent}). Its link AUPRC was \GvV{A.link.cj-gbdt.auprc} overall, against \GvV{A.link.grail-k3.auprc} for the GraIL-style model and \GvV{A.link.rules-v2+grail-k3:lr.auprc} for the best rules + GNN combination. At 3 hops it reached \GvV{A3.ref} against \GvV{A3.method} (difference \GvV{A3.diff} \GvV{A3.diffci}), and at 4 hops \GvV{A4.ref} against \GvV{A4.method}. At entity level it reached AUPRC \GvV{A5.ref} against \GvV{A5.method} for the RDL-style model. It also read the records best: its link twin AUROC was \GvV{A.link.cj-gbdt.twins:auroc} \GvV{A.link.cj-gbdt.twins:auroc.ci}, above Rules v2 (\GvV{A.link.rules-v2.twins:auroc}). The deterministic chained-join query reaches 4-hop colluders (AUPRC \GvV{A.link.cj-count.auprc|hop:4}) but cannot read records (twin AUROC \GvV{A.link.cj-count.twins:auroc}). Following the pre-specified interpretation rule, the H1 finding must be restated. \emph{Multi-hop models} reach colluders that one-step rules cannot name, and a tuned model on explicit path features did so better than the subgraph and path GNNs in our setting. The finding is not a GNN-specific advantage.

\paragraph{Record discrimination depends on rule specification (A1 supported, A2 not).} Generic rules separated frauds from their twins much worse than Rules v2 (link twin AUROC \GvV{A1.ref} vs.\ \GvV{A1.method}; difference \GvV{A1.diff} \GvV{A1.diffci}, Holm $p=\GvV{A1.holm}$). Against these rules, GraIL-style had AUROC \GvV{A2.method} versus \GvV{A2.ref}, a difference of \GvV{A2.diff} \GvV{A2.diffci}. The directional hypothesis was unsupported; this is not equivalence. Weights, exemptions and logic also change. In accordance with the extension's interpretation rule, explanations beyond generator knowledge remain; this comparison does not identify knowledge as the predominant cause of H2's result. Generic rules were also far weaker at entity level (AUPRC \GvV{A.entity.rules-generic.auprc} vs.\ \GvV{A.entity.rules-v2.auprc}).

\begin{table}[t]
\centering
\caption{Stage A, link level, on the \texttt{hrgraph-004} test organizations (483,912 candidate pairs). $^\ast$\,New arm of addendum v2; all other rows are the frozen scores of the confirmatory run (GNN values = mean of 3 seeds). AUPRC per hop = hop-$h$ positives vs.\ all negatives; twin AUROC = planted positives vs.\ their matched benign twins. Brackets: 95\% org-bootstrap CI.}
\label{tab:v2-a-link}
\small
\begin{adjustbox}{max width=\textwidth}
\begin{tabular}{lrrrllll}
\toprule
method & AUPRC & h1 & h2 & h3 & h4 & twin AUROC \\
\midrule
Rules v2 & 0.347 & 0.324 & 0.650 & 0.081 {\scriptsize [0.066, 0.098]} & 0.000 {\scriptsize [0.000, 0.000]} & 0.796 {\scriptsize [0.773, 0.818]} \\
Generic rules$^\ast$ & 0.186 & 0.084 & 0.292 & 0.025 {\scriptsize [0.022, 0.029]} & 0.000 {\scriptsize [0.000, 0.000]} & 0.667 {\scriptsize [0.650, 0.682]} \\
Path features + LR (len.\,3) & 0.295 & 0.051 & 0.177 & 0.239 {\scriptsize [0.222, 0.260]} & 0.001 {\scriptsize [0.001, 0.002]} & 0.619 {\scriptsize [0.602, 0.637]} \\
Chained-join query$^\ast$ & 0.311 & 0.000 & 0.282 & 0.220 {\scriptsize [0.209, 0.230]} & 0.104 {\scriptsize [0.092, 0.117]} & 0.500 {\scriptsize [0.494, 0.506]} \\
Chained-join GBDT$^\ast$ & 0.823 & 0.170 & 0.603 & 0.692 {\scriptsize [0.631, 0.746]} & 0.646 {\scriptsize [0.536, 0.755]} & 0.853 {\scriptsize [0.826, 0.876]} \\
GraIL-style $k{=}3$ & 0.663 & 0.140 & 0.442 & 0.501 {\scriptsize [0.462, 0.549]} & 0.272 {\scriptsize [0.220, 0.337]} & 0.674 {\scriptsize [0.643, 0.705]} \\
NBFNet-style & 0.673 & 0.386 & 0.518 & 0.461 {\scriptsize [0.420, 0.502]} & 0.154 {\scriptsize [0.132, 0.188]} & 0.668 {\scriptsize [0.642, 0.691]} \\
Opening-subgraph & 0.584 & 0.045 & 0.371 & 0.394 {\scriptsize [0.357, 0.444]} & 0.197 {\scriptsize [0.174, 0.229]} & 0.635 {\scriptsize [0.611, 0.663]} \\
Rules v2 + GraIL $k{=}3$ (LR) & 0.735 & 0.201 & 0.730 & 0.437 {\scriptsize [0.412, 0.468]} & 0.190 {\scriptsize [0.160, 0.224]} & 0.763 {\scriptsize [0.742, 0.783]} \\
\bottomrule
\end{tabular}
\end{adjustbox}
\end{table}

\begin{table}[t]
\centering
\caption{Stage A, entity level, \texttt{hrgraph-004} test organizations (2,496 new entities, 26.9\% fraudulent). AUPRC@2\% = AUPRC with negatives re-weighted to 2\% prevalence (descriptive); P@$b$ = precision when each organization reviews its top $b$ of new entities. $^\ast$\,New arm; other rows frozen.}
\label{tab:v2-a-ent}
\small
\begin{adjustbox}{max width=\textwidth}
\begin{tabular}{lrllrrr}
\toprule
method & AUROC & AUPRC & twin AUROC & AUPRC@2\% & P@2\% & P@5\% \\
\midrule
Rules v2 & 0.872 & 0.698 {\scriptsize [0.669, 0.728]} & 0.817 {\scriptsize [0.793, 0.839]} & 0.158 & 0.858 & 0.826 \\
Generic rules$^\ast$ & 0.780 & 0.490 {\scriptsize [0.467, 0.515]} & 0.735 {\scriptsize [0.712, 0.756]} & 0.054 & 0.578 & 0.616 \\
Path features + LR (len.\,3) & 0.876 & 0.725 {\scriptsize [0.692, 0.757]} & 0.815 {\scriptsize [0.788, 0.837]} & 0.165 & 0.889 & 0.908 \\
Chained-join query$^\ast$ & 0.628 & 0.325 {\scriptsize [0.323, 0.327]} & 0.500 {\scriptsize [0.495, 0.504]} & 0.027 & 0.117 & 0.146 \\
Chained-join GBDT$^\ast$ & 0.952 & 0.861 {\scriptsize [0.832, 0.884]} & 0.881 {\scriptsize [0.860, 0.898]} & 0.329 & 0.958 & 0.950 \\
R-GCN entity & 0.900 & 0.776 {\scriptsize [0.744, 0.808]} & 0.821 {\scriptsize [0.796, 0.846]} & 0.237 & 0.931 & 0.922 \\
RDL-style entity & 0.890 & 0.752 {\scriptsize [0.719, 0.784]} & 0.821 {\scriptsize [0.797, 0.843]} & 0.198 & 0.896 & 0.881 \\
Rules v2 + RDL-style (LR) & 0.914 & 0.792 {\scriptsize [0.763, 0.822]} & 0.844 {\scriptsize [0.822, 0.866]} & 0.225 & 0.889 & 0.894 \\
\bottomrule
\end{tabular}
\end{adjustbox}
\end{table}

\begin{figure}[t]
\centering
\begin{tikzpicture}
\begin{axis}[width=0.72\textwidth, height=5.6cm, xlabel={recall}, ylabel={precision (interpolated)}, xmin=0, xmax=1, ymin=0, ymax=1.02,
  legend style={font=\scriptsize, at={(1.02,1)}, anchor=north west}, grid=major, grid style={gray!20}, tick label style={font=\small}]
\addplot[c1, thick, solid, mark=none] coordinates {(0.05,0.8600) (0.10,0.8600) (0.15,0.8239) (0.20,0.8239) (0.25,0.8239) (0.30,0.8239) (0.35,0.7158) (0.40,0.7158) (0.45,0.7158) (0.50,0.7158) (0.55,0.7158) (0.60,0.6458) (0.65,0.1046) (0.70,0.1046) (0.75,0.1046) (0.80,0.1046) (0.85,0.1046) (0.90,0.0200) (0.95,0.0200) (1.00,0.0200)};
\addlegendentry{Rules v2}
\addplot[c4, thick, solid, mark=none] coordinates {(0.05,0.3971) (0.10,0.3971) (0.15,0.3971) (0.20,0.1611) (0.25,0.1611) (0.30,0.1611) (0.35,0.1611) (0.40,0.1611) (0.45,0.1611) (0.50,0.1611) (0.55,0.1611) (0.60,0.0471) (0.65,0.0471) (0.70,0.0471) (0.75,0.0471) (0.80,0.0471) (0.85,0.0471) (0.90,0.0471) (0.95,0.0200) (1.00,0.0200)};
\addlegendentry{Generic rules}
\addplot[gray, thick, dotted, mark=none] coordinates {(0.05,0.3235) (0.10,0.3235) (0.15,0.3235) (0.20,0.3235) (0.25,0.3235) (0.30,0.3235) (0.35,0.3235) (0.40,0.3235) (0.45,0.3235) (0.50,0.3235) (0.55,0.3235) (0.60,0.3235) (0.65,0.3235) (0.70,0.3235) (0.75,0.3235) (0.80,0.0200) (0.85,0.0200) (0.90,0.0200) (0.95,0.0200) (1.00,0.0200)};
\addlegendentry{Chained-join query}
\addplot[c3, thick, solid, mark=none] coordinates {(0.05,1.0000) (0.10,0.9219) (0.15,0.9219) (0.20,0.9079) (0.25,0.8878) (0.30,0.8736) (0.35,0.8736) (0.40,0.8736) (0.45,0.8736) (0.50,0.8416) (0.55,0.8253) (0.60,0.8145) (0.65,0.7789) (0.70,0.7628) (0.75,0.7347) (0.80,0.6985) (0.85,0.6502) (0.90,0.6084) (0.95,0.5186) (1.00,0.1193)};
\addlegendentry{Chained-join GBDT}
\addplot[c2, thick, dashed, mark=none] coordinates {(0.05,0.9710) (0.10,0.9238) (0.15,0.8889) (0.20,0.8542) (0.25,0.8416) (0.30,0.8310) (0.35,0.8077) (0.40,0.8054) (0.45,0.7868) (0.50,0.7639) (0.55,0.7198) (0.60,0.7064) (0.65,0.6667) (0.70,0.6392) (0.75,0.6063) (0.80,0.5547) (0.85,0.4324) (0.90,0.2854) (0.95,0.1740) (1.00,0.0254)};
\addlegendentry{R-GCN entity}
\addplot[c2, thick, solid, mark=none] coordinates {(0.05,0.9343) (0.10,0.9135) (0.15,0.9068) (0.20,0.8960) (0.25,0.8830) (0.30,0.8526) (0.35,0.8459) (0.40,0.8073) (0.45,0.7910) (0.50,0.7863) (0.55,0.7712) (0.60,0.7526) (0.65,0.7263) (0.70,0.6964) (0.75,0.6598) (0.80,0.6062) (0.85,0.5482) (0.90,0.3472) (0.95,0.0679) (1.00,0.0212)};
\addlegendentry{RDL-style entity}
\end{axis}
\end{tikzpicture}
\caption{Stage B: precision--recall curves at the natural 2.0\% prevalence of \texttt{hrgraph-005} (pooled 24 test organizations; GNN curves = mean over 3 seeds of the interpolated precision at recall $0.05,\dots,1$). Exact values are in \cref{tab:v2-b}.}
\label{fig:v2-pr}
\end{figure}
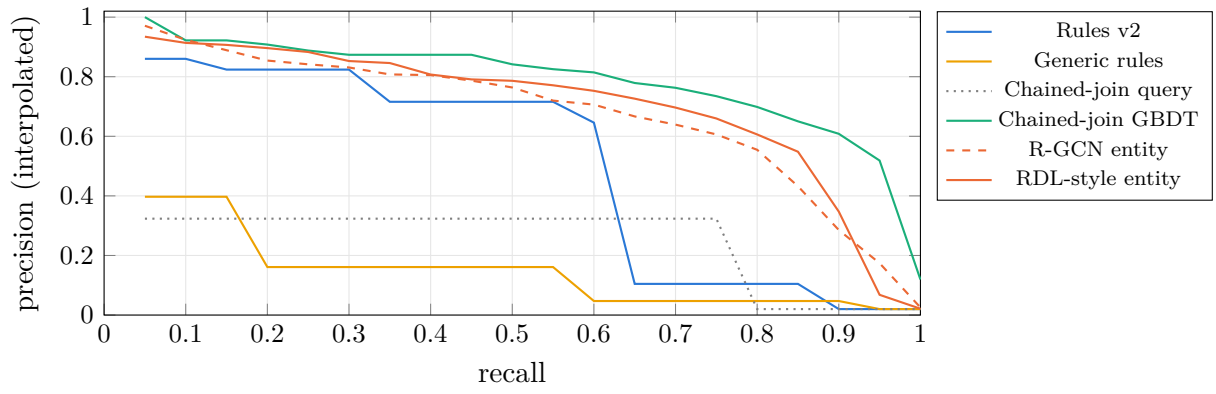

\paragraph{Ranking and review precision on the 2\% cohort (B1, B2 supported).} On \texttt{hrgraph-005} the RDL-style model reached entity AUPRC \GvV{B1.method} \GvV{B1.methodci} against \GvV{B1.ref} \GvV{B1.refci} for Rules v2 (difference \GvV{B1.diff} \GvV{B1.diffci}; \cref{tab:v2-b}, \cref{fig:v2-pr}). With a review budget of 2\% of new entities per organization, \GvV{B2.method} of the reviewed entities were frauds for the RDL-style model and \GvV{B2.ref} for the rules (\GvV{B2.diff} \GvV{B2.diffci}). At a 5\% budget the RDL-style model recovered \GvV{B.rdl-ent.r@budget0.05} of all frauds and the rules \GvV{B.rules-v2.r@budget0.05}. This differs from the fixed-score re-weighting in \cref{sec:posthoc}. Organization size, new-entity counts, benign composition and training data also change, so the comparison does not isolate prevalence as a cause. One possible reason, which we did not test, is that the rules' coarse, tied scores rank many ordinary entities with a single missing or late record alongside the frauds. The chained-join GBDT was again the best (AUPRC \GvV{B.cj-gbdt-ent.auprc}; P@2\% \GvV{B.cj-gbdt-ent.p@budget0.02}; R@5\% \GvV{B.cj-gbdt-ent.r@budget0.05}). Generic rules fell to AUPRC \GvV{B.rules-generic.auprc}, and on the twin pool no method exceeded \GvV{B.cj-gbdt-ent.twins:auroc}. These numbers come from one synthetic mix of benign entities and are not deployment precision.

\begin{table}[t]
\centering
\caption{Stage B: a 2\% positive-rate synthetic cohort. New corpus \texttt{hrgraph-005} (24 unseen test organizations, 16,800 new entities, 336 fraudulent = 2.0\%); every model retrained on its 12 dev organizations. P@$b$ / R@$b$ = precision / recall when each organization reviews its top $b$ of new entities (at 2\%, 14 reviews per organization, equal to its number of frauds, so P@2\% = R@2\%). AUPRC@1\% = negatives re-weighted to 1\% prevalence (descriptive). $^\ast$\,Not a GNN, new in addendum v2. Brackets: 95\% org-bootstrap CI.}
\label{tab:v2-b}
\small
\begin{adjustbox}{max width=\textwidth}
\begin{tabular}{lrlrrlrrr}
\toprule
method & AUROC & AUPRC & twin AUROC & P@1\% & P@2\% & P@5\% & R@5\% & AUPRC@1\% \\
\midrule
Rules v2 & 0.900 & 0.504 {\scriptsize [0.453, 0.559]} & 0.792 & 0.800 & 0.602 {\scriptsize [0.561, 0.645]} & 0.267 & 0.668 & 0.405 \\
Generic rules$^\ast$ & 0.855 & 0.155 {\scriptsize [0.123, 0.198]} & 0.728 & 0.343 & 0.239 {\scriptsize [0.203, 0.276]} & 0.174 & 0.434 & 0.091 \\
Chained-join query$^\ast$ & 0.871 & 0.211 {\scriptsize [0.208, 0.215]} & 0.500 & 0.098 & 0.184 {\scriptsize [0.171, 0.197]} & 0.314 & 0.786 & 0.121 \\
Chained-join GBDT$^\ast$ & 0.995 & 0.798 {\scriptsize [0.761, 0.838]} & 0.828 & 0.851 & 0.735 {\scriptsize [0.702, 0.768]} & 0.398 & 0.994 & 0.681 \\
R-GCN entity & 0.974 & 0.686 {\scriptsize [0.651, 0.726]} & 0.749 & 0.780 & 0.670 {\scriptsize [0.639, 0.700]} & 0.358 & 0.894 & 0.561 \\
RDL-style entity & 0.959 & 0.707 {\scriptsize [0.669, 0.754]} & 0.772 & 0.808 & 0.695 {\scriptsize [0.672, 0.721]} & 0.360 & 0.900 & 0.586 \\
\bottomrule
\end{tabular}
\end{adjustbox}
\end{table}

\paragraph{Both advantages survive failed identity joins (C1, C2 supported).} With 20\% of identifier links split, the NBFNet-style model kept 3-hop link AUPRC \GvV{C1.method} against \GvV{C1.ref} for Rules v2, and the RDL-style model kept entity AUPRC \GvV{C2.method} against \GvV{C2.ref} (\cref{tab:v2-c}). The degradation is real. The 4-hop AUPRC of the NBFNet-style model fell from \GvV{C.clean.link.nbfnet.auprc|hop:4} to \GvV{C.idsplit-40.link.nbfnet.auprc|hop:4} at $p=0.4$, and that of the chained-join GBDT from \GvV{C.clean.link.cj-gbdt.auprc|hop:4} to \GvV{C.idsplit-40.link.cj-gbdt.auprc|hop:4}, because a 4-hop chain breaks if any of its identifier links is split. The entity-level GNN degraded least: at $p=0.4$ the RDL-style model (\GvV{C.idsplit-40.entity.rdl-ent.auprc}) was ahead of both the chained-join GBDT (\GvV{C.idsplit-40.entity.cj-gbdt-ent.auprc}) and Rules v2 (\GvV{C.idsplit-40.entity.rules-v2.auprc}). Missing relationship records hurt the 3-hop link results most (chained-join GBDT \GvV{C.clean.link.cj-gbdt.auprc|hop:3} to \GvV{C.recdrop-40.link.cj-gbdt.auprc|hop:3} at $q=0.4$), because they remove the middle of the chain. Our noise model is simple: duplicates are clean singletons, and the joins never create false links.

\begin{table}[t]
\centering
\caption{Stage C: identity-resolution noise on the \texttt{hrgraph-004} test organizations. ``split $p$'': each identifier link (to a bank account, address or phone) is re-pointed with probability $p$ to a duplicate record, so the join fails; ``drop $q$'': each emergency-contact, account-holder, representative and conflict-of-interest record is missing with probability $q$. Models trained on clean dev organizations; NBFNet-style and RDL-style retrained with the frozen settings (3 seeds). Pre-specified condition: split 0.2.}
\label{tab:v2-c}
\small
\begin{adjustbox}{max width=\textwidth}
\begin{tabular}{lrrrrrrr}
\toprule
method & clean & split 0.1 & split 0.2 & split 0.4 & drop 0.1 & drop 0.2 & drop 0.4 \\
\midrule
\multicolumn{8}{l}{\emph{link AUPRC, all hops}} \\
Rules v2 & 0.347 & 0.305 & 0.292 & 0.201 & 0.347 & 0.344 & 0.338 \\
Generic rules & 0.186 & 0.165 & 0.161 & 0.122 & 0.185 & 0.184 & 0.184 \\
Chained-join query & 0.311 & 0.251 & 0.219 & 0.123 & 0.279 & 0.262 & 0.208 \\
Chained-join GBDT & 0.823 & 0.716 & 0.664 & 0.474 & 0.750 & 0.701 & 0.576 \\
NBFNet-style & 0.688 & 0.605 & 0.553 & 0.399 & 0.610 & 0.556 & 0.447 \\
\midrule
\multicolumn{8}{l}{\emph{link AUPRC, 3-hop colluders}} \\
Rules v2 & 0.081 & 0.084 & 0.090 & 0.086 & 0.077 & 0.073 & 0.066 \\
Generic rules & 0.025 & 0.026 & 0.027 & 0.032 & 0.025 & 0.024 & 0.024 \\
Chained-join query & 0.220 & 0.198 & 0.173 & 0.117 & 0.171 & 0.134 & 0.071 \\
Chained-join GBDT & 0.692 & 0.638 & 0.608 & 0.453 & 0.577 & 0.480 & 0.340 \\
NBFNet-style & 0.502 & 0.478 & 0.441 & 0.334 & 0.385 & 0.304 & 0.179 \\
\midrule
\multicolumn{8}{l}{\emph{entity AUPRC}} \\
Rules v2 & 0.698 & 0.674 & 0.684 & 0.626 & 0.688 & 0.685 & 0.655 \\
Generic rules & 0.490 & 0.480 & 0.494 & 0.471 & 0.484 & 0.482 & 0.466 \\
Chained-join query & 0.325 & 0.310 & 0.305 & 0.279 & 0.320 & 0.320 & 0.310 \\
Chained-join GBDT & 0.861 & 0.800 & 0.778 & 0.683 & 0.821 & 0.808 & 0.736 \\
RDL-style entity & 0.752 & 0.735 & 0.731 & 0.712 & 0.737 & 0.731 & 0.705 \\
\bottomrule
\end{tabular}
\end{adjustbox}
\end{table}

\paragraph{What changes in the paper's reading.} (i) The reach claim is restated as a claim about multi-hop methods, not about GNNs. On these graphs, a tuned GBDT on explicit chained-join features is a stronger and cheaper baseline than any GNN we ran. (ii) Record discrimination is sensitive to rule specification; the generic-rule comparison establishes neither equivalence nor a causal knowledge effect. (iii) The entity GNN exceeds rules in the new 2\% cohort and specified split-noise conditions; boosting remains strongest except in the high split-noise condition. (iv) Findings concern the stated synthetic populations and perturbations, not estimated operational performance.

\FloatBarrier

\section{Original Frozen Study: Results and Recorded Decisions}
\label{sec:results}

Subsections~\ref{sec:h1}--\ref{sec:h3} report the five confirmatory tests; the secondary comparisons and further results are descriptive, and the exploratory post-hoc analyses follow in \cref{sec:posthoc}. \GhNSupported{} of the \GhNFamily{} pre-specified hypotheses are supported and \GhNNotSupported{} are not (\cref{tab:hypotheses}). The GraIL-style model's multi-hop reach beats the one-step record-aware rules for colluders 3 and 4 hops away, and the RDL-style entity model beats them at entity level; the GraIL-style model does not beat the generator-aware rules for 2-hop collusion or at separating a fraud from its benign twin. Every outcome has the direction recorded from the dev rehearsal before the freeze (\cref{tab:devtest}), and, in an exploratory check, every difference has the same sign for each of the \GhNSeeds{} GNN seeds (\cref{sec:posthoc}).

\subsection{Run integrity}

The frozen run script was executed once. It verified the \GhFreezeFiles{} frozen files and all \GhManifestFiles{} data files (\GhManifestMismatch{} mismatches) before any training, trained every model with \GhNSeeds{} seeds, and finished without a crash in \GhWallHours{} hours of wall time (\GhWallSec{} s; the GraIL-style $k=3$ model alone took \GhGraILThreeHours{} hours for \GhNSeeds{} seeds on CPU). The formal run log records no crash or rerun of this stage; development and extension work are tracked separately (\cref{sec:deviations}). The evaluation pool is exactly the pre-specified one: \GhNTestOrgs{} test organizations, \GhNEntities{} entities (\GhNEntityPos{} positive) and \GhNPairs{} pairs (\GhNPairPos{} positive). Early stopping ended every training run before its epoch cap except \GhNCapHit{} (\cref{tab:training}).

% generated by xiaobo-experiment/render-results-draft.py; do not edit
\begin{table}[t]\centering\footnotesize\setlength{\tabcolsep}{3.5pt}
\caption{Pre-specified confirmatory tests on the 24 unseen test organizations: the GraIL-style model's multi-hop reach beats one-step record-aware rules at 3 and 4 hops and the RDL-style model beats them at entity level, but the GraIL-style model does not beat the generator-aware rules at 2 hops or on the twin pool. Reference = Rules v2 in every row; one-sided tests of model $-$ Rules v2 $>0$; paired org-cluster bootstrap (1{,}000 replicates, $p$ floor 0.001); Holm over the five tests; one-sided $\alpha=0.025$; CIs are per-test (unadjusted).}
\label{tab:hypotheses}
\begin{adjustbox}{max width=\linewidth}
\begin{tabular}{llllrrrrrl}
\toprule
id & pool / group & metric & model & \makecell[l]{model\\ {}[95\% CI]} & \makecell[l]{Rules v2\\ {}[95\% CI]} & \makecell[l]{difference\\ {}[95\% CI]} & $p$ & Holm $p$ & outcome \\
\midrule
H1-h2 & \makecell[l]{link, all pool\\ {}hop 2 ($n_+{=}$192)} & AUPRC & \makecell[l]{GraIL-style\\ {}$k{=}3$} & \makecell[r]{0.442\\ {\scriptsize [0.400, 0.497]}} & \makecell[r]{0.650\\ {\scriptsize [0.585, 0.716]}} & \makecell[r]{$-$0.208\\ {\scriptsize [$-$0.262, $-$0.150]}} & 1.000 & 1.000 & not supported \\
H1-h3 & \makecell[l]{link, all pool\\ {}hop 3 ($n_+{=}$240)} & AUPRC & \makecell[l]{GraIL-style\\ {}$k{=}3$} & \makecell[r]{0.501\\ {\scriptsize [0.462, 0.549]}} & \makecell[r]{0.081\\ {\scriptsize [0.066, 0.098]}} & \makecell[r]{+0.420\\ {\scriptsize [0.375, 0.469]}} & 0.001 & 0.005 & \textbf{supported} \\
H1-h4 & \makecell[l]{link, all pool\\ {}hop 4 ($n_+{=}$96)} & AUPRC & \makecell[l]{GraIL-style\\ {}$k{=}3$} & \makecell[r]{0.272\\ {\scriptsize [0.220, 0.337]}} & \makecell[r]{0.000\\ {\scriptsize [0.000, 0.000]}} & \makecell[r]{+0.272\\ {\scriptsize [0.219, 0.337]}} & 0.001 & 0.005 & \textbf{supported} \\
H2 & \makecell[l]{link, twin pool\\ {}overall ($n_+{=}$576)} & AUROC & \makecell[l]{GraIL-style\\ {}$k{=}3$} & \makecell[r]{0.674\\ {\scriptsize [0.643, 0.705]}} & \makecell[r]{0.796\\ {\scriptsize [0.773, 0.818]}} & \makecell[r]{$-$0.122\\ {\scriptsize [$-$0.151, $-$0.094]}} & 1.000 & 1.000 & not supported \\
H3 & \makecell[l]{entity, all pool\\ {}overall ($n_+{=}$672)} & AUPRC & \makecell[l]{RDL-style\\ {}(entity)} & \makecell[r]{0.752\\ {\scriptsize [0.719, 0.784]}} & \makecell[r]{0.698\\ {\scriptsize [0.669, 0.728]}} & \makecell[r]{+0.053\\ {\scriptsize [0.025, 0.086]}} & 0.002 & 0.006 & \textbf{supported} \\
\bottomrule
\end{tabular}
\end{adjustbox}
\end{table}

\subsection{H1 (confirmatory): multi-hop reach beats one-step rules at 3 and 4 hops}
\label{sec:h1}

\paragraph{H1-h3 and H1-h4 are supported.} For 3-hop collusion the GraIL-style model had link AUPRC \GhHOneHopThreeMethod{} \GhHOneHopThreeMethodCI{} against \GhHOneHopThreeRef{} \GhHOneHopThreeRefCI{} for Rules v2, a difference of \GhHOneHopThreeDiff{} \GhHOneHopThreeDiffCI{} (Holm $p=\GhHOneHopThreePHolm$). For 4-hop collusion the difference is \GhHOneHopFourDiff{} \GhHOneHopFourDiffCI{} (Holm $p=\GhHOneHopFourPHolm$). Rules v2 has 4-hop AUPRC \GhHOneHopFourRef{}, at chance level: a one-step rule cannot name a colluder whose link to the new entity passes through an account holder and that holder's phone. H1-h4 therefore checks multi-hop access relative to a reference with a shorter evidence horizon; H1-h3 is less so, because Rules v2 names some 3-hop colluders through the manager or approver edge. The recorded bootstrap-tail statistics reach their simulation floor (\GhPMin{} before adjustment), not a certified bound on a population $p$-value.

\paragraph{H1-h2 is not supported.} For 2-hop collusion---where the evidence is a direct identifier match such as a shared account, address or phone---the GraIL-style model had AUPRC \GhHOneHopTwoMethod{} \GhHOneHopTwoMethodCI{} against \GhHOneHopTwoRef{} \GhHOneHopTwoRefCI{} for the rules, a difference of \GhHOneHopTwoDiff{} \GhHOneHopTwoDiffCI. The one-sided test cannot show that the rules are better; descriptively they are, and the dev rehearsal had predicted this direction (\GhHOneHopTwoDevDiff).

\paragraph{What the hop-$h$ AUPRC measures.} The hop-$h$ pool contains every matched twin as a negative, so a hop-$h$ AUPRC mixes two abilities: reaching the colluder and ranking the fraud above its twin. A detector that reaches every hop-$h$ colluder but cannot tell fraud from twin scores about 0.5 (\cref{sec:leakage}); the GraIL-style model's \GhHOneHopThreeMethod{} at 3 hops is at that level. The exploratory analysis of \cref{sec:posthoc} separates the two parts.

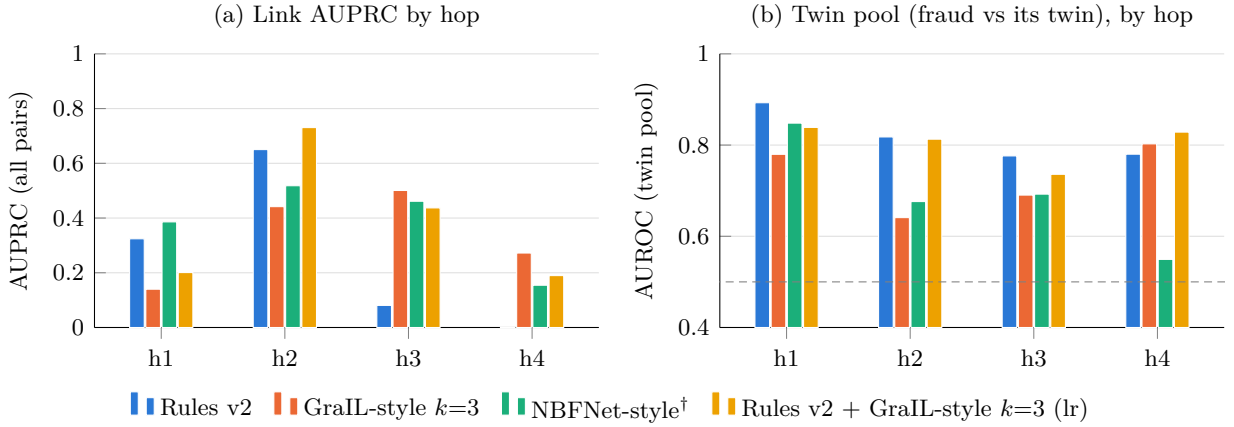
\begin{figure}[t]\centering
\begin{tikzpicture}
\begin{groupplot}[group style={group size=2 by 1, horizontal sep=1.6cm}, width=0.48\linewidth, height=5.2cm, ybar=0.6pt, /pgf/bar width=5.5pt, xtick={1,2,3,4}, xticklabels={h1,h2,h3,h4}, enlarge x limits=0.18, ymin=0, axis lines*=left, ymajorgrids, grid style={gray!25}, tick label style={font=\footnotesize}, label style={font=\footnotesize}, title style={font=\footnotesize}]
\nextgroupplot[ymax=1, ylabel={AUPRC (all pairs)}, title={(a) Link AUPRC by hop}, legend to name=hoplegend, legend columns=4, legend style={font=\footnotesize, draw=none, /tikz/every even column/.append style={column sep=6pt}}]
\addplot[fill=c1, draw=white, line width=0.4pt] coordinates {(1,0.3241) (2,0.6501) (3,0.0807) (4,0.0004)};
\addlegendentry{Rules v2}
\addplot[fill=c2, draw=white, line width=0.4pt] coordinates {(1,0.1399) (2,0.4416) (3,0.5007) (4,0.2721)};
\addlegendentry{GraIL-style $k{=}3$}
\addplot[fill=c3, draw=white, line width=0.4pt] coordinates {(1,0.3859) (2,0.5180) (3,0.4613) (4,0.1544)};
\addlegendentry{NBFNet-style$^\dagger$}
\addplot[fill=c4, draw=white, line width=0.4pt] coordinates {(1,0.2007) (2,0.7305) (3,0.4370) (4,0.1896)};
\addlegendentry{Rules v2 + GraIL-style $k{=}3$ (lr)}
\nextgroupplot[ymin=0.4, ymax=1, ylabel={AUROC (twin pool)}, title={(b) Twin pool (fraud vs its twin), by hop}]
\addplot[fill=c1, draw=white, line width=0.4pt] coordinates {(1,0.8928) (2,0.8178) (3,0.7762) (4,0.7800)};
\addplot[fill=c2, draw=white, line width=0.4pt] coordinates {(1,0.7797) (2,0.6409) (3,0.6903) (4,0.8026)};
\addplot[fill=c3, draw=white, line width=0.4pt] coordinates {(1,0.8481) (2,0.6761) (3,0.6924) (4,0.5493)};
\addplot[fill=c4, draw=white, line width=0.4pt] coordinates {(1,0.8387) (2,0.8129) (3,0.7359) (4,0.8284)};
\draw[densely dashed, gray] (axis cs:0.4,0.5) -- (axis cs:4.6,0.5);
\end{groupplot}
\end{tikzpicture}
\par\smallskip\pgfplotslegendfromname{hoplegend}
\caption{Original-stage comparison: subgraph and path GNNs reach distant colluders; the generator-aware rules are descriptively higher on the twin pool. (a) Link AUPRC per planted hop (hop-$h$ positives vs all negatives; not comparable across hops): record-aware rules beat both GNNs at 2 hops but cannot name a colluder at 4 hops, where the GraIL-style model is best. (b) On the twin pool (fraud vs its structurally identical benign twin; dashed line = chance) the rules lead at every hop except h4. Exact values: Tables~\ref{tab:link-main} and~\ref{tab:link-twins}. $^\dagger$Not a pre-specified primary model.}
\label{fig:hop}
\end{figure}

\Cref{fig:hop}(a) and \cref{tab:link-main} put these tests in context. The pattern is the same for every learned path or subgraph model: NBFNet-style (not pre-specified) reaches AUPRC \GhLinkNBFNetAUPRCHopThree{} at h3 and \GhLinkNBFNetAUPRCHopFour{} at h4, the opening-subgraph model (not pre-specified) \GhLinkOSFAAUPRCHopThree{} and \GhLinkOSFAAUPRCHopFour. GraIL-style $k=2$ reaches \GhLinkGraILTwoAUPRCHopThree{} at h3 but only \GhLinkGraILTwoAUPRCHopFour{} at h4, as expected from the $h\le k+1$ constraint. Path features + LR, the only non-GNN multi-hop model, is limited to paths of length 3 and has \GhLinkPathLRAUPRCHopThree{} at h3 and \GhLinkPathLRAUPRCHopFour{} at h4; a four-hop non-GNN reference was absent from this original method set and is supplied by the extension (\cref{sec:addendum}). The h1 cell (a ghost employee whose manager is also the hiring approver; \GhLinkPosHopOne{} positives) was not tested.

% generated by xiaobo-experiment/render-results-draft.py; do not edit

% generated by xiaobo-experiment/render-results-draft.py; do not edit
\begin{table}[t]\centering\footnotesize\setlength{\tabcolsep}{3.5pt}
\caption{Original-stage link level, all 483{,}912 candidate pairs: subgraph and path GNNs lead on 3- and 4-hop collusion, record-aware rules beat every single GNN on 2-hop collusion, and the rules + GraIL-style combination has the best overall AUPRC within this original method set. Per-hop columns compare hop-$h$ positives with all negatives, other-hop positives excluded ($n_+$ = 48 / 192 / 240 / 96; chance AUPRC = prevalence, at most 0.0005); AUPRC values are not comparable across hops because prevalence differs. GNN values are seed means $\pm$ seed SD; brackets are 95\% org-cluster bootstrap CIs. $^\dagger$GNN that is not a pre-specified primary model (descriptive).}
\label{tab:link-main}
\begin{adjustbox}{max width=\linewidth}
\begin{tabular}{lrrrrrrr}
\toprule
method & AUROC & \makecell[l]{AUPRC\\ {}[95\% CI]} & \makecell[l]{recall at\\ {}0.1\% FPR} & \makecell[l]{AUPRC\\ {}h1} & \makecell[l]{AUPRC\\ {}h2} & \makecell[l]{AUPRC\\ {}h3} & \makecell[l]{AUPRC\\ {}h4} \\
\midrule
Rules tuned & 0.765 & \makecell[r]{0.237\\ {\scriptsize [0.222, 0.253]}} & 0.434 & 0.084 & 0.495 & 0.020 & 0.000 \\
Rules v2 (record-aware) & 0.808 & \makecell[r]{0.347\\ {\scriptsize [0.321, 0.376]}} & 0.429 & 0.324 & 0.650 & 0.081 & 0.000 \\
Personalized PageRank & 0.993 & \makecell[r]{0.093\\ {\scriptsize [0.090, 0.100]}} & 0.057 & 0.070 & 0.047 & 0.037 & 0.006 \\
Path features + LR & 0.981 & \makecell[r]{0.295\\ {\scriptsize [0.281, 0.315]}} & 0.530 & 0.051 & 0.177 & 0.239 & 0.001 \\
RDL-style (link)$^\dagger$ & 0.953 & \makecell[r]{0.082$\pm$0.023\\ {\scriptsize [0.075, 0.092]}} & 0.148 & 0.014 & 0.078 & 0.021 & 0.011 \\
GraIL-style $k{=}2$$^\dagger$ & 0.979 & \makecell[r]{0.473$\pm$0.012\\ {\scriptsize [0.456, 0.493]}} & 0.789 & 0.036 & 0.424 & 0.392 & 0.002 \\
GraIL-style $k{=}3$ & 0.999 & \makecell[r]{0.663$\pm$0.071\\ {\scriptsize [0.630, 0.698]}} & 0.906 & 0.140 & 0.442 & 0.501 & 0.272 \\
NBFNet-style$^\dagger$ & 0.999 & \makecell[r]{0.673$\pm$0.017\\ {\scriptsize [0.645, 0.698]}} & 0.936 & 0.386 & 0.518 & 0.461 & 0.154 \\
Opening-subgraph (OSFA-insp.)$^\dagger$ & 0.999 & \makecell[r]{0.584$\pm$0.033\\ {\scriptsize [0.559, 0.617]}} & 0.900 & 0.045 & 0.371 & 0.394 & 0.197 \\
\midrule
Rules v2 + GraIL-style $k{=}3$ (lr) & 1.000 & \makecell[r]{0.735$\pm$0.026\\ {\scriptsize [0.711, 0.763]}} & 0.961 & 0.201 & 0.730 & 0.437 & 0.190 \\
Rules v2 + NBFNet-style$^\dagger$ (lr) & 0.999 & \makecell[r]{0.626$\pm$0.111\\ {\scriptsize [0.605, 0.648]}} & 0.910 & 0.222 & 0.481 & 0.409 & 0.164 \\
\bottomrule
\end{tabular}
\end{adjustbox}
\end{table}

\subsection{H2 (confirmatory): graph models do not read control records better than generator-aware rules}

\paragraph{H2 is not supported.} On the twin pool---the \GhXConcNLink{} planted fraud pairs and their \GhXConcNLink{} structurally identical benign twins---the GraIL-style model has AUROC \GhHTwoMethod{} \GhHTwoMethodCI{} and Rules v2 \GhHTwoRef{} \GhHTwoRefCI, a difference of \GhHTwoDiff{} \GhHTwoDiffCI. Topology cannot help here: a method can only rank a fraud above its twin by reading the declaration, authorization, verification, approval and activity records. The test was one-sided, so its failure shows only that the GNN did not beat the rules; descriptively, the rules are higher. Two features of the design favour the rules on this test: Rules v2 was written knowing which record types the generator emits, and the GNN hyperparameters were fixed on generator v1 and never tuned for v2. The comparison is therefore between generator-aware rules and untuned GNNs, and the result is conditional on the generator-informed rule specification, not a performance bound for other analysts.

The descriptive pattern holds beyond the pre-specified model (\cref{tab:link-twins}; unadjusted CIs). The paired-difference CI lies below zero for \GhNLinkTwinBelowRules{} of \GhNLinkSingleOthers{} other single link methods and for \GhNComboTwinBelowRules{} of \GhNLinkCombos{} rules + GNN combinations. The highest learned single model on the twin pool is the RDL-style link model (not pre-specified) (\GhLinkTwinRDLLinkAUROC{}, $\Delta={}$\GhLinkTwinRDLLinkDiff \GhLinkTwinRDLLinkDiffCI), whose node-wise embeddings are weak at finding the candidate (overall AUPRC \GhLinkRDLLinkAUPRC) but encode the entity's own records. Structure-only heuristics are at chance (\cref{tab:link-twins-full}). \Cref{fig:hop}(b) shows that the rules lead at h1--h3 and that the GraIL-style model is level with or above them only at h4 (\GhLinkTwinGraILThreeAUROCHopFour{} vs.\ \GhLinkTwinRulesVTwoAUROCHopFour), a descriptive per-hop view. At entity level the gap disappears: on the entity twin pool the pre-specified RDL-style entity model has a close point estimate to the rules (\cref{sec:h3}).

% generated by xiaobo-experiment/render-results-draft.py; do not edit
\begin{table}[t]\centering\footnotesize\setlength{\tabcolsep}{3.5pt}
\caption{Twin pool (each of the 576 planted fraud pairs against its structurally identical benign twin, 1{,}152 pairs in all): no learned link model separates fraud from its twin as well as the generator-aware Rules v2 (descriptive except for the GraIL-style $k{=}3$ row, which is H2). Per-hop columns compare same-hop positives and twins; $\Delta$ is the paired difference against Rules v2 with an unadjusted 95\% CI. $^\dagger$GNN that is not a pre-specified primary model (descriptive).}
\label{tab:link-twins}
\begin{adjustbox}{max width=\linewidth}
\begin{tabular}{lrrrrrr}
\toprule
method & \makecell[l]{twin AUROC\\ {}[95\% CI]} & h1 & h2 & h3 & h4 & \makecell[l]{$\Delta$ vs Rules v2\\ {}[95\% CI]} \\
\midrule
Rules tuned & \makecell[r]{0.547\\ {\scriptsize [0.537, 0.555]}} & 0.780 & 0.501 & 0.547 & 0.502 & \makecell[r]{$-$0.250\\ {\scriptsize [$-$0.269, $-$0.230]}} \\
Rules v2 (record-aware) & \makecell[r]{0.796\\ {\scriptsize [0.773, 0.818]}} & 0.893 & 0.818 & 0.776 & 0.780 & -- \\
Path features + LR & \makecell[r]{0.619\\ {\scriptsize [0.602, 0.637]}} & 0.829 & 0.629 & 0.644 & 0.742 & \makecell[r]{$-$0.177\\ {\scriptsize [$-$0.201, $-$0.151]}} \\
Personalized PageRank & \makecell[r]{0.501\\ {\scriptsize [0.496, 0.507]}} & 0.475 & 0.499 & 0.504 & 0.503 & \makecell[r]{$-$0.295\\ {\scriptsize [$-$0.317, $-$0.271]}} \\
RDL-style (link)$^\dagger$ & \makecell[r]{0.738$\pm$0.005\\ {\scriptsize [0.715, 0.764]}} & 0.856 & 0.743 & 0.720 & 0.718 & \makecell[r]{$-$0.058\\ {\scriptsize [$-$0.080, $-$0.034]}} \\
GraIL-style $k{=}2$$^\dagger$ & \makecell[r]{0.573$\pm$0.006\\ {\scriptsize [0.559, 0.588]}} & 0.748 & 0.592 & 0.609 & 0.740 & \makecell[r]{$-$0.224\\ {\scriptsize [$-$0.244, $-$0.202]}} \\
GraIL-style $k{=}3$ & \makecell[r]{0.674$\pm$0.057\\ {\scriptsize [0.643, 0.705]}} & 0.780 & 0.641 & 0.690 & 0.803 & \makecell[r]{$-$0.122\\ {\scriptsize [$-$0.151, $-$0.094]}} \\
NBFNet-style$^\dagger$ & \makecell[r]{0.668$\pm$0.025\\ {\scriptsize [0.642, 0.691]}} & 0.848 & 0.676 & 0.692 & 0.549 & \makecell[r]{$-$0.129\\ {\scriptsize [$-$0.152, $-$0.103]}} \\
Opening-subgraph (OSFA-insp.)$^\dagger$ & \makecell[r]{0.635$\pm$0.030\\ {\scriptsize [0.611, 0.663]}} & 0.874 & 0.612 & 0.648 & 0.713 & \makecell[r]{$-$0.161\\ {\scriptsize [$-$0.187, $-$0.131]}} \\
\midrule
Rules v2 + GraIL-style $k{=}3$ (lr) & \makecell[r]{0.763$\pm$0.036\\ {\scriptsize [0.742, 0.783]}} & 0.839 & 0.813 & 0.736 & 0.828 & \makecell[r]{$-$0.034\\ {\scriptsize [$-$0.047, $-$0.019]}} \\
Rules v2 + RDL-style (link)$^\dagger$ (lr) & \makecell[r]{0.772$\pm$0.003\\ {\scriptsize [0.751, 0.794]}} & 0.863 & 0.816 & 0.735 & 0.729 & \makecell[r]{$-$0.025\\ {\scriptsize [$-$0.039, $-$0.010]}} \\
\bottomrule
\end{tabular}
\end{adjustbox}
\end{table}

\subsection{H3 (confirmatory): an entity-level GNN ranks fraudulent new entities higher}
\label{sec:h3}

\paragraph{H3 is supported.} Over all \GhNEntities{} unseen entities, the RDL-style entity model has AUPRC \GhHThreeMethod{} \GhHThreeMethodCI{} against \GhHThreeRef{} \GhHThreeRefCI{} for Rules v2, a difference of \GhHThreeDiff{} \GhHThreeDiffCI{} (Holm $p=\GhHThreePHolm$). The gain survives removing the \GhXNoSoDDropped{} SoD conflicts from the positives (exploratory; $\Delta={}$\GhXNoSoDRDLEntDiff \GhXNoSoDRDLEntDiffCI; \cref{tab:posthoc-entity}). Link models aggregated by the maximum over candidates are weaker at entity level (GraIL-style \GhEntityGraILThreeAUPRC), mainly because they cannot see the entity-only SoD conflict.

On the entity twin pool the comparison did not establish a gain (secondary S3): RDL-style has twin AUROC \GhEntityTwinRDLEntAUROC{} against \GhEntityTwinRulesVTwoAUROC{} for the rules, a difference of \GhEntityTwinRDLEntDiff{} \GhEntityTwinRDLEntDiffCI. The entity GNN is therefore better at \emph{ranking} across all entities, which combines many weak cues, but not at reading the records of a given fraud--twin pair. The pre-specified combination Rules v2 + RDL-style (\texttt{lr}) reaches entity AUPRC \GhEntityRulesVTwoPlusRDLEntLRAUPRC{} (descriptive). Two models that were not pre-specified are higher still: the R-GCN entity model (\GhEntityRGCNEntAUPRC{} \GhEntityRGCNEntAUPRCCI) and its combination with the rules (\GhEntityRulesVTwoPlusRGCNEntLRAUPRC); we report them without inference. Low-FPR entity recall remains modest for every method: at 1\% FPR the rules recover \GhEntityRulesVTwoRFprOne{} of fraudulent entities and the RDL-style model \GhEntityRDLEntRFprOne{} ($\pm$\GhEntityRDLEntRFprOneSD{} across seeds).

% generated by xiaobo-experiment/render-results-draft.py; do not edit
\begin{table}[t]\centering\footnotesize\setlength{\tabcolsep}{3.5pt}
\caption{Entity level (2{,}496 unseen new hires and vendors, 672 planted, 26.9\% prevalence by design): the pre-specified RDL-style entity model ranks fraudulent entities above record-aware rules, but on the twin pool it has a close point estimate to the rules; combining the rules with an entity GNN helps on both views (descriptive). Link models are aggregated by the maximum over candidates. $^\dagger$GNN that is not a pre-specified primary model (descriptive).}
\label{tab:entity-main}
\begin{adjustbox}{max width=\linewidth}
\begin{tabular}{lrrrrrr}
\toprule
method & AUROC & \makecell[l]{AUPRC\\ {}[95\% CI]} & \makecell[l]{recall at\\ {}1\% FPR} & \makecell[l]{recall at\\ {}5\% FPR} & \makecell[l]{twin\\ {}AUROC} & \makecell[l]{twin $\Delta$ vs\\ {}Rules v2 [95\% CI]} \\
\midrule
Rules severity & 0.758 & \makecell[r]{0.486\\ {\scriptsize [0.475, 0.499]}} & 0.010 & 0.124 & 0.564 & \makecell[r]{$-$0.253\\ {\scriptsize [$-$0.271, $-$0.234]}} \\
Rules v2 (record-aware) & 0.872 & \makecell[r]{0.698\\ {\scriptsize [0.669, 0.728]}} & 0.104 & 0.387 & 0.817 & -- \\
Path features + LR & 0.876 & \makecell[r]{0.725\\ {\scriptsize [0.692, 0.757]}} & 0.192 & 0.463 & 0.815 & \makecell[r]{$-$0.002\\ {\scriptsize [$-$0.021, 0.018]}} \\
RDL-style (entity) & 0.890 & \makecell[r]{0.752$\pm$0.018\\ {\scriptsize [0.719, 0.784]}} & 0.197 & 0.524 & 0.821 & \makecell[r]{+0.004\\ {\scriptsize [$-$0.013, 0.023]}} \\
R-GCN (entity)$^\dagger$ & 0.900 & \makecell[r]{0.776$\pm$0.015\\ {\scriptsize [0.744, 0.808]}} & 0.233 & 0.546 & 0.821 & \makecell[r]{+0.004\\ {\scriptsize [$-$0.017, 0.025]}} \\
RDL-style (link)$^\dagger$ & 0.780 & \makecell[r]{0.638$\pm$0.016\\ {\scriptsize [0.614, 0.665]}} & 0.120 & 0.381 & 0.718 & \makecell[r]{$-$0.099\\ {\scriptsize [$-$0.114, $-$0.080]}} \\
GraIL-style $k{=}3$ & 0.782 & \makecell[r]{0.612$\pm$0.060\\ {\scriptsize [0.584, 0.641]}} & 0.100 & 0.339 & 0.631 & \makecell[r]{$-$0.186\\ {\scriptsize [$-$0.208, $-$0.162]}} \\
NBFNet-style$^\dagger$ & 0.791 & \makecell[r]{0.621$\pm$0.017\\ {\scriptsize [0.596, 0.643]}} & 0.129 & 0.309 & 0.624 & \makecell[r]{$-$0.193\\ {\scriptsize [$-$0.213, $-$0.172]}} \\
\midrule
Rules v2 + RDL-style (entity) (lr) & 0.914 & \makecell[r]{0.792$\pm$0.004\\ {\scriptsize [0.763, 0.822]}} & 0.266 & 0.567 & 0.844 & \makecell[r]{+0.027\\ {\scriptsize [0.019, 0.037]}} \\
Rules v2 + R-GCN (entity)$^\dagger$ (lr) & 0.921 & \makecell[r]{0.801$\pm$0.004\\ {\scriptsize [0.769, 0.832]}} & 0.287 & 0.603 & 0.847 & \makecell[r]{+0.030\\ {\scriptsize [0.019, 0.041]}} \\
\bottomrule
\end{tabular}
\end{adjustbox}
\end{table}

\subsection{Secondary: the strongest combination within the original method set}

The pre-specified secondary comparisons (\cref{tab:secondary}) are descriptive. The two-feature logistic combination of Rules v2 and the GraIL-style model has the highest overall link AUPRC of the \GhNLinkMethods{} original-stage link methods (\GhSOneAMethod), above Rules v2 (\GhSOneARef; $\Delta={}$\GhSOneADiff \GhSOneADiffCI) and above the GraIL-style model alone (\GhSOneBRef; $\Delta={}$\GhSOneBDiff \GhSOneBDiffCI). It inherits the rules' strength at h2 (AUPRC \GhLinkRulesVTwoPlusGraILThreeLRAUPRCHopTwo, the best h2 value of any method) but gives up part of the GNN's advantage at h3 (\GhLinkRulesVTwoPlusGraILThreeLRAUPRCHopThree{} vs.\ \GhHOneHopThreeMethod) and h4 (\GhLinkRulesVTwoPlusGraILThreeLRAUPRCHopFour{} vs.\ \GhHOneHopFourMethod); on the twin pool it reaches \GhLinkTwinRulesVTwoPlusGraILThreeLRAUROC{}, still below the rules alone.

% generated by xiaobo-experiment/render-results-draft.py; do not edit
\begin{table}[t]\centering\footnotesize\setlength{\tabcolsep}{3.5pt}
\caption{Pre-specified secondary comparisons (descriptive, outside the Holm family, unadjusted CIs): the rules + GraIL-style logistic combination has the highest overall link AUPRC within the original method set, above both of its inputs, and the RDL-style entity model has a close point estimate to the rules on the entity twin pool.}
\label{tab:secondary}
\begin{adjustbox}{max width=\linewidth}
\begin{tabular}{llllrlrrr}
\toprule
id & pool / group & metric & method & value & reference & value & difference [95\% CI] & \makecell[l]{one-sided $p$\\ {}(unadjusted)} \\
\midrule
S1a & link, all, overall & AUPRC & \makecell[l]{Rules v2 +\\ {}GraIL-style\\ {}$k{=}3$ (lr)} & 0.735 & Rules v2 & 0.347 & \makecell[r]{+0.388\\ {\scriptsize [0.364, 0.412]}} & 0.001 \\
S1b & link, all, overall & AUPRC & \makecell[l]{Rules v2 +\\ {}GraIL-style\\ {}$k{=}3$ (lr)} & 0.735 & \makecell[l]{GraIL-style\\ {}$k{=}3$} & 0.663 & \makecell[r]{+0.072\\ {\scriptsize [0.047, 0.099]}} & 0.001 \\
S2 & link, all, overall & AUPRC & \makecell[l]{GraIL-style\\ {}$k{=}3$} & 0.663 & Rules v2 & 0.347 & \makecell[r]{+0.316\\ {\scriptsize [0.274, 0.355]}} & 0.001 \\
S3 & entity, twin pool, overall & AUROC & \makecell[l]{RDL-style\\ {}(entity)} & 0.821 & Rules v2 & 0.817 & \makecell[r]{+0.004\\ {\scriptsize [$-$0.013, 0.023]}} & 0.348 \\
\bottomrule
\end{tabular}
\end{adjustbox}
\end{table}

\subsection{Further descriptive results}

\paragraph{Low-FPR link recall is saturated, as pre-specified.} \GhNLinkSaturated{} of the \GhNLinkMethods{} link methods recall at least 99\% of the positives at 1\% FPR (\cref{tab:link-full}), including the GraIL-style model (\GhLinkGraILThreeRFprOne). At 0.1\% FPR the measure is still informative: GraIL-style \GhLinkGraILThreeRFprTenth{} against \GhLinkRulesVTwoRFprTenth{} for the rules.

\paragraph{Graph heuristics.} Personalized PageRank has the highest AUROC of any non-learned method (\GhLinkPPRAUROC) but AUPRC \GhLinkPPRAUPRC: it finds the neighbourhood of the colluder but not the colluder.

\paragraph{Per audit test.} \Cref{tab:link-control} gives link AUPRC per audit test. Rules v2 is strongest on the two tests whose evidence is usually an edge the rules can read directly---ghost employee (\GhLinkRulesVTwoAUPRCCtlGhost) and approver--payee link (\GhLinkRulesVTwoAUPRCCtlApprover)---and weakest on shared payroll accounts (\GhLinkRulesVTwoAUPRCCtlSharedAcct), bank-detail changes before a run (\GhLinkRulesVTwoAUPRCCtlBankChange) and employee--vendor links (\GhLinkRulesVTwoAUPRCCtlEmpVendor), where half or more of the planted chains are 3 or 4 hops long. The GraIL-style model reverses that ordering (employee--vendor \GhLinkGraILThreeAUPRCCtlEmpVendor, ghost \GhLinkGraILThreeAUPRCCtlGhost). These per-test and per-pattern cells (\cref{tab:link-patternhop,tab:entity-pattern}) are exploratory and have different prevalences.

\paragraph{Seed variance.} The pre-specified GraIL-style model has a seed SD of \GhLinkGraILThreeAUPRCSD{} in overall link AUPRC. The organization bootstrap does not include seed variance; the reported CIs are for the seed-mean metric and are conditional on these \GhNSeeds{} seeds (\cref{sec:limitations}).

% generated by xiaobo-experiment/render-results-draft.py; do not edit
\begin{table}[t]\centering\footnotesize\setlength{\tabcolsep}{3.5pt}
\caption{Every confirmatory outcome matched the direction recorded from the dev rehearsal before the freeze (dev: 4 held-out dev organizations; test: 24 unseen test organizations).}
\label{tab:devtest}
\begin{tabular}{lllll}
\toprule
id & dev difference (point) & dev direction $>0$ & test difference [95\% CI] & test outcome \\
\midrule
H1-h2 & $-$0.034 & no & $-$0.208 [$-$0.262, $-$0.150] & not supported \\
H1-h3 & +0.405 & yes & +0.420 [0.375, 0.469] & supported \\
H1-h4 & +0.286 & yes & +0.272 [0.219, 0.337] & supported \\
H2 & $-$0.061 & no & $-$0.122 [$-$0.151, $-$0.094] & not supported \\
H3 & +0.087 & yes & +0.053 [0.025, 0.086] & supported \\
\bottomrule
\end{tabular}
\end{table}

\FloatBarrier

\section{Exploratory Post-hoc Analyses}
\label{sec:posthoc}

Every analysis in this section is exploratory and was not pre-specified. The analyses were run after the confirmatory report, on the frozen score files of the confirmatory run and the test-organization ground truth, with no model trained or re-run, and none replaces a pre-specified result; all CIs here are unadjusted. Recomputing the confirmatory values from the score files reproduced the report to within \GhXSanityMaxDev{} in AUPRC.

\paragraph{Seeds.} For each of the \GhXNSeedHyp{} pre-specified tests, the difference has the same sign for every GNN seed as for the seed mean (\cref{tab:posthoc-seeds}); for example, H1-h4 ranges from \GhXHOneHopFourSeedDiffMin{} to \GhXHOneHopFourSeedDiffMax{} across seeds and H2 from \GhXHTwoSeedDiffMin{} to \GhXHTwoSeedDiffMax.

\paragraph{Reach without the twins.} Removing the matched twins of all hops from the negatives leaves a pool in which a method has to find the colluder among all other candidate pairs, including the other hand-built hard negatives, but never has to order a fraud against its twin. There the GraIL-style model reaches link AUPRC \GhXReachGraILThreeHopTwo, \GhXReachGraILThreeHopThree{} and \GhXReachGraILThreeHopFour{} at hops 2, 3 and 4, against \GhXReachRulesVTwoHopTwo, \GhXReachRulesVTwoHopThree{} and \GhXReachRulesVTwoHopFour{} for Rules v2 (paired differences \GhXReachGraILThreeHopTwoDiff{} \GhXReachGraILThreeHopTwoDiffCI, \GhXReachGraILThreeHopThreeDiff{} \GhXReachGraILThreeHopThreeDiffCI{} and \GhXReachGraILThreeHopFourDiff{} \GhXReachGraILThreeHopFourDiffCI; unadjusted). The subgraph and path GNNs find nearly every planted colluder at hops 2--4 (GraIL-style $k=2$ excepted at 4 hops, outside its radius), so their loss on H1-h2 and their sub-ceiling AUPRC at 3 and 4 hops come from ordering frauds against their twins. At hop 1 the rules lead (reach \GhXReachRulesVTwoHopOne{} against \GhXReachGraILThreeHopOne{} for the GraIL-style model). Two caveats apply: Rules v2's reach values still include its record exemptions, so they mix reach and reading; and because a twin copies its fraud's planting side-effects, near-perfect GNN reach may partly reflect chain shapes that are rare in the generated background rather than a general ability to follow multi-hop relations. Path features + LR, limited to length-3 paths, reaches \GhXReachPathLRHopThree{} at h3 and \GhXReachPathLRHopFour{} at h4.

\paragraph{More look-alikes per fraud.} The generator plants one benign twin per fraud (look-alike ratio \GhLookalikeRatio). Counting each twin of every hop $w$ times among the negatives is a deterministic stress test for a higher ratio: the duplicated twins keep their scores exactly, which is not the same as sampling new look-alikes. Newly sampled look-alikes would also vary in their records, so the proxy may be optimistic or pessimistic for a given method; its direction is unknown. At $w=\GhXWMax$ the GraIL-style model's AUPRC falls to \GhXTwGraILThreeHopThreeWEight{} at h3 and \GhXTwGraILThreeHopFourWEight{} at h4, and Rules v2 to \GhXTwRulesVTwoHopThreeWEight{} and \GhXTwRulesVTwoHopFourWEight{} (\cref{tab:posthoc-link}). The sign of the H1-h3 and H1-h4 differences is unchanged at every weight, but the absolute AUPRC values of the confirmatory test depend strongly on the planted ratio and should not be read as deployment precision.

\paragraph{Lower entity base rates.} Fraud prevalence among the new entities is \GhXObsPrevPct\% by design, and \GhXNegLookalikePct\% of the \GhXNegTotal{} entity negatives are matched twins (\GhXNegTwins) or other hand-built hard negatives (\GhXNegOtherHard); only \GhXNegOrdinary{} are ordinary. We lowered the prevalence in two ways (\cref{tab:posthoc-entity}). Scaling all negatives keeps that look-alike share and supplies one fixed-score scenario: at 1\% prevalence entity AUPRC falls to \GhXBaseRulesVTwoPrevOne{} for Rules v2 and \GhXBaseRDLEntPrevOne{} for the RDL-style model. Scaling only ordinary negatives supplies a different mixture scenario: at 1\% the rules keep \GhXBaseOrdRulesVTwoPrevOne{} and the RDL-style model has \GhXBaseOrdRDLEntPrevOne, and at 0.1\% \GhXBaseOrdRulesVTwoPrevTenth{} against \GhXBaseOrdRDLEntPrevTenth. The entity-level ranking of H3 therefore depends on what the additional benign entities look like; these scenarios are not estimates of real rates or mixtures.

\paragraph{Within-pair concordance.} The cleanest view of record reading compares each planted fraud with its own twin. The share of frauds scored above their twin is \GhXConcLinkRulesVTwo{} \GhXConcLinkRulesVTwoCI{} for Rules v2 and \GhXConcLinkGraILThree{} \GhXConcLinkGraILThreeCI{} for the GraIL-style model at link level ($\Delta={}$\GhXConcLinkGraILThreeDiff{} \GhXConcLinkGraILThreeDiffCI), and every single GNN is below the rules (\cref{tab:posthoc-concordance}); at entity level the observed estimates are close, without an equivalence test (\GhXConcEntRDLEnt{} against \GhXConcEntRulesVTwo). This agrees with H2 and S3.

\paragraph{H3 without SoD conflicts.} Removing the \GhXNoSoDDropped{} SoD conflicts but keeping their twins as negatives gives an RDL-style gain of \GhXNoSoDRDLEntDiff{} \GhXNoSoDRDLEntDiffCI{}; removing the twins as well gives \GhXNoSoDTwRDLEntDiff{} \GhXNoSoDTwRDLEntDiffCI{} (\cref{tab:posthoc-entity} shows the first variant).

% generated by xiaobo-experiment/render-results-draft.py; do not edit
\begin{table}[htbp]\centering\footnotesize\setlength{\tabcolsep}{3.5pt}
\caption{Exploratory, post hoc: with the matched twins of all hops removed from the negatives (``reach''; the other hand-built hard negatives stay), the GraIL-style $k{=}3$, NBFNet-style and opening-subgraph models find almost every planted colluder at hops 2--4 (GraIL-style $k{=}2$ cannot see 4 hops) but not at hop 1, while one-step Rules v2 finds mainly 1- and 2-hop colluders. When each twin of every hop counts $w$ times among the negatives (a deterministic stress test: duplicated twins keep their scores) the AUPRC of these models falls steeply, but their lead over the rules at 3 and 4 hops keeps its sign. $w{=}1$ is the confirmatory pool; GNN values are seed means; no CIs. $^\dagger$GNN that is not a pre-specified primary model (descriptive).}
\label{tab:posthoc-link}
\begin{adjustbox}{max width=\linewidth}
\begin{tabular}{lrrrrrrrrrrr}
\toprule
& \multicolumn{5}{c}{reach (twins removed)} & \multicolumn{3}{c}{h3, twin weight $w$} & \multicolumn{3}{c}{h4, twin weight $w$} \\ \cmidrule(lr){2-6}\cmidrule(lr){7-9}\cmidrule(lr){10-12} method & h1 & h2 & h3 & h4 & all & 1 & 4 & 8 & 1 & 4 & 8 \\
\midrule
Rules v2 & 0.833 & 0.854 & 0.180 & 0.000 & 0.431 & 0.081 & 0.032 & 0.018 & 0.000 & 0.000 & 0.000 \\
PPR & 0.076 & 0.053 & 0.041 & 0.006 & 0.103 & 0.037 & 0.028 & 0.021 & 0.006 & 0.005 & 0.004 \\
Path features + LR & 0.081 & 0.243 & 0.336 & 0.001 & 0.380 & 0.239 & 0.129 & 0.081 & 0.001 & 0.001 & 0.001 \\
GraIL-style $k{=}2$$^\dagger$ & 0.487 & 0.965 & 0.984 & 0.003 & 0.809 & 0.392 & 0.143 & 0.077 & 0.002 & 0.002 & 0.002 \\
GraIL-style $k{=}3$ & 0.562 & 0.982 & 0.995 & 0.989 & 0.981 & 0.501 & 0.215 & 0.126 & 0.272 & 0.093 & 0.050 \\
NBFNet-style$^\dagger$ & 0.778 & 0.991 & 0.986 & 0.958 & 0.989 & 0.461 & 0.221 & 0.148 & 0.154 & 0.045 & 0.023 \\
OSFA-insp.$^\dagger$ & 0.543 & 0.956 & 0.963 & 0.971 & 0.972 & 0.394 & 0.148 & 0.083 & 0.197 & 0.059 & 0.030 \\
RDL-style (link)$^\dagger$ & 0.014 & 0.080 & 0.021 & 0.011 & 0.084 & 0.021 & 0.019 & 0.018 & 0.011 & 0.010 & 0.009 \\
Rules v2 + GraIL-style $k{=}3$ (lr) & 0.816 & 0.996 & 0.995 & 0.988 & 0.990 & 0.437 & 0.166 & 0.091 & 0.190 & 0.057 & 0.030 \\
\bottomrule
\end{tabular}
\end{adjustbox}
\end{table}

% generated by xiaobo-experiment/render-results-draft.py; do not edit
\begin{table}[htbp]\centering\footnotesize\setlength{\tabcolsep}{3.5pt}
\caption{Exploratory, post hoc: every pre-specified difference has the same sign for each of the 3 GNN seeds as for the seed mean (point values only; per-seed tests were not run). Confirmatory test pool, recomputed from the frozen score files; Rules v2 is deterministic.}
\label{tab:posthoc-seeds}
\begin{adjustbox}{max width=\linewidth}
\begin{tabular}{llrrrrrrrr}
\toprule
& & & \multicolumn{3}{c}{model, seed} & \multicolumn{3}{c}{difference, seed} & \\ \cmidrule(lr){4-6}\cmidrule(lr){7-9} test & metric & Rules v2 & 9131 & 9132 & 9133 & 9131 & 9132 & 9133 & seed mean \\
\midrule
H1-h2 & AUPRC & 0.650 & 0.521 & 0.392 & 0.412 & $-$0.129 & $-$0.258 & $-$0.238 & $-$0.208 \\
H1-h3 & AUPRC & 0.081 & 0.504 & 0.427 & 0.571 & +0.424 & +0.346 & +0.490 & +0.420 \\
H1-h4 & AUPRC & 0.000 & 0.379 & 0.130 & 0.308 & +0.378 & +0.130 & +0.307 & +0.272 \\
H2 & AUROC & 0.796 & 0.722 & 0.611 & 0.689 & $-$0.074 & $-$0.185 & $-$0.108 & $-$0.122 \\
H3 & AUPRC & 0.698 & 0.747 & 0.771 & 0.737 & +0.049 & +0.073 & +0.039 & +0.053 \\
\bottomrule
\end{tabular}
\end{adjustbox}
\end{table}

% generated by xiaobo-experiment/render-results-draft.py; do not edit
\begin{table}[htbp]\centering\footnotesize\setlength{\tabcolsep}{3.5pt}
\caption{Exploratory, post hoc: entity-level AUPRC at lower fraud prevalence depends on what the added negatives look like. Scaling all negatives (hard-negative mix fixed: 67\% twins and hand-built hard negatives) is a pessimistic stress test under which AP decreases for every listed method; scaling only the 600 ordinary negatives keeps most of the rules' AUPRC and reverses the ranking in their favour. The RDL-style gain over Rules v2 survives removing the 96 SoD conflicts (control deficiencies). Observed prevalence 26.9\%; base-rate columns have no CIs; brackets are unadjusted org-cluster bootstrap CIs. $^\dagger$GNN that is not a pre-specified primary model (descriptive).}
\label{tab:posthoc-entity}
\begin{adjustbox}{max width=\linewidth}
\begin{tabular}{lrrrrrrr}
\toprule
& & \multicolumn{2}{c}{all negatives scaled} & \multicolumn{2}{c}{ordinary scaled} & \multicolumn{2}{c}{without SoD conflicts} \\ \cmidrule(lr){3-4}\cmidrule(lr){5-6}\cmidrule(lr){7-8} method & observed & 1\% & 0.1\% & 1\% & 0.1\% & AUPRC [95\% CI] & $\Delta$ vs Rules v2 \\
\midrule
Rules severity & 0.486 & 0.027 & 0.003 & -- & -- & -- & -- \\
Rules v2 & 0.698 & 0.089 & 0.012 & 0.538 & 0.518 & \makecell[r]{0.667\\ {\scriptsize [0.633, 0.700]}} & -- \\
Path features + LR & 0.725 & 0.092 & 0.012 & 0.389 & 0.282 & \makecell[r]{0.681\\ {\scriptsize [0.642, 0.717]}} & \makecell[r]{+0.014\\ {\scriptsize [$-$0.015, 0.047]}} \\
RDL-style (entity) & 0.752 & 0.119 & 0.028 & 0.393 & 0.269 & \makecell[r]{0.719\\ {\scriptsize [0.681, 0.754]}} & \makecell[r]{+0.052\\ {\scriptsize [0.021, 0.088]}} \\
R-GCN (entity)$^\dagger$ & 0.776 & 0.147 & 0.022 & 0.418 & 0.244 & \makecell[r]{0.740\\ {\scriptsize [0.704, 0.777]}} & \makecell[r]{+0.074\\ {\scriptsize [0.041, 0.111]}} \\
Rules v2 + RDL-style (entity) (lr) & 0.792 & 0.132 & 0.020 & 0.590 & 0.481 & \makecell[r]{0.766\\ {\scriptsize [0.730, 0.800]}} & \makecell[r]{+0.099\\ {\scriptsize [0.082, 0.116]}} \\
Rules v2 + R-GCN (entity)$^\dagger$ (lr) & 0.801 & 0.134 & 0.018 & 0.606 & 0.498 & -- & -- \\
\bottomrule
\end{tabular}
\end{adjustbox}
\end{table}

% generated by xiaobo-experiment/render-results-draft.py; do not edit
\begin{table}[htbp]\centering\footnotesize\setlength{\tabcolsep}{3.5pt}
\caption{Exploratory, post hoc: within-pair concordance, the share of planted frauds scored above their own matched twin (ties count half; 576 link pairs, 672 entity pairs). At link level every single GNN is below the generator-aware Rules v2 and the rules + GraIL-style combination has a close point estimate; at entity level GNN point estimates are close to the rules and the combinations are above them. Seed means; brackets are unadjusted org-cluster bootstrap CIs. $^\dagger$GNN that is not a pre-specified primary model (descriptive).}
\label{tab:posthoc-concordance}
\begin{adjustbox}{max width=\linewidth}
\begin{tabular}{lrrrrrr}
\toprule
method & \makecell[l]{concordance\\ {}[95\% CI]} & h1 & h2 & h3 & h4 & \makecell[l]{$\Delta$ vs Rules v2\\ {}[95\% CI]} \\
\midrule
Rules v2 & \makecell[r]{0.820\\ {\scriptsize [0.792, 0.847]}} & 0.885 & 0.839 & 0.796 & 0.812 & -- \\
PPR & \makecell[r]{0.503\\ {\scriptsize [0.470, 0.536]}} & 0.479 & 0.479 & 0.525 & 0.510 & \makecell[r]{$-$0.317\\ {\scriptsize [$-$0.354, $-$0.280]}} \\
Path features + LR & \makecell[r]{0.735\\ {\scriptsize [0.701, 0.770]}} & 0.792 & 0.698 & 0.692 & 0.891 & \makecell[r]{$-$0.085\\ {\scriptsize [$-$0.115, $-$0.051]}} \\
RDL-style (link)$^\dagger$ & \makecell[r]{0.759\\ {\scriptsize [0.736, 0.786]}} & 0.882 & 0.733 & 0.733 & 0.816 & \makecell[r]{$-$0.061\\ {\scriptsize [$-$0.087, $-$0.032]}} \\
GraIL-style $k{=}2$$^\dagger$ & \makecell[r]{0.684\\ {\scriptsize [0.650, 0.718]}} & 0.764 & 0.639 & 0.678 & 0.750 & \makecell[r]{$-$0.136\\ {\scriptsize [$-$0.168, $-$0.104]}} \\
GraIL-style $k{=}3$ & \makecell[r]{0.728\\ {\scriptsize [0.694, 0.762]}} & 0.812 & 0.665 & 0.725 & 0.819 & \makecell[r]{$-$0.092\\ {\scriptsize [$-$0.119, $-$0.066]}} \\
NBFNet-style$^\dagger$ & \makecell[r]{0.697\\ {\scriptsize [0.666, 0.726]}} & 0.861 & 0.715 & 0.694 & 0.583 & \makecell[r]{$-$0.124\\ {\scriptsize [$-$0.155, $-$0.090]}} \\
OSFA-insp.$^\dagger$ & \makecell[r]{0.696\\ {\scriptsize [0.663, 0.730]}} & 0.889 & 0.646 & 0.672 & 0.760 & \makecell[r]{$-$0.124\\ {\scriptsize [$-$0.149, $-$0.099]}} \\
Rules v2 + GraIL-style $k{=}3$ (lr) & \makecell[r]{0.819\\ {\scriptsize [0.786, 0.853]}} & 0.875 & 0.830 & 0.790 & 0.844 & \makecell[r]{$-$0.001\\ {\scriptsize [$-$0.019, 0.018]}} \\
\midrule
entity: Rules v2 & \makecell[r]{0.830\\ {\scriptsize [0.800, 0.856]}} & 0.885 & 0.836 & 0.792 & 0.812 & -- \\
entity: Path features + LR & \makecell[r]{0.857\\ {\scriptsize [0.827, 0.882]}} & 0.854 & 0.810 & 0.854 & 0.922 & \makecell[r]{+0.028\\ {\scriptsize [0.005, 0.053]}} \\
entity: RDL-style (entity) & \makecell[r]{0.831\\ {\scriptsize [0.804, 0.860]}} & 0.861 & 0.788 & 0.829 & 0.875 & \makecell[r]{+0.002\\ {\scriptsize [$-$0.024, 0.029]}} \\
entity: R-GCN (entity)$^\dagger$ & \makecell[r]{0.845\\ {\scriptsize [0.818, 0.872]}} & 0.889 & 0.809 & 0.843 & 0.896 & \makecell[r]{+0.016\\ {\scriptsize [$-$0.003, 0.036]}} \\
entity: Rules v2 + RDL-style (entity) (lr) & \makecell[r]{0.858\\ {\scriptsize [0.829, 0.883]}} & 0.882 & 0.828 & 0.853 & 0.889 & \makecell[r]{+0.029\\ {\scriptsize [0.013, 0.045]}} \\
entity: Rules v2 + R-GCN (entity)$^\dagger$ (lr) & \makecell[r]{0.865\\ {\scriptsize [0.839, 0.891]}} & 0.903 & 0.847 & 0.856 & 0.896 & \makecell[r]{+0.035\\ {\scriptsize [0.024, 0.047]}} \\
\bottomrule
\end{tabular}
\end{adjustbox}
\end{table}

\section{Discussion}
\label{sec:discussion}

\subsection{Multi-hop discovery is an information-access advantage}
The comparison separates direct-record checks from methods that combine several relations. In the original study, GraIL-style exceeded one-step rules at three and four hops. The extension supplied the non-GNN control: a chained-join query reached four-hop candidates, and tuned boosting on typed paths exceeded the evaluated GNNs. At four hops its AP was \GvV{A4.ref}, versus \GvV{A4.method} for GraIL-style. The original contrast therefore identifies multi-hop access relative to the implemented one-step rules, rather than a benefit unique to neural message passing.

Both graph and relational-query methods can follow records linked through account holders, representatives, contacts and identifiers. GADBench's tree-ensemble result~\citep{tang2023gadbench} concerns anomalous nodes in other datasets; the present evidence adds organization-held-out relations and matched benign counterparts. The contribution is the task, controls and decomposition of measured behavior, not a universal algorithm-family ranking.

\subsection{Record discrimination is separate from finding a chain}
On matched twins, the original GraIL-style model had AUROC \GhHTwoMethod{} against \GhHTwoRef{} for generator-informed Rules v2. With generic rules as reference, the difference was \GvV{A2.diff} with interval \GvV{A2.diffci}. This comparison did not establish a difference in either direction; no equivalence margin or equivalence test was specified.

Rule specification matters, but the comparison does not isolate analyst knowledge: weights, exemptions and logic also change, and both rule sets were authored with knowledge of the generator. The extension protocol retained explanations beyond generator knowledge for this outcome. In the full method set, boosting achieved twin AUROC \GvV{A.link.cj-gbdt.twins:auroc}, above Rules v2's \GvV{A.link.rules-v2.twins:auroc}. Thus rules are not necessarily the best record-reading layer; connectivity alone is insufficient, and multiple method classes can use record features.

\subsection{Review priorities depend on the cohort}
At the new cohort's \GvV{B.prevpct}\% positive rate, RDL-style achieved entity AP \GvV{B1.method} against \GvV{B1.ref} for Rules v2. At a 2\% review budget, their precisions were \GvV{B2.method} and \GvV{B2.ref}; boosting achieved \GvV{B.cj-gbdt-ent.p@budget0.02}. These operating points connect ranking to a defined review workload rather than equating AUROC with a useful shortlist.

The cohorts differ in organization size, new-entity counts, benign composition and training data as well as prevalence. Their comparison does not identify prevalence as the sole cause of the changed gap. Re-weighting holds written scores fixed; the new cohort evaluates methods retrained on a different synthetic population. Neither estimates real payroll-fraud prevalence or deployment precision.

\subsection{Identity noise changes the useful representation}
Failed joins weaken explicit paths because a missing join can interrupt an evidence chain. At 40\% split noise, boosting's four-hop AP declined from \GvV{C.clean.link.cj-gbdt.auprc|hop:4} to \GvV{C.idsplit-40.link.cj-gbdt.auprc|hop:4}. The entity GNN retained more entity-ranking performance: AP \GvV{C.idsplit-40.entity.rdl-ent.auprc}, versus boosting's \GvV{C.idsplit-40.entity.cj-gbdt-ent.auprc}. This is a descriptive contrast under the specified perturbation, not superiority under every identity-error process.

Explicit paths are strong on the resolved graphs evaluated here; node-wise aggregation supplies a complementary reference when joins fail. The perturbations are splits and missing records, not false merges, free-text matching or an estimated operational error distribution.

\subsection{Practical implications and transferable audit checks}
Method selection should compare candidate retrieval, discrimination using control records, and ranking under an actual review budget. A cascade using different methods for retrieval and final assessment is a possible design, but was not evaluated end to end here.

For benchmark construction, attribute-free audits, matched benign controls, univariate screens and metric-saturation checks make shortcuts visible. The v1-to-v2 audit demonstrates one correction: initial high scores reflected the planting procedure, while the revised twin pool requires control-record discrimination. Independent generators and case data would test the transfer of that diagnostic pattern.

\section{Study Scope, Reproducibility and Responsible Use}
\label{sec:limitations}

\paragraph{Population and target.}
Both cohorts are synthetic outputs of the same HR/ERP generator. Testing on new organizations measures transfer within that mechanism, not to a new schema or an independently sampled payroll population. Positive labels denote planted cases; the original entity task also includes SoD control deficiencies, with an exploratory exclusion analysis. The \GhEntityPrevalencePct\% and \GvV{B.prevpct}\% positive rates and one-twin-per-case ratio are study parameters, not estimates of real-world prevalence. Operational misconduct determinations require evidence beyond a screening score.

\paragraph{Implementation and selection.}
The GNNs are the feature-augmented style implementations specified in \cref{sec:methods}, not original-system reproductions. Their hyperparameters remained fixed from generator-v1 development; extension GBDTs were validation-tuned. Results compare these implemented procedures, not algorithm families under equal tuning effort. Rules v2 is a generator-informed reference, not an upper bound on other analysts or rule systems. Typed path features also use schema knowledge.

\paragraph{Evidence and uncertainty.}
\Cref{tab:study-lineage} separates original testing, exploratory analyses, reused-test extensions and the new synthetic cohort. Both prospective protocols are internal freezes; no public registry identifier is asserted. All five original and nine extension decisions are retained. Organization-bootstrap intervals condition on fitted models and the \GhNSeeds{} recorded seeds; they do not include all training or generator variability. GraIL-style's overall seed SD was \GhLinkGraILThreeAUPRCSD. Non-family intervals and per-pattern cells are descriptive. The recorded bootstrap-tail decision convention is approximate, not an exact randomization test; Holm adjustment does not independently validate its calibration. Unsupported directional hypotheses do not establish equivalence.

\paragraph{Perturbations and operating points.}
The identity experiments model failed joins and missing records, not false merges or a measured operational linkage process~\citep{fellegi1969recordlinkage}. Noise results use fixed perturbation realizations. Budget precision describes the synthetic cohort and tie-handling rule, not projected deployment yield. Re-weighted AP is a fixed-score sensitivity analysis rather than a newly sampled cohort.

\paragraph{Run provenance.}
Original and extension runs verified code/data checksums and wrote separate outputs. The archived history distinguishes design changes, superseded exports, exploratory work and test reuse (\cref{sec:deviations}). Changing CPU threads altered retrained NBFNet-style scores (clean AP \GvV{C.repro.nbfnet.retrained} versus \GvV{C.repro.nbfnet.frozen}); the RDL-style check reproduced its written scores. Original outputs remain unchanged. This supports an auditable specification, without asserting bitwise identity across all training environments.

\paragraph{Responsible application.}
No real employee or customer data are used; no person is evaluated for hiring, promotion or performance. Shared addresses, family contacts and joint accounts can be legitimate. Operational applications should treat signals as review leads, retain human decisions, protect personal data and validate population-specific error rates. The contribution is a controlled comparison and audit procedure, not an automated finding of wrongdoing.

\clearpage
\section{Conclusion}
\label{sec:conclusion}

We evaluated inductive payroll-fraud screening through employee--payee link discovery, record discrimination and review prioritization. The synthetic HR/ERP design combines unseen organizations, complete planted labels, typed evidence chains and benign twins, with organization-level paired analyses and preserved stage provenance.

On the resolved graphs, a tuned GBDT on chained-join features achieved the strongest overall link ranking (AP \GvV{A.link.cj-gbdt.auprc}, versus \GvV{A.link.grail-k3.auprc} for GraIL-style) and highest measured twin discrimination. The original GNN advantage over one-step rules at three and four hops is a multi-hop evidence advantage, not a GNN-specific result. On the new cohort with \GvV{B.prevpct}\% positive entities, boosting again led; the entity GNN also exceeded record-aware rules in AP (\GvV{B1.method} versus \GvV{B1.ref}) and precision at a 2\% review budget (\GvV{B2.method} versus \GvV{B2.ref}). At 40\% failed joins, that GNN ranked above boosting in the evaluated condition (AP \GvV{C.idsplit-40.entity.rdl-ent.auprc} versus \GvV{C.idsplit-40.entity.cj-gbdt-ent.auprc}, descriptively).

A graph benchmark must distinguish finding a planted chain from interpreting the records that make it suspicious. Structure-only audits exposed an initial shortcut; matched twins supplied a direct control for that channel. These results provide reproducible evidence for comparing relational screening under explicit data, training and review conditions. Balanced tuning, independently sourced cases and richer identity-error processes are the next validation targets.

\paragraph{Artifact availability.}
Generators, synthetic datasets and run records are available from the authors on request.

\bibliographystyle{plainnat}
\bibliography{references}

@inproceedings{teru2020grail,
  title         = {{Inductive Relation Prediction by Subgraph Reasoning}},
  author        = {Teru, Komal K. and Denis, Etienne and Hamilton, William L.},
  year          = {2020},
  eprint        = {1911.06962},
  archivePrefix = {arXiv},
  primaryClass  = {cs.LG},
  url           = {https://arxiv.org/abs/1911.06962},
  booktitle     = {Proceedings of the International Conference on Machine Learning}
}

@inproceedings{zhu2021nbfnet,
  title         = {{Neural Bellman-Ford Networks: A General Graph Neural Network Framework for Link Prediction}},
  author        = {Zhu, Zhaocheng and Zhang, Zuobai and Xhonneux, Louis-Pascal and Tang, Jian},
  year          = {2021},
  eprint        = {2106.06935},
  archivePrefix = {arXiv},
  primaryClass  = {cs.LG},
  url           = {https://arxiv.org/abs/2106.06935},
  booktitle     = {Advances in Neural Information Processing Systems}
}

@inproceedings{galkin2024ultra,
  title         = {{Towards Foundation Models for Knowledge Graph Reasoning}},
  author        = {Galkin, Mikhail and Yuan, Xinyu and Mostafa, Hesham and Tang, Jian and Zhu, Zhaocheng},
  year          = {2024},
  eprint        = {2310.04562},
  archivePrefix = {arXiv},
  primaryClass  = {cs.CL},
  url           = {https://arxiv.org/abs/2310.04562},
  booktitle     = {Proceedings of the International Conference on Learning Representations}
}

@inproceedings{lee2023ingram,
  title         = {{InGram: Inductive Knowledge Graph Embedding via Relation Graphs}},
  author        = {Lee, Jaejun and Chung, Chanyoung and Whang, Joyce Jiyoung},
  year          = {2023},
  eprint        = {2305.19987},
  archivePrefix = {arXiv},
  primaryClass  = {cs.LG},
  url           = {https://arxiv.org/abs/2305.19987},
  booktitle     = {Proceedings of the International Conference on Machine Learning}
}

@inproceedings{zhang2022redgnn,
  title         = {{Knowledge Graph Reasoning with Relational Digraph}},
  author        = {Zhang, Yongqi and Yao, Quanming},
  year          = {2022},
  eprint        = {2108.06040},
  archivePrefix = {arXiv},
  primaryClass  = {cs.AI},
  url           = {https://arxiv.org/abs/2108.06040},
  booktitle     = {Proceedings of The Web Conference}
}

@misc{xie2024onesubgraph,
  title         = {{One Subgraph for All: Efficient Reasoning on Opening Subgraphs for Inductive Knowledge Graph Completion}},
  author        = {Xie, Zhiwen and Zhang, Yi and Zhou, Guangyou and Liu, Jin and Tu, Xinhui and Huang, Jimmy Xiangji},
  year          = {2024},
  eprint        = {2404.15807},
  archivePrefix = {arXiv},
  primaryClass  = {cs.CL},
  url           = {https://arxiv.org/abs/2404.15807}
}

@misc{sajja2026travelfraudbench,
  title         = {{TravelFraudBench: A Configurable Evaluation Framework for GNN Fraud Ring Detection in Travel Networks}},
  author        = {Sajja, Bhavana},
  year          = {2026},
  eprint        = {2604.21093},
  archivePrefix = {arXiv},
  primaryClass  = {cs.LG},
  url           = {https://arxiv.org/abs/2604.21093}
}

@misc{tritscher2022openerp,
  title         = {{Open ERP System Data For Occupational Fraud Detection}},
  author        = {Tritscher, Julian and Gwinner, Fabian and Schl{\"o}r, Daniel and Krause, Anna and Hotho, Andreas},
  year          = {2022},
  eprint        = {2206.04460},
  archivePrefix = {arXiv},
  primaryClass  = {cs.AI},
  url           = {https://arxiv.org/abs/2206.04460}
}

@misc{cheng2024gnnfraudreview,
  title         = {{Graph Neural Networks for Financial Fraud Detection: A Review}},
  author        = {Cheng, Dawei and Zou, Yao and Xiang, Sheng and Jiang, Changjun},
  year          = {2024},
  eprint        = {2411.05815},
  archivePrefix = {arXiv},
  primaryClass  = {q-fin.ST},
  url           = {https://arxiv.org/abs/2411.05815},
  note          = {Frontiers of Computer Science 2025}
}

@misc{altman2023amlworld,
  title         = {{Realistic Synthetic Financial Transactions for Anti-Money Laundering Models}},
  author        = {Altman, Erik and Blanu{\v{s}}a, Jovan and von Nieder{\"a}usern, Luc and Egressy, B{\'e}ni and Anghel, Andreea and Atasu, Kubilay},
  year          = {2023},
  eprint        = {2306.16424},
  archivePrefix = {arXiv},
  primaryClass  = {cs.AI},
  url           = {https://arxiv.org/abs/2306.16424}
}

@misc{samudrala2026seqfct,
  title         = {{ERP Data Provisioning Financial Control Testing}},
  author        = {Samudrala, Anitha},
  year          = {2026},
  eprint        = {2607.09712},
  archivePrefix = {arXiv},
  primaryClass  = {cs.LG},
  url           = {https://arxiv.org/abs/2607.09712}
}

@misc{mishra2026erpriskbench,
  title         = {{ERP-RiskBench: Leakage-Safe Ensemble Learning for Financial Risk}},
  author        = {Mishra, Sanjay},
  year          = {2026},
  eprint        = {2603.06671},
  archivePrefix = {arXiv},
  primaryClass  = {cs.LG},
  url           = {https://arxiv.org/abs/2603.06671}
}

@misc{kadir2025auditcopilot,
  title         = {{AuditCopilot: Leveraging LLMs for Fraud Detection in Double-Entry Bookkeeping}},
  author        = {Kadir, Md Abdul and Macharla Vasu, Sai Suresh and Nair, Sidharth S. and Sonntag, Daniel},
  year          = {2025},
  eprint        = {2512.02726},
  archivePrefix = {arXiv},
  primaryClass  = {cs.AI},
  url           = {https://arxiv.org/abs/2512.02726}
}

@misc{huang2025mled,
  title         = {{Can LLMs Find Fraudsters? Multi-level LLM Enhanced Graph Fraud Detection}},
  author        = {Huang, Tairan and Wang, Yili and Li, Qiutong and He, Changlong and Gao, Jianliang},
  year          = {2025},
  eprint        = {2507.11997},
  archivePrefix = {arXiv},
  primaryClass  = {cs.LG},
  url           = {https://arxiv.org/abs/2507.11997},
  note          = {ACM MM 2025}
}

@misc{robinson2024relbench,
  title         = {{RelBench: A Benchmark for Deep Learning on Relational Databases}},
  author        = {Robinson, Joshua and Ranjan, Rishabh and Hu, Weihua and Huang, Kexin and Han, Jiaqi and Dobles, Alejandro and Fey, Matthias and Lenssen, Jan E. and Yuan, Yiwen and Zhang, Zecheng and others},
  year          = {2024},
  eprint        = {2407.20060},
  archivePrefix = {arXiv},
  primaryClass  = {cs.LG},
  url           = {https://arxiv.org/abs/2407.20060}
}

@misc{fey2023rdl,
  title         = {{Relational Deep Learning: Graph Representation Learning on Relational Databases}},
  author        = {Fey, Matthias and Hu, Weihua and Huang, Kexin and Lenssen, Jan Eric and Ranjan, Rishabh and Robinson, Joshua and Ying, Rex and You, Jiaxuan and Leskovec, Jure},
  year          = {2023},
  eprint        = {2312.04615},
  archivePrefix = {arXiv},
  primaryClass  = {cs.LG},
  url           = {https://arxiv.org/abs/2312.04615}
}

@misc{weber2019elliptic,
  title         = {{Anti-Money Laundering in Bitcoin: Experimenting with Graph Convolutional Networks for Financial Forensics}},
  author        = {Weber, Mark and Domeniconi, Giacomo and Chen, Jie and Weidele, Daniel Karl I. and Bellei, Claudio and Robinson, Tom and Leiserson, Charles E.},
  year          = {2019},
  eprint        = {1908.02591},
  archivePrefix = {arXiv},
  primaryClass  = {cs.SI},
  url           = {https://arxiv.org/abs/1908.02591},
  note          = {Tutorial in the Anomaly Detection in Finance Workshop at the 25th SIGKDD Conference on Knowledge Discovery and Data Mining}
}

@misc{schreyer2017erp,
  title         = {{Detection of Anomalies in Large Scale Accounting Data using Deep Autoencoder Networks}},
  author        = {Schreyer, Marco and Sattarov, Timur and Borth, Damian and Dengel, Andreas and Reimer, Bernd},
  year          = {2017},
  eprint        = {1709.05254},
  archivePrefix = {arXiv},
  primaryClass  = {cs.LG},
  url           = {https://arxiv.org/abs/1709.05254}
}

@misc{schreyer2021multiview,
  title         = {{Multi-view Contrastive Self-Supervised Learning of Accounting Data Representations for Downstream Audit Tasks}},
  author        = {Schreyer, Marco and Sattarov, Timur and Borth, Damian},
  year          = {2021},
  eprint        = {2109.11201},
  archivePrefix = {arXiv},
  primaryClass  = {cs.LG},
  url           = {https://arxiv.org/abs/2109.11201}
}

@misc{schlichtkrull2018rgcn,
  title         = {{Modeling Relational Data with Graph Convolutional Networks}},
  author        = {Schlichtkrull, Michael and Kipf, Thomas N. and Bloem, Peter and van den Berg, Rianne and Titov, Ivan and Welling, Max},
  year          = {2017},
  eprint        = {1703.06103},
  archivePrefix = {arXiv},
  primaryClass  = {stat.ML},
  url           = {https://arxiv.org/abs/1703.06103},
  note          = {ESWC 2018}
}

@misc{zhang2018seal,
  title         = {{Link Prediction Based on Graph Neural Networks}},
  author        = {Zhang, Muhan and Chen, Yixin},
  year          = {2018},
  eprint        = {1802.09691},
  archivePrefix = {arXiv},
  primaryClass  = {cs.LG},
  url           = {https://arxiv.org/abs/1802.09691},
  note          = {NIPS 2018}
}

@misc{zhang2021labeling,
  title         = {{Labeling Trick: A Theory of Using Graph Neural Networks for Multi-Node Representation Learning}},
  author        = {Zhang, Muhan and Li, Pan and Xia, Yinglong and Wang, Kai and Jin, Long},
  year          = {2020},
  eprint        = {2010.16103},
  archivePrefix = {arXiv},
  primaryClass  = {cs.LG},
  url           = {https://arxiv.org/abs/2010.16103},
  note          = {NeurIPS 2021}
}

@misc{hamilton2017graphsage,
  title         = {{Inductive Representation Learning on Large Graphs}},
  author        = {Hamilton, William L. and Ying, Rex and Leskovec, Jure},
  year          = {2017},
  eprint        = {1706.02216},
  archivePrefix = {arXiv},
  primaryClass  = {cs.SI},
  url           = {https://arxiv.org/abs/1706.02216},
  note          = {NIPS 2017}
}

@misc{dou2020caregnn,
  title         = {{Enhancing Graph Neural Network-based Fraud Detectors against Camouflaged Fraudsters}},
  author        = {Dou, Yingtong and Liu, Zhiwei and Sun, Li and Deng, Yutong and Peng, Hao and Yu, Philip S.},
  year          = {2020},
  eprint        = {2008.08692},
  archivePrefix = {arXiv},
  primaryClass  = {cs.SI},
  url           = {https://arxiv.org/abs/2008.08692},
  note          = {CIKM 2020. Related DOI: 10.1145/3340531.3411903}
}

@misc{kapoor2023leakage,
  title         = {{Leakage and the Reproducibility Crisis in ML-based Science}},
  author        = {Kapoor, Sayash and Narayanan, Arvind},
  year          = {2022},
  eprint        = {2207.07048},
  archivePrefix = {arXiv},
  primaryClass  = {cs.LG},
  url           = {https://arxiv.org/abs/2207.07048}
}

@misc{geirhos2020shortcut,
  title         = {{Shortcut Learning in Deep Neural Networks}},
  author        = {Geirhos, Robert and Jacobsen, J{\"o}rn-Henrik and Michaelis, Claudio and Zemel, Richard and Brendel, Wieland and Bethge, Matthias and Wichmann, Felix A.},
  year          = {2020},
  eprint        = {2004.07780},
  archivePrefix = {arXiv},
  primaryClass  = {cs.CV},
  url           = {https://arxiv.org/abs/2004.07780},
  note          = {Perspective article published at Nature Machine Intelligence. Related DOI: 10.1038/s42256-020-00257-z}
}

@inproceedings{liu2021pcgnn,
  title     = {{Pick and Choose: A GNN-based Imbalanced Learning Approach for Fraud Detection}},
  author    = {Liu, Yang and Ao, Xiang and Qin, Zidi and Chi, Jianfeng and Feng, Jinghua and Yang, Hao and He, Qing},
  booktitle = {Proceedings of the Web Conference 2021},
  publisher = {ACM},
  year      = {2021},
  pages     = {3168--3177},
  doi       = {10.1145/3442381.3449989}
}

@article{adamic2003friends,
  title   = {{Friends and neighbors on the Web}},
  author  = {Adamic, Lada A. and Adar, Eytan},
  journal = {Social Networks},
  year    = {2003},
  volume  = {25},
  number  = {3},
  pages   = {211--230},
  doi     = {10.1016/S0378-8733(03)00009-1}
}

@article{katz1953,
  title   = {{A New Status Index Derived from Sociometric Analysis}},
  author  = {Katz, Leo},
  journal = {Psychometrika},
  year    = {1953},
  volume  = {18},
  number  = {1},
  pages   = {39--43},
  doi     = {10.1007/BF02289026}
}

@article{libennowell2007,
  title   = {{The link-prediction problem for social networks}},
  author  = {Liben-Nowell, David and Kleinberg, Jon},
  journal = {Journal of the American Society for Information Science and Technology},
  year    = {2007},
  volume  = {58},
  number  = {7},
  pages   = {1019--1031},
  doi     = {10.1002/asi.20591}
}

@inproceedings{davis2006prroc,
  title     = {{The relationship between Precision-Recall and ROC curves}},
  author    = {Davis, Jesse and Goadrich, Mark},
  booktitle = {{Proceedings of the 23rd International Conference on Machine Learning (ICML '06)}},
  publisher = {ACM Press},
  year      = {2006},
  pages     = {233--240},
  doi       = {10.1145/1143844.1143874}
}

@article{saito2015prc,
  title   = {{The Precision-Recall Plot Is More Informative than the ROC Plot When Evaluating Binary Classifiers on Imbalanced Datasets}},
  author  = {Saito, Takaya and Rehmsmeier, Marc},
  journal = {PLOS ONE},
  year    = {2015},
  volume  = {10},
  number  = {3},
  pages   = {e0118432},
  doi     = {10.1371/journal.pone.0118432}
}

@article{kaufman2012leakage,
  title   = {{Leakage in data mining: Formulation, detection, and avoidance}},
  author  = {Kaufman, Shachar and Rosset, Saharon and Perlich, Claudia and Stitelman, Ori},
  journal = {ACM Transactions on Knowledge Discovery from Data},
  year    = {2012},
  volume  = {6},
  number  = {4},
  pages   = {1--21},
  doi     = {10.1145/2382577.2382579}
}

@article{nosek2018prereg,
  title   = {{The preregistration revolution}},
  author  = {Nosek, Brian A. and Ebersole, Charles R. and DeHaven, Alexander C. and Mellor, David T.},
  journal = {Proceedings of the National Academy of Sciences},
  year    = {2018},
  volume  = {115},
  number  = {11},
  pages   = {2600--2606},
  doi     = {10.1073/pnas.1708274114}
}

@article{efron1979bootstrap,
  title   = {{Bootstrap Methods: Another Look at the Jackknife}},
  author  = {Efron, B.},
  journal = {The Annals of Statistics},
  year    = {1979},
  volume  = {7},
  number  = {1},
  doi     = {10.1214/aos/1176344552}
}

@article{field2007clusterbootstrap,
  title   = {{Bootstrapping Clustered Data}},
  author  = {Field, C. A. and Welsh, A. H.},
  journal = {Journal of the Royal Statistical Society Series B: Statistical Methodology},
  year    = {2007},
  volume  = {69},
  number  = {3},
  pages   = {369--390},
  doi     = {10.1111/j.1467-9868.2007.00593.x}
}

@inproceedings{haveliwala2002ppr,
  title     = {{Topic-sensitive PageRank}},
  author    = {Haveliwala, Taher H.},
  booktitle = {Proceedings of the 11th international conference on World Wide Web},
  publisher = {ACM},
  year      = {2002},
  pages     = {517--526},
  doi       = {10.1145/511446.511513}
}

@article{holm1979,
  title   = {{A Simple Sequentially Rejective Multiple Test Procedure}},
  author  = {Holm, Sture},
  journal = {Scandinavian Journal of Statistics},
  volume  = {6},
  number  = {2},
  pages   = {65--70},
  year    = {1979}
}

@misc{acfe2024rttn,
  title        = {{Occupational Fraud 2024: A Report to the Nations}},
  author       = {{Association of Certified Fraud Examiners}},
  year         = {2024},
  howpublished = {Association of Certified Fraud Examiners, Austin, TX},
  url          = {https://www.acfe.com/fraud-resources/report-to-the-nations-archive}
}

@book{wells2017corporate,
  title     = {{Corporate Fraud Handbook: Prevention and Detection}},
  author    = {Wells, Joseph T.},
  publisher = {Wiley},
  year      = {2017},
  doi       = {10.1002/9781119351962}
}

@book{vona2017fraudanalytics,
  title     = {{Fraud Data Analytics Methodology: The Fraud Scenario Approach to Uncovering Fraud in Core Business Systems}},
  author    = {Vona, Leonard W.},
  publisher = {Wiley},
  year      = {2017},
  doi       = {10.1002/9781119270331}
}

@book{nigrini2012benford,
  title     = {{Benford's Law: Applications for Forensic Accounting, Auditing, and Fraud Detection}},
  author    = {Nigrini, Mark J.},
  publisher = {Wiley},
  year      = {2012},
  doi       = {10.1002/9781119203094}
}

@article{alles2008continuous,
  title   = {{Putting Continuous Auditing Theory into Practice: Lessons from Two Pilot Implementations}},
  author  = {Alles, Michael G. and Kogan, Alexander and Vasarhelyi, Miklos A.},
  journal = {Journal of Information Systems},
  volume  = {22},
  number  = {2},
  pages   = {195--214},
  year    = {2008},
  doi     = {10.2308/jis.2008.22.2.195}
}

@article{jans2013processmining,
  title   = {{The case for process mining in auditing: Sources of value added and areas of application}},
  author  = {Jans, Mieke and Alles, Michael and Vasarhelyi, Miklos},
  journal = {International Journal of Accounting Information Systems},
  volume  = {14},
  number  = {1},
  pages   = {1--20},
  year    = {2013},
  doi     = {10.1016/j.accinf.2012.06.015}
}

@article{fellegi1969recordlinkage,
  title   = {{A Theory for Record Linkage}},
  author  = {Fellegi, Ivan P. and Sunter, Alan B.},
  journal = {Journal of the American Statistical Association},
  volume  = {64},
  number  = {328},
  pages   = {1183--1210},
  year    = {1969},
  doi     = {10.1080/01621459.1969.10501049}
}

@article{debrah2026payrollreview,
  title   = {{Artificial Intelligence and Machine Learning for Payroll Fraud Detection in the United States Public Sector Payroll Systems: A Scoping Review}},
  author  = {Debrah, Regina and Oman-Amoako, Matthew and Tetteh, Ebenezer},
  journal = {Asian Journal of Economics, Business and Accounting},
  volume  = {26},
  number  = {7},
  pages   = {65--90},
  year    = {2026},
  doi     = {10.9734/ajeba/2026/v26i72317}
}

@misc{saha2025payrollerp,
  title        = {{Leveraging AI to Predict Payroll Fraud in Enterprise Resource Planning (ERP) Systems}},
  author       = {Saha, Biswanath},
  year         = {2025},
  howpublished = {SSRN preprint},
  doi          = {10.2139/ssrn.5224746},
  url          = {https://doi.org/10.2139/ssrn.5224746}
}

@article{sultana2025hris,
  title   = {{HRIS-Integrated Fraud Detection for U.S. Home Health Agency Workforce Compliance: Identifying Ghost Worker Schemes and Payroll Irregularities in Medicare and Medicaid-Funded Care Programs}},
  author  = {Sultana, Nasrin and Nasir, Abu and Hasan, Asif},
  journal = {International Journal of Computational and Experimental Science and Engineering},
  volume  = {11},
  number  = {1},
  year    = {2025},
  doi     = {10.22399/ijcesen.5496}
}

@inproceedings{bukhori2023inductive,
  title     = {{Inductive Link Prediction Banking Fraud Detection System Using Homogeneous Graph-Based Machine Learning Model}},
  author    = {Bukhori, Hilmi Aziz and Munir, Rinaldi},
  booktitle = {2023 IEEE 13th Annual Computing and Communication Workshop and Conference (CCWC)},
  pages     = {246--251},
  year      = {2023},
  doi       = {10.1109/ccwc57344.2023.10099180}
}

@misc{rahim2023oracle,
  title         = {{Oracle R12 EBusiness Suite Role Based Access Control and Roles Lifecycle Management}},
  author        = {Rahim, Sajid},
  year          = {2023},
  eprint        = {2304.13514},
  archivePrefix = {arXiv},
  url           = {https://arxiv.org/abs/2304.13514}
}

@misc{galkin2021nodepiece,
  title         = {{NodePiece: Compositional and Parameter-Efficient Representations of Large Knowledge Graphs}},
  author        = {Galkin, Mikhail and Denis, Etienne and Wu, Jiapeng and Hamilton, William L.},
  year          = {2021},
  eprint        = {2106.12144},
  archivePrefix = {arXiv},
  url           = {https://arxiv.org/abs/2106.12144},
  note          = {ICLR 2022}
}

@misc{galkin2022ilpc,
  title         = {{An Open Challenge for Inductive Link Prediction on Knowledge Graphs}},
  author        = {Galkin, Mikhail and Berrendorf, Max and Hoyt, Charles Tapley},
  year          = {2022},
  eprint        = {2203.01520},
  archivePrefix = {arXiv},
  url           = {https://arxiv.org/abs/2203.01520}
}

@inproceedings{li2023heart,
  title         = {{Evaluating Graph Neural Networks for Link Prediction: Current Pitfalls and New Benchmarking}},
  author        = {Li, Juanhui and Shomer, Harry and Mao, Haitao and Zeng, Shenglai and Ma, Yao and Shah, Neil and Tang, Jiliang and Yin, Dawei},
  year          = {2023},
  eprint        = {2306.10453},
  archivePrefix = {arXiv},
  url           = {https://arxiv.org/abs/2306.10453},
  booktitle     = {Advances in Neural Information Processing Systems, Datasets and Benchmarks Track}
}

@misc{tang2022bwgnn,
  title         = {{Rethinking Graph Neural Networks for Anomaly Detection}},
  author        = {Tang, Jianheng and Li, Jiajin and Gao, Ziqi and Li, Jia},
  year          = {2022},
  eprint        = {2205.15508},
  archivePrefix = {arXiv},
  url           = {https://arxiv.org/abs/2205.15508},
  note          = {ICML 2022}
}

@inproceedings{tang2023gadbench,
  title         = {{GADBench: Revisiting and Benchmarking Supervised Graph Anomaly Detection}},
  author        = {Tang, Jianheng and Hua, Fengrui and Gao, Ziqi and Zhao, Peilin and Li, Jia},
  year          = {2023},
  eprint        = {2306.12251},
  archivePrefix = {arXiv},
  url           = {https://arxiv.org/abs/2306.12251},
  booktitle     = {Advances in Neural Information Processing Systems, Datasets and Benchmarks Track}
}

@misc{elmougy2023ellipticpp,
  title         = {{Demystifying Fraudulent Transactions and Illicit Nodes in the Bitcoin Network for Financial Forensics}},
  author        = {Elmougy, Youssef and Liu, Ling},
  year          = {2023},
  eprint        = {2306.06108},
  archivePrefix = {arXiv},
  url           = {https://arxiv.org/abs/2306.06108}
}

@misc{gardner2020contrast,
  title         = {{Evaluating Models' Local Decision Boundaries via Contrast Sets}},
  author        = {Gardner, Matt and Artzi, Yoav and Basmova, Victoria and Berant, Jonathan and Bogin, Ben and Chen, Sihao and Dasigi, Pradeep and Dua, Dheeru and Elazar, Yanai and Gottumukkala, Ananth and Gupta, Nitish and Hajishirzi, Hanna and Ilharco, Gabriel and Khashabi, Daniel and Lin, Kevin and Liu, Jiangming and Liu, Nelson F. and Mulcaire, Phoebe and Ning, Qiang and Singh, Sameer and Smith, Noah A. and Subramanian, Sanjay and Tsarfaty, Reut and Wallace, Eric and Zhang, Ally and Zhou, Ben},
  year          = {2020},
  eprint        = {2004.02709},
  archivePrefix = {arXiv},
  url           = {https://arxiv.org/abs/2004.02709},
  note          = {Findings of EMNLP 2020}
}

@misc{coso2013internal,
  author       = {{Committee of Sponsoring Organizations of the Treadway Commission (COSO)}},
  title        = {Internal Control---Integrated Framework},
  year         = {2013},
  howpublished = {COSO},
  note         = {\url{https://www.coso.org/guidance-on-ic}}
}

@misc{iaasb2025isa240,
  author       = {{International Auditing and Assurance Standards Board (IAASB)}},
  title        = {{ISA} 240 (Revised), The Auditor's Responsibilities Relating to Fraud in an Audit of Financial Statements},
  year         = {2025},
  howpublished = {IAASB / IFAC},
  note         = {Effective for periods beginning on or after 15 December 2026; available from \url{https://www.iaasb.org}}
}

@misc{pedregosa2011sklearn,
  title         = {{Scikit-learn: Machine Learning in Python}},
  author        = {Pedregosa, Fabian and Varoquaux, Ga{\"e}l and Gramfort, Alexandre and Michel, Vincent and Thirion, Bertrand and Grisel, Olivier and Blondel, Mathieu and M{\"u}ller, Andreas and Nothman, Joel and Louppe, Gilles and Prettenhofer, Peter and Weiss, Ron and Dubourg, Vincent and Vanderplas, Jake and Passos, Alexandre and Cournapeau, David and Brucher, Matthieu and Perrot, Matthieu and Duchesnay, {\'E}douard},
  year          = {2011},
  eprint        = {1201.0490},
  archivePrefix = {arXiv},
  primaryClass  = {cs.LG},
  url           = {https://arxiv.org/abs/1201.0490},
  note          = {Journal of Machine Learning Research 12, 2011}
}

\clearpage
\appendix
\section{Protocol History and Execution Provenance}
\label{sec:deviations}

The development record distinguishes pilots, the original frozen study, score-only analyses and the extension (\cref{tab:study-lineage}). Their results remain separately identified.

\begin{enumerate}[leftmargin=*,itemsep=3pt]
\item \textbf{Exploratory development.} Phases 1 and 2 used generator v1 before the prospective protocol was finalized. RDL-style and audit-test breakdowns were added after a phase-2 development run. These outputs are exploratory (\cref{sec:leakage,tab:phase2}).
\item \textbf{Protocol finalization.} Before original testing, the family expanded from three hop tests to five tests, the decision became the Holm-adjusted bootstrap-tail rule, and the combination hypothesis became descriptive S1. These changes are in \texttt{freeze-phase3-001}.
\item \textbf{Seed and export qualification.} The exporter refused an adjacent seed that would reproduce earlier organizations. Final seed \GhBaseSeed{} was checked against earlier corpora. An initial v2 export matched two-hop variants only in distribution; a test exposed this before GNN results. Corrected \texttt{hrgraph-004} supplied the study. Only dev organizations of superseded \texttt{hrgraph-003} were analyzed; interrupted dev GNN runs supplied no result used here.
\item \textbf{Audit corrections.} Intermediate dev audits mixed hops in a twin comparison and used an incorrect entity lookup. Corrected audits superseded them; reported values use the final audit.
\item \textbf{Development-file retention.} A phase-1 debug output exposing an omitted four-hop intermediary and two seed-refusal logs were deleted during development; none contributes a result. Later debug runs were retained separately. Partial extension smoke outputs and scratch noise data were deleted; the final dev-only smoke run was retained. A number-check directory deleted during manuscript rebuilding was regenerated. The run log records these events separately from preserved formal outputs.
\item \textbf{Rule construction.} Severity weights followed phase-1 inspection of ties. Rules v2 has hand-set weights and generator-informed exemptions. Generic rules have unit weights and different exemptions; both were authored with knowledge of the generator, not as a blinded manipulation of analyst knowledge.
\item \textbf{Twin analysis.} The report forms pools from written scores. Original twin intervals and comparisons were implemented before the freeze and regression-checked on dev values.
\item \textbf{Implementation.} GraIL-style concatenates features and distance labels; NBFNet-style injects features at each layer; the opening-subgraph model is inspired by GLAR~\citep{xie2024onesubgraph}. Hyperparameters stayed fixed from phase 2. Saturating validation-calibrated \texttt{max} combinations remain descriptive appendix rows.
\item \textbf{Score-only analyses.} Frozen scores and test truth supplied seed comparisons, twin removal, concordance, re-weighting and SoD exclusions after the original report. No model was trained for them and no original decision replaced (\cref{sec:posthoc}).
\item \textbf{Extension and test reuse.} The extension was designed after original results were examined and frozen before new formal computations. A/C reuse hrgraph-004 test organizations; B uses new hrgraph-005. Dev-only smoke tests preceded the freeze. NBFNet-style replaces GraIL as the stage-C link model.
\item \textbf{Extension interpretation.} The protocol's actual A1-supported/A2-unsupported outcome retained explanations beyond generator knowledge for H2. The final interpretation therefore describes sensitivity to rule specification without identifying knowledge as the unique or predominant cause. All nine recorded decisions are retained.
\item \textbf{Thread-count reproduction.} Four rather than six CPU threads reproduced RDL-style scores but changed NBFNet-style clean AP to \GvV{C.repro.nbfnet.retrained} from \GvV{C.repro.nbfnet.frozen}. Original and extension outputs remain distinct.
\end{enumerate}

Internal simulated manuscript reviews in the run log were preparation activities, not journal peer review or evidence for performance.

\FloatBarrier
\section{Literature Search}
\label{app:litsearch}

The novelty statement in \cref{sec:related} rests on a set of arXiv API queries covering 2018--2026, run on 14 October 2026, with abstracts read for the closest hits. The following queries returned no relevant hit: ``ghost employee''; ``payroll fraud''; payroll AND fraud; ``inductive link prediction'' AND fraud; ``knowledge graph completion'' AND fraud; ``vendor fraud''; ``shared bank account'' or ``shared address'' AND fraud; procurement AND fraud AND ``knowledge graph''; ``fraud ring'' AND synthetic AND graph; inductive AND ``unseen entities'' AND fraud; ``knowledge graph'' AND audit AND fraud. Semantic Scholar was rate-limited during that search. To cover accounting, auditing and information-systems journals and books outside arXiv, we queried the Crossref API on 15--16 October 2026 (bibliographic queries ``ghost employee payroll fraud'', ``payroll fraud detection machine learning'' and ``graph link prediction occupational fraud audit employee vendor''; top 20 hits each; a fourth query was rate-limited and not repeated) and read the abstracts of the relevant hits. The search returned practitioner handbook chapters on payroll schemes~\citep{wells2017corporate,vona2017fraudanalytics}, public-sector and policy studies of ghost workers, a scoping review and a preprint on machine learning for payroll fraud~\citep{debrah2026payrollreview,saha2025payrollerp}, a synthetic HRIS study with tabular classifiers and network-style collusion indicators~\citep{sultana2025hris}, and an inductive link-prediction system for banking fraud~\citep{bukhori2023inductive}; none formulates payroll or employee--vendor fraud as link prediction. Crossref coverage of accounting journals is incomplete, so the novelty statement remains ``to our knowledge''. Every reference was checked against its arXiv abstract page, its Crossref DOI record, or (for the ACFE report) the publisher's archive page on 15--16 October 2026; Holm (1979) has no DOI and was checked against a secondary reference list.

\FloatBarrier
\section{Design Details}
\label{app:design}

\paragraph{Planted counts.} \Cref{tab:design-counts} lists the planted positives per pattern and hop in the test organizations. Each test organization contributes \GhPosPerOrg{} planted frauds and the same number of matched twins.

% generated by xiaobo-experiment/render-results-draft.py; do not edit
\begin{table}[htbp]\centering\footnotesize\setlength{\tabcolsep}{3.5pt}
\caption{Planted positives per pattern and hop in the 24 test organizations (analytic counts from the generator configuration). Every planted pair also has one structurally matched benign twin.}
\label{tab:design-counts}
\begin{tabular}{lrrr}
\toprule
pattern & hop & entity $n_+$ & link $n_+$ \\
\midrule
account change before run & 2 & 48 & 48 \\
account change before run & 3 & 48 & 48 \\
approver link & 2 & 48 & 48 \\
approver link & 3 & 48 & 48 \\
employee--vendor account & 2 & 32 & 32 \\
employee--vendor account & 3 & 32 & 32 \\
employee--vendor account & 4 & 32 & 32 \\
employee--vendor contact & 2 & 32 & 32 \\
employee--vendor contact & 3 & 32 & 32 \\
employee--vendor contact & 4 & 32 & 32 \\
ghost employee & 1 & 48 & 48 \\
ghost employee & 3 & 48 & 48 \\
shared payroll account & 2 & 32 & 32 \\
shared payroll account & 3 & 32 & 32 \\
shared payroll account & 4 & 32 & 32 \\
SoD conflict & 0 & 96 & -- \\
\bottomrule
\end{tabular}
\end{table}

\paragraph{Rules v2 exemptions.} An identifier match (shared account, address, phone, approver link) is exempt when a conflict-of-interest declaration is on file, filed on time, and the approval was independent. A holder mismatch is exempt when a third-party payee authorization is on file. A recent bank-detail change is exempt when it was call-back verified. ``Paid without activity'' fires on zero tracked attendance, or on low badge \emph{and} login activity for a tracked employee not on leave. A separate flag fires when the hiring or change approval was not independent. The link score adds direct identifier matches between $u$ and $c$ (unless exempt), plus one if $u$ is paid without activity and reports to $c$, plus one if $u$ has an approver link and was hired by $c$, plus a small multiple of the weighted entity flags.

\paragraph{Node features.} Type one-hot; log degree, total and per relation; the phase-1 entity-attribute flags; new-entity indicator; runs paid; attendance, start and onboarding times (log); account change; roles; address and phone kind; and the v2 control records (conflict-of-interest declaration and its timing, payee authorization, call-back verification, independent approval, badge and login activity, time-sheets, SoD exception), each with a missing indicator. The feature dimension is \GhFeatureDim.

\paragraph{Tests.} Before the freeze, the JavaScript test suite (generator determinism, label validity, hard-negative behaviour, metrics on hand-computed cases including ties, leakage checks, export reproducibility, and the twin audit on dev data) and the Python test suite (subgraph extraction against a reference implementation, labelling, that the loader never opens label columns or test organizations in dev mode, that no feature equals the label, invariance of every model class to identifier relabelling, and metric parity with the JavaScript implementation) all passed.

\FloatBarrier
\section{Hyperparameters and Training}
\label{app:training}

% generated by xiaobo-experiment/render-results-draft.py; do not edit
\begin{table}[htbp]\centering\footnotesize\setlength{\tabcolsep}{3.5pt}
\caption{GNN hyperparameters, fixed in phase 2 before generator v2 existed and not tuned (as recorded in the confirmatory run's metrics file).}
\label{tab:hparams}
\begin{adjustbox}{max width=\linewidth}
\begin{tabular}{ll}
\toprule
model & hyperparameters \\
\midrule
R-GCN (entity) & hid 64, layers 2, bases 8, dropout 0.2, lr 0.005, max epochs 300, patience 30 \\
RDL-style (entity, link) & hid 64, layers 2, dropout 0.2, lr 0.005, max epochs 300, patience 30 \\
GraIL-style ($k{=}2$, $k{=}3$) & hid 32, layers 3, bases 4, dropout 0.1, lr 0.002, max epochs 40, patience 8, batch 256, neg frac 0.25 \\
NBFNet-style & hid 32, layers 6, lr 0.002, max epochs 60, patience 10, chunk 28 \\
Opening-subgraph (OSFA-insp.) & k 4, hid 32, bases 4, dropout 0.1, lr 0.002, max epochs 60, patience 10, chunk 14 \\
\bottomrule
\end{tabular}
\end{adjustbox}
\end{table}

% generated by xiaobo-experiment/render-results-draft.py; do not edit
\begin{table}[htbp]\centering\footnotesize\setlength{\tabcolsep}{3.5pt}
\caption{Training on the confirmatory run (seeds 9131 / 9132 / 9133): early stopping on the held-out dev orgs ended almost every run before its epoch cap; runs that reached the cap: Opening-subgraph (OSFA-insp.) seed 9132.}
\label{tab:training}
\begin{adjustbox}{max width=\linewidth}
\begin{tabular}{lllrlr}
\toprule
model & best epoch & epochs run & cap & val AUPRC & wall h (3 seeds) \\
\midrule
R-GCN (entity) & 91 / 33 / 28 & 122 / 64 / 59 & 300 & 0.777 / 0.771 / 0.764 & 0.10 \\
RDL-style (entity) & 73 / 43 / 62 & 104 / 74 / 93 & 300 & 0.802 / 0.799 / 0.778 & 0.11 \\
RDL-style (link) & 140 / 140 / 193 & 171 / 171 / 224 & 300 & 0.099 / 0.141 / 0.082 & 0.27 \\
GraIL-style $k{=}2$ & 17 / 25 / 15 & 26 / 34 / 24 & 40 & 0.493 / 0.572 / 0.553 & 2.23 \\
GraIL-style $k{=}3$ & 24 / 11 / 18 & 33 / 20 / 27 & 40 & 0.712 / 0.586 / 0.629 & 9.06 \\
NBFNet-style & 18 / 21 / 34 & 29 / 32 / 45 & 60 & 0.649 / 0.620 / 0.662 & 0.36 \\
Opening-subgraph (OSFA-insp.) & 29 / 52 / 25 & 40 / 60 / 36 & 60 & 0.553 / 0.540 / 0.501 & 1.02 \\
\bottomrule
\end{tabular}
\end{adjustbox}
\end{table}

\FloatBarrier
\section{Full Original-Study Tables}
\label{app:tables}

All tables in this appendix are generated from the confirmatory report. GNN values are seed means; ``$\pm$'' is the seed SD; brackets are 95\% organization-cluster bootstrap CIs over the \GhNTestOrgs{} test organizations. Per-pattern and per-control breakdowns are exploratory.

% generated by xiaobo-experiment/render-results-draft.py; do not edit
\begin{table}[htbp]\centering\footnotesize\setlength{\tabcolsep}{3.5pt}
\caption{All link-level methods on the test pool: recall at 1\% FPR is saturated (9 of 24 methods reach at least 0.99), as declared in the prospective protocol, so no claim rests on it. P@100 (precision in the top 100) was not pre-specified. $^\dagger$GNN that is not a pre-specified primary model (descriptive).}
\label{tab:link-full}
\begin{adjustbox}{max width=\linewidth}
\begin{tabular}{lrlllrrrl}
\toprule
method & AUROC & AUPRC & 95\% CI & P@100 & \makecell[l]{recall at\\ {}0.1\% FPR} & \makecell[l]{recall at\\ {}1\% FPR} & \makecell[l]{recall at\\ {}5\% FPR} & \makecell[l]{per-org\\ {}AUROC range} \\
\midrule
Rules naive & 0.698 & 0.024 & [0.020, 0.029] & 0.178 & 0.090 & 0.101 & 0.448 & 0.631--0.771 \\
Rules tuned & 0.765 & 0.237 & [0.222, 0.253] & 0.505 & 0.434 & 0.434 & 0.434 & 0.707--0.834 \\
Rules severity & 0.772 & 0.239 & [0.224, 0.256] & 0.505 & 0.434 & 0.434 & 0.434 & 0.714--0.837 \\
Rules v2 & 0.808 & 0.347 & [0.321, 0.376] & 0.790 & 0.429 & 0.429 & 0.436 & 0.709--0.891 \\
Common neighbours & 0.645 & 0.003 & [0.003, 0.003] & 0.003 & 0.000 & 0.014 & 0.014 & 0.599--0.717 \\
Adamic--Adar & 0.659 & 0.124 & [0.113, 0.137] & 0.215 & 0.333 & 0.340 & 0.359 & 0.613--0.732 \\
Katz & 0.829 & 0.023 & [0.022, 0.025] & 0.050 & 0.042 & 0.337 & 0.356 & 0.779--0.886 \\
PPR & 0.993 & 0.093 & [0.090, 0.100] & 0.090 & 0.057 & 0.797 & 1.000 & 0.990--0.994 \\
Path features + LR & 0.981 & 0.295 & [0.281, 0.315] & 0.410 & 0.530 & 0.819 & 0.896 & 0.964--0.992 \\
NBFNet-style$^\dagger$ & 0.999 & 0.673$\pm$0.017 & [0.645, 0.698] & 0.830 & 0.936 & 0.999 & 0.999 & 0.992--1.000 \\
OSFA-insp.$^\dagger$ & 0.999 & 0.584$\pm$0.033 & [0.559, 0.617] & 0.577 & 0.900 & 0.998 & 0.998 & 0.994--1.000 \\
RDL-style (link)$^\dagger$ & 0.953 & 0.082$\pm$0.023 & [0.075, 0.092] & 0.280 & 0.148 & 0.469 & 0.806 & 0.924--0.978 \\
Rules v2 + NBFNet-style$^\dagger$ (max) & 0.999 & 0.619$\pm$0.018 & [0.527, 0.720] & 0.857 & 0.792 & 0.999 & 0.999 & 0.988--1.000 \\
Rules v2 + NBFNet-style$^\dagger$ (lr) & 0.999 & 0.626$\pm$0.111 & [0.605, 0.648] & 0.717 & 0.910 & 0.999 & 0.999 & 0.992--1.000 \\
Rules v2 + OSFA-insp.$^\dagger$ (max) & 0.999 & 0.583$\pm$0.011 & [0.491, 0.691] & 0.658 & 0.852 & 0.998 & 0.998 & 0.988--1.000 \\
Rules v2 + OSFA-insp.$^\dagger$ (lr) & 0.999 & 0.624$\pm$0.063 & [0.599, 0.654] & 0.627 & 0.921 & 0.998 & 0.998 & 0.994--1.000 \\
Rules v2 + RDL-style (link)$^\dagger$ (max) & 0.955 & 0.292$\pm$0.014 & [0.256, 0.351] & 0.626 & 0.448 & 0.621 & 0.796 & 0.914--0.990 \\
Rules v2 + RDL-style (link)$^\dagger$ (lr) & 0.966 & 0.391$\pm$0.002 & [0.364, 0.423] & 0.803 & 0.446 & 0.617 & 0.856 & 0.932--0.993 \\
GraIL-style $k{=}2$$^\dagger$ & 0.979 & 0.473$\pm$0.012 & [0.456, 0.493] & 0.597 & 0.789 & 0.841 & 0.915 & 0.940--0.997 \\
GraIL-style $k{=}3$ & 0.999 & 0.663$\pm$0.071 & [0.630, 0.698] & 0.760 & 0.906 & 0.997 & 0.999 & 0.998--1.000 \\
Rules v2 + GraIL-style $k{=}2$$^\dagger$ (max) & 0.979 & 0.462$\pm$0.012 & [0.392, 0.544] & 0.625 & 0.737 & 0.851 & 0.900 & 0.953--0.997 \\
Rules v2 + GraIL-style $k{=}2$$^\dagger$ (lr) & 0.983 & 0.591$\pm$0.009 & [0.567, 0.618] & 0.747 & 0.809 & 0.850 & 0.924 & 0.958--0.998 \\
Rules v2 + GraIL-style $k{=}3$ (max) & 0.999 & 0.655$\pm$0.061 & [0.594, 0.726] & 0.777 & 0.855 & 0.999 & 0.999 & 0.996--1.000 \\
Rules v2 + GraIL-style $k{=}3$ (lr) & 1.000 & 0.735$\pm$0.026 & [0.711, 0.763] & 0.777 & 0.961 & 0.999 & 0.999 & 0.998--1.000 \\
\bottomrule
\end{tabular}
\end{adjustbox}
\end{table}

% generated by xiaobo-experiment/render-results-draft.py; do not edit
\begin{table}[p]\centering\footnotesize\setlength{\tabcolsep}{3.5pt}
\caption{Link recall per hop at 0.1\% and 1\% FPR (hop-$h$ positives vs all negatives). At 1\% FPR the GraIL-style $k{=}3$, NBFNet-style and opening-subgraph models recall nearly all positives at every hop; the one-step rules recall none of the 4-hop positives at either FPR. $^\dagger$GNN that is not a pre-specified primary model (descriptive).}
\label{tab:link-hop-recall}
\begin{adjustbox}{max width=\linewidth}
\begin{tabular}{lrrrrrrrr}
\toprule
& \multicolumn{4}{c}{recall at 0.1\% FPR} & \multicolumn{4}{c}{recall at 1\% FPR} \\ \cmidrule(lr){2-5}\cmidrule(lr){6-9} method & h1 & h2 & h3 & h4 & h1 & h2 & h3 & h4 \\
\midrule
Rules naive & 0.000 & 0.260 & 0.008 & 0.000 & 0.000 & 0.276 & 0.021 & 0.000 \\
Rules tuned & 0.667 & 1.000 & 0.108 & 0.000 & 0.667 & 1.000 & 0.108 & 0.000 \\
Rules severity & 0.667 & 1.000 & 0.108 & 0.000 & 0.667 & 1.000 & 0.108 & 0.000 \\
Rules v2 & 0.833 & 0.854 & 0.179 & 0.000 & 0.833 & 0.854 & 0.179 & 0.000 \\
Common neighbours & 0.000 & 0.000 & 0.000 & 0.000 & 0.000 & 0.031 & 0.008 & 0.000 \\
Adamic--Adar & 0.000 & 1.000 & 0.000 & 0.000 & 0.000 & 1.000 & 0.017 & 0.000 \\
Katz & 0.479 & 0.005 & 0.000 & 0.000 & 1.000 & 0.250 & 0.408 & 0.000 \\
PPR & 0.667 & 0.005 & 0.000 & 0.000 & 1.000 & 1.000 & 0.787 & 0.312 \\
Path features + LR & 0.604 & 0.568 & 0.696 & 0.000 & 1.000 & 0.958 & 0.963 & 0.094 \\
NBFNet-style$^\dagger$ & 0.715 & 0.983 & 0.942 & 0.938 & 0.993 & 1.000 & 1.000 & 1.000 \\
OSFA-insp.$^\dagger$ & 0.583 & 0.913 & 0.926 & 0.969 & 0.986 & 0.998 & 1.000 & 1.000 \\
RDL-style (link)$^\dagger$ & 0.222 & 0.220 & 0.086 & 0.122 & 0.556 & 0.505 & 0.439 & 0.431 \\
Rules v2 + NBFNet-style$^\dagger$ (max) & 0.833 & 0.955 & 0.711 & 0.649 & 0.993 & 1.000 & 1.000 & 1.000 \\
Rules v2 + NBFNet-style$^\dagger$ (lr) & 0.708 & 0.906 & 0.943 & 0.938 & 0.993 & 1.000 & 1.000 & 1.000 \\
Rules v2 + OSFA-insp.$^\dagger$ (max) & 0.833 & 0.944 & 0.796 & 0.819 & 0.986 & 0.998 & 1.000 & 1.000 \\
Rules v2 + OSFA-insp.$^\dagger$ (lr) & 0.771 & 0.929 & 0.926 & 0.969 & 0.986 & 0.998 & 1.000 & 1.000 \\
Rules v2 + RDL-style (link)$^\dagger$ (max) & 0.833 & 0.858 & 0.197 & 0.066 & 0.833 & 0.880 & 0.454 & 0.413 \\
Rules v2 + RDL-style (link)$^\dagger$ (lr) & 0.833 & 0.842 & 0.194 & 0.087 & 0.833 & 0.868 & 0.456 & 0.410 \\
GraIL-style $k{=}2$$^\dagger$ & 0.632 & 0.970 & 0.990 & 0.000 & 0.826 & 0.979 & 1.000 & 0.177 \\
GraIL-style $k{=}3$ & 0.375 & 0.917 & 0.972 & 0.986 & 0.979 & 0.998 & 0.999 & 1.000 \\
Rules v2 + GraIL-style $k{=}2$$^\dagger$ (max) & 0.833 & 0.970 & 0.826 & 0.000 & 0.903 & 0.995 & 1.000 & 0.163 \\
Rules v2 + GraIL-style $k{=}2$$^\dagger$ (lr) & 0.819 & 0.991 & 0.985 & 0.000 & 0.882 & 0.993 & 1.000 & 0.174 \\
Rules v2 + GraIL-style $k{=}3$ (max) & 0.833 & 0.944 & 0.842 & 0.722 & 1.000 & 1.000 & 0.999 & 1.000 \\
Rules v2 + GraIL-style $k{=}3$ (lr) & 0.771 & 0.986 & 0.968 & 0.990 & 0.993 & 1.000 & 0.999 & 1.000 \\
\bottomrule
\end{tabular}
\end{adjustbox}
\end{table}

% generated by xiaobo-experiment/render-results-draft.py; do not edit
\begin{table}[htbp]\centering\footnotesize\setlength{\tabcolsep}{3.5pt}
\caption{All link-level methods on the twin pool: structure-only heuristics sit at chance (as designed) and no single method or combination exceeds the generator-aware Rules v2 (descriptive; unadjusted CIs). $^\dagger$GNN that is not a pre-specified primary model (descriptive).}
\label{tab:link-twins-full}
\begin{adjustbox}{max width=\linewidth}
\begin{tabular}{lrlrrrrrl}
\toprule
method & AUROC & 95\% CI & AUPRC & h1 & h2 & h3 & h4 & $\Delta$ vs Rules v2 [95\% CI] \\
\midrule
Rules naive & 0.536 & [0.526, 0.548] & 0.523 & 0.748 & 0.510 & 0.523 & 0.495 & $-$0.260 [$-$0.285, $-$0.234] \\
Rules tuned & 0.547 & [0.537, 0.555] & 0.527 & 0.780 & 0.501 & 0.547 & 0.502 & $-$0.250 [$-$0.269, $-$0.230] \\
Rules severity & 0.548 & [0.539, 0.557] & 0.529 & 0.780 & 0.499 & 0.542 & 0.505 & $-$0.248 [$-$0.267, $-$0.229] \\
Rules v2 & 0.796 & [0.773, 0.818] & 0.742 & 0.893 & 0.818 & 0.776 & 0.780 & -- \\
Common neighbours & 0.498 & [0.489, 0.509] & 0.499 & 0.479 & 0.497 & 0.502 & 0.495 & $-$0.298 [$-$0.326, $-$0.270] \\
Adamic--Adar & 0.497 & [0.489, 0.505] & 0.497 & 0.479 & 0.479 & 0.503 & 0.494 & $-$0.300 [$-$0.328, $-$0.273] \\
Katz & 0.494 & [0.485, 0.502] & 0.500 & 0.514 & 0.514 & 0.487 & 0.467 & $-$0.303 [$-$0.327, $-$0.278] \\
PPR & 0.501 & [0.496, 0.507] & 0.507 & 0.475 & 0.499 & 0.504 & 0.503 & $-$0.295 [$-$0.317, $-$0.271] \\
Path features + LR & 0.619 & [0.602, 0.637] & 0.602 & 0.829 & 0.629 & 0.644 & 0.742 & $-$0.177 [$-$0.201, $-$0.151] \\
NBFNet-style$^\dagger$ & 0.668 & [0.642, 0.691] & 0.676 & 0.848 & 0.676 & 0.692 & 0.549 & $-$0.129 [$-$0.152, $-$0.103] \\
OSFA-insp.$^\dagger$ & 0.635 & [0.611, 0.663] & 0.593 & 0.874 & 0.612 & 0.648 & 0.713 & $-$0.161 [$-$0.187, $-$0.131] \\
RDL-style (link)$^\dagger$ & 0.738 & [0.715, 0.764] & 0.727 & 0.856 & 0.743 & 0.720 & 0.718 & $-$0.058 [$-$0.080, $-$0.034] \\
Rules v2 + NBFNet-style$^\dagger$ (max) & 0.711 & [0.686, 0.734] & 0.703 & 0.851 & 0.766 & 0.703 & 0.549 & $-$0.086 [$-$0.105, $-$0.064] \\
Rules v2 + NBFNet-style$^\dagger$ (lr) & 0.622 & [0.602, 0.642] & 0.630 & 0.846 & 0.616 & 0.668 & 0.547 & $-$0.174 [$-$0.198, $-$0.149] \\
Rules v2 + OSFA-insp.$^\dagger$ (max) & 0.714 & [0.690, 0.738] & 0.670 & 0.887 & 0.762 & 0.665 & 0.713 & $-$0.083 [$-$0.106, $-$0.058] \\
Rules v2 + OSFA-insp.$^\dagger$ (lr) & 0.669 & [0.648, 0.692] & 0.631 & 0.883 & 0.672 & 0.653 & 0.716 & $-$0.127 [$-$0.151, $-$0.102] \\
Rules v2 + RDL-style (link)$^\dagger$ (max) & 0.769 & [0.748, 0.790] & 0.737 & 0.880 & 0.814 & 0.725 & 0.718 & $-$0.028 [$-$0.040, $-$0.013] \\
Rules v2 + RDL-style (link)$^\dagger$ (lr) & 0.772 & [0.751, 0.794] & 0.744 & 0.863 & 0.816 & 0.735 & 0.729 & $-$0.025 [$-$0.039, $-$0.010] \\
GraIL-style $k{=}2$$^\dagger$ & 0.573 & [0.559, 0.588] & 0.564 & 0.748 & 0.592 & 0.609 & 0.740 & $-$0.224 [$-$0.244, $-$0.202] \\
GraIL-style $k{=}3$ & 0.674 & [0.643, 0.705] & 0.670 & 0.780 & 0.641 & 0.690 & 0.803 & $-$0.122 [$-$0.151, $-$0.094] \\
Rules v2 + GraIL-style $k{=}2$$^\dagger$ (max) & 0.634 & [0.618, 0.650] & 0.608 & 0.839 & 0.706 & 0.620 & 0.765 & $-$0.163 [$-$0.184, $-$0.142] \\
Rules v2 + GraIL-style $k{=}2$$^\dagger$ (lr) & 0.672 & [0.656, 0.688] & 0.677 & 0.817 & 0.803 & 0.671 & 0.755 & $-$0.125 [$-$0.140, $-$0.108] \\
Rules v2 + GraIL-style $k{=}3$ (max) & 0.749 & [0.725, 0.774] & 0.714 & 0.860 & 0.758 & 0.719 & 0.802 & $-$0.048 [$-$0.067, $-$0.029] \\
Rules v2 + GraIL-style $k{=}3$ (lr) & 0.763 & [0.742, 0.783] & 0.738 & 0.839 & 0.813 & 0.736 & 0.828 & $-$0.034 [$-$0.047, $-$0.019] \\
\bottomrule
\end{tabular}
\end{adjustbox}
\end{table}

% generated by xiaobo-experiment/render-results-draft.py; do not edit
\begin{table}[htbp]\centering\footnotesize\setlength{\tabcolsep}{3.5pt}
\caption{All entity-level methods on the test pool: entity GNNs and the rules + entity-GNN combinations have the highest AUPRC, and link models aggregated by the maximum miss the entity-only SoD conflict (h0). Per-hop columns: AUPRC of hop-$h$ positives vs all negatives. $^\dagger$GNN that is not a pre-specified primary model (descriptive).}
\label{tab:entity-full}
\begin{adjustbox}{max width=\linewidth}
\begin{tabular}{lrllrrrrrrr}
\toprule
method & AUROC & AUPRC & 95\% CI & \makecell[l]{recall at\\ {}1\% FPR} & \makecell[l]{recall at\\ {}5\% FPR} & h0 & h1 & h2 & h3 & h4 \\
\midrule
Rules naive & 0.633 & 0.343 & [0.332, 0.357] & 0.016 & 0.016 & 0.076 & 0.027 & 0.192 & 0.150 & 0.048 \\
Rules tuned & 0.739 & 0.447 & [0.437, 0.458] & 0.012 & 0.012 & 0.084 & 0.043 & 0.337 & 0.220 & 0.080 \\
Rules severity & 0.758 & 0.486 & [0.475, 0.499] & 0.010 & 0.124 & 0.107 & 0.054 & 0.372 & 0.251 & 0.104 \\
Rules v2 & 0.872 & 0.698 & [0.669, 0.728] & 0.104 & 0.387 & 0.327 & 0.261 & 0.608 & 0.411 & 0.229 \\
Common neighbours & 0.461 & 0.254 & [0.250, 0.261] & 0.000 & 0.013 & 0.063 & 0.029 & 0.108 & 0.105 & 0.040 \\
Adamic--Adar & 0.517 & 0.302 & [0.295, 0.311] & 0.000 & 0.009 & 0.044 & 0.021 & 0.366 & 0.091 & 0.038 \\
Katz & 0.444 & 0.239 & [0.231, 0.250] & 0.004 & 0.042 & 0.080 & 0.061 & 0.074 & 0.092 & 0.047 \\
PPR & 0.454 & 0.235 & [0.229, 0.245] & 0.004 & 0.024 & 0.072 & 0.077 & 0.083 & 0.086 & 0.033 \\
Path features + LR & 0.876 & 0.725 & [0.692, 0.757] & 0.192 & 0.463 & 0.412 & 0.249 & 0.323 & 0.587 & 0.339 \\
R-GCN (entity)$^\dagger$ & 0.900 & 0.776$\pm$0.015 & [0.744, 0.808] & 0.233 & 0.546 & 0.502 & 0.283 & 0.455 & 0.628 & 0.416 \\
NBFNet-style$^\dagger$ & 0.791 & 0.621$\pm$0.017 & [0.596, 0.643] & 0.129 & 0.309 & 0.029 & 0.388 & 0.519 & 0.463 & 0.155 \\
OSFA-insp.$^\dagger$ & 0.787 & 0.550$\pm$0.023 & [0.529, 0.578] & 0.023 & 0.245 & 0.032 & 0.049 & 0.374 & 0.398 & 0.198 \\
RDL-style (entity) & 0.890 & 0.752$\pm$0.018 & [0.719, 0.784] & 0.197 & 0.524 & 0.411 & 0.243 & 0.445 & 0.576 & 0.405 \\
RDL-style (link)$^\dagger$ & 0.780 & 0.638$\pm$0.016 & [0.614, 0.665] & 0.120 & 0.381 & 0.028 & 0.273 & 0.488 & 0.480 & 0.264 \\
Rules v2 + R-GCN (entity)$^\dagger$ (max) & 0.916 & 0.782$\pm$0.013 & [0.752, 0.811] & 0.187 & 0.557 & 0.433 & 0.241 & 0.563 & 0.581 & 0.354 \\
Rules v2 + R-GCN (entity)$^\dagger$ (lr) & 0.921 & 0.801$\pm$0.004 & [0.769, 0.832] & 0.287 & 0.603 & 0.438 & 0.311 & 0.638 & 0.583 & 0.333 \\
Rules v2 + RDL-style (entity) (max) & 0.907 & 0.760$\pm$0.019 & [0.729, 0.789] & 0.089 & 0.494 & 0.359 & 0.207 & 0.553 & 0.536 & 0.345 \\
Rules v2 + RDL-style (entity) (lr) & 0.914 & 0.792$\pm$0.004 & [0.763, 0.822] & 0.266 & 0.567 & 0.411 & 0.283 & 0.640 & 0.565 & 0.334 \\
GraIL-style $k{=}2$$^\dagger$ & 0.726 & 0.506$\pm$0.007 & [0.492, 0.523] & 0.043 & 0.194 & 0.027 & 0.046 & 0.429 & 0.393 & 0.059 \\
GraIL-style $k{=}3$ & 0.782 & 0.612$\pm$0.060 & [0.584, 0.641] & 0.100 & 0.339 & 0.028 & 0.144 & 0.443 & 0.501 & 0.272 \\
\bottomrule
\end{tabular}
\end{adjustbox}
\end{table}

% generated by xiaobo-experiment/render-results-draft.py; do not edit
\begin{table}[htbp]\centering\footnotesize\setlength{\tabcolsep}{3.5pt}
\caption{All entity-level methods on the twin pool: entity GNNs tie Rules v2 and rules + entity-GNN combinations exceed it (descriptive; unadjusted CIs). $^\dagger$GNN that is not a pre-specified primary model (descriptive).}
\label{tab:entity-twins-full}
\begin{adjustbox}{max width=\linewidth}
\begin{tabular}{lrlrrrrrl}
\toprule
method & AUROC & 95\% CI & h0 & h1 & h2 & h3 & h4 & $\Delta$ vs Rules v2 [95\% CI] \\
\midrule
Rules naive & 0.519 & [0.500, 0.536] & 0.503 & 0.704 & 0.510 & 0.511 & 0.495 & $-$0.298 [$-$0.327, $-$0.269] \\
Rules tuned & 0.560 & [0.550, 0.572] & 0.787 & 0.756 & 0.503 & 0.539 & 0.502 & $-$0.257 [$-$0.276, $-$0.236] \\
Rules severity & 0.564 & [0.553, 0.575] & 0.789 & 0.780 & 0.499 & 0.542 & 0.505 & $-$0.253 [$-$0.271, $-$0.234] \\
Rules v2 & 0.817 & [0.793, 0.839] & 0.912 & 0.893 & 0.816 & 0.781 & 0.780 & -- \\
Common neighbours & 0.487 & [0.467, 0.507] & 0.472 & 0.542 & 0.494 & 0.479 & 0.493 & $-$0.330 [$-$0.365, $-$0.297] \\
Adamic--Adar & 0.491 & [0.480, 0.503] & 0.466 & 0.509 & 0.479 & 0.488 & 0.485 & $-$0.326 [$-$0.354, $-$0.298] \\
Katz & 0.492 & [0.479, 0.506] & 0.462 & 0.514 & 0.491 & 0.483 & 0.502 & $-$0.325 [$-$0.349, $-$0.299] \\
PPR & 0.504 & [0.489, 0.519] & 0.451 & 0.475 & 0.496 & 0.512 & 0.522 & $-$0.313 [$-$0.338, $-$0.288] \\
Path features + LR & 0.815 & [0.788, 0.837] & 0.906 & 0.865 & 0.798 & 0.790 & 0.844 & $-$0.002 [$-$0.021, 0.018] \\
R-GCN (entity)$^\dagger$ & 0.821 & [0.796, 0.846] & 0.877 & 0.883 & 0.763 & 0.827 & 0.856 & +0.004 [$-$0.017, 0.025] \\
NBFNet-style$^\dagger$ & 0.624 & [0.605, 0.641] & 0.506 & 0.848 & 0.676 & 0.692 & 0.549 & $-$0.193 [$-$0.213, $-$0.172] \\
OSFA-insp.$^\dagger$ & 0.605 & [0.587, 0.624] & 0.694 & 0.874 & 0.612 & 0.648 & 0.713 & $-$0.212 [$-$0.235, $-$0.188] \\
RDL-style (entity) & 0.821 & [0.797, 0.843] & 0.858 & 0.856 & 0.793 & 0.815 & 0.857 & +0.004 [$-$0.013, 0.023] \\
RDL-style (link)$^\dagger$ & 0.718 & [0.699, 0.737] & 0.860 & 0.855 & 0.760 & 0.787 & 0.826 & $-$0.099 [$-$0.114, $-$0.080] \\
Rules v2 + R-GCN (entity)$^\dagger$ (max) & 0.841 & [0.815, 0.864] & 0.879 & 0.884 & 0.816 & 0.832 & 0.849 & +0.024 [0.008, 0.039] \\
Rules v2 + R-GCN (entity)$^\dagger$ (lr) & 0.847 & [0.823, 0.868] & 0.908 & 0.894 & 0.828 & 0.828 & 0.838 & +0.030 [0.019, 0.041] \\
Rules v2 + RDL-style (entity) (max) & 0.835 & [0.811, 0.856] & 0.866 & 0.868 & 0.816 & 0.827 & 0.846 & +0.018 [0.005, 0.033] \\
Rules v2 + RDL-style (entity) (lr) & 0.844 & [0.822, 0.866] & 0.900 & 0.879 & 0.824 & 0.824 & 0.844 & +0.027 [0.019, 0.037] \\
GraIL-style $k{=}2$$^\dagger$ & 0.559 & [0.548, 0.569] & 0.644 & 0.749 & 0.592 & 0.609 & 0.752 & $-$0.258 [$-$0.278, $-$0.237] \\
GraIL-style $k{=}3$ & 0.631 & [0.607, 0.653] & 0.623 & 0.784 & 0.641 & 0.690 & 0.803 & $-$0.186 [$-$0.208, $-$0.162] \\
\bottomrule
\end{tabular}
\end{adjustbox}
\end{table}

% generated by xiaobo-experiment/render-results-draft.py; do not edit
\begin{table}[htbp]\centering\footnotesize\setlength{\tabcolsep}{3.5pt}
\caption{Link AUPRC per planted pattern and hop (cell positives vs all negatives; exploratory, not pre-specified as a test): Rules v2 leads most 2-hop cells and scores zero in every 3- and 4-hop cell except the ghost-employee cell, whose evidence includes the manager edge, while the GraIL-style $k{=}3$, NBFNet-style and opening-subgraph models score above zero in every cell. $^\dagger$GNN that is not a pre-specified primary model (descriptive).}
\label{tab:link-patternhop}
\begin{adjustbox}{max width=\linewidth}
\begin{tabular}{lrrrrrrrrr}
\toprule
pattern & hop & $n_+$ & Rules v2 & \makecell[l]{Path\\ {}features + LR} & \makecell[l]{GraIL-style\\ {}$k{=}2^\dagger$} & \makecell[l]{GraIL-style\\ {}$k{=}3$} & \makecell[l]{NBFNet-\\ {}style$^\dagger$} & OSFA-insp.$^\dagger$ & \makecell[l]{Rules v2 +\\ {}GraIL-style\\ {}$k{=}3$ (lr)} \\
\midrule
account change before run & 2 & 48 & 0.199 & 0.027 & 0.222 & 0.136 & 0.184 & 0.102 & 0.214 \\
account change before run & 3 & 48 & 0.000 & 0.050 & 0.086 & 0.179 & 0.090 & 0.113 & 0.127 \\
approver link & 2 & 48 & 0.818 & 0.201 & 0.242 & 0.200 & 0.330 & 0.187 & 0.854 \\
approver link & 3 & 48 & 0.000 & 0.056 & 0.096 & 0.173 & 0.136 & 0.084 & 0.094 \\
employee--vendor account & 2 & 32 & 0.279 & 0.040 & 0.132 & 0.195 & 0.238 & 0.129 & 0.268 \\
employee--vendor account & 3 & 32 & 0.000 & 0.055 & 0.073 & 0.196 & 0.082 & 0.116 & 0.118 \\
employee--vendor account & 4 & 32 & 0.000 & 0.002 & 0.002 & 0.131 & 0.076 & 0.086 & 0.085 \\
employee--vendor contact & 2 & 32 & 0.265 & 0.017 & 0.054 & 0.262 & 0.128 & 0.071 & 0.287 \\
employee--vendor contact & 3 & 32 & 0.000 & 0.015 & 0.080 & 0.102 & 0.065 & 0.061 & 0.062 \\
employee--vendor contact & 4 & 32 & 0.000 & 0.000 & 0.001 & 0.146 & 0.057 & 0.068 & 0.075 \\
ghost employee & 1 & 48 & 0.324 & 0.051 & 0.036 & 0.140 & 0.386 & 0.045 & 0.201 \\
ghost employee & 3 & 48 & 0.399 & 0.159 & 0.296 & 0.186 & 0.595 & 0.136 & 0.352 \\
shared payroll account & 2 & 32 & 0.295 & 0.026 & 0.136 & 0.105 & 0.176 & 0.144 & 0.236 \\
shared payroll account & 3 & 32 & 0.000 & 0.042 & 0.092 & 0.195 & 0.081 & 0.217 & 0.093 \\
shared payroll account & 4 & 32 & 0.000 & 0.000 & 0.001 & 0.106 & 0.052 & 0.105 & 0.071 \\
\bottomrule
\end{tabular}
\end{adjustbox}
\end{table}

% generated by xiaobo-experiment/render-results-draft.py; do not edit
\begin{table}[htbp]\centering\footnotesize\setlength{\tabcolsep}{3.5pt}
\caption{Link AUPRC per audit test (exploratory): Rules v2 is strongest on ghost-employee and approver--payee tests, whose evidence is usually a direct edge, and the GraIL-style model on employee--vendor tests. The employee--vendor test merges the account and contact patterns; the sixth test, the SoD conflict, is entity-only and appears in Table~\ref{tab:entity-pattern}. $^\dagger$GNN that is not a pre-specified primary model (descriptive).}
\label{tab:link-control}
\begin{adjustbox}{max width=\linewidth}
\begin{tabular}{lrrrrrrr}
\toprule
audit test (named after its scheme) & $n_+$ & Rules tuned & Rules v2 & \makecell[l]{Path\\ {}features + LR} & \makecell[l]{GraIL-style\\ {}$k{=}3$} & \makecell[l]{NBFNet-\\ {}style$^\dagger$} & \makecell[l]{Rules v2 +\\ {}GraIL-style\\ {}$k{=}3$ (lr)} \\
\midrule
Approver--payee link & 96 & 0.258 & 0.409 & 0.192 & 0.288 & 0.327 & 0.566 \\
Bank-account change before payroll run & 96 & 0.142 & 0.100 & 0.066 & 0.258 & 0.216 & 0.267 \\
Employee--vendor account/contact & 192 & 0.074 & 0.125 & 0.065 & 0.466 & 0.355 & 0.425 \\
Ghost employee & 96 & 0.126 & 0.474 & 0.164 & 0.231 & 0.560 & 0.409 \\
Shared payroll account & 96 & 0.041 & 0.099 & 0.040 & 0.277 & 0.210 & 0.266 \\
\bottomrule
\end{tabular}
\end{adjustbox}
\end{table}

% generated by xiaobo-experiment/render-results-draft.py; do not edit
\begin{table}[htbp]\centering\footnotesize\setlength{\tabcolsep}{3.5pt}
\caption{Entity AUPRC per planted pattern (exploratory): entity GNNs gain most over Rules v2 on the employee--vendor patterns and the SoD conflict (a control deficiency, read from role attributes), the rules and their combination lead on approver links, and the link GNN aggregated by the maximum misses the SoD conflict. $n_+$ counts all hops of a pattern. $^\dagger$GNN that is not a pre-specified primary model (descriptive).}
\label{tab:entity-pattern}
\begin{adjustbox}{max width=\linewidth}
\begin{tabular}{lrrrrrrr}
\toprule
pattern & $n_+$ & Rules v2 & \makecell[l]{Path\\ {}features + LR} & \makecell[l]{R-GCN\\ {}(entity)$^\dagger$} & \makecell[l]{RDL-style\\ {}(entity)} & \makecell[l]{GraIL-style\\ {}$k{=}3$} & \makecell[l]{Rules v2 +\\ {}RDL-style\\ {}(entity) (lr)} \\
\midrule
account change before run & 96 & 0.306 & 0.284 & 0.352 & 0.350 & 0.258 & 0.362 \\
approver link & 96 & 0.427 & 0.172 & 0.375 & 0.302 & 0.289 & 0.476 \\
employee--vendor account & 96 & 0.373 & 0.523 & 0.508 & 0.511 & 0.349 & 0.488 \\
employee--vendor contact & 96 & 0.130 & 0.227 & 0.300 & 0.238 & 0.306 & 0.197 \\
ghost employee & 96 & 0.467 & 0.448 & 0.508 & 0.437 & 0.235 & 0.566 \\
shared payroll account & 96 & 0.366 & 0.262 & 0.312 & 0.310 & 0.277 & 0.386 \\
SoD conflict & 96 & 0.327 & 0.412 & 0.502 & 0.411 & 0.028 & 0.411 \\
\bottomrule
\end{tabular}
\end{adjustbox}
\end{table}

\FloatBarrier
\section{Reproducibility and Provenance}
\label{app:repro}

\paragraph{Number provenance.} Result macros and numerical table bodies are retained from the original and extension report exporters. Source bindings record the report hashes. This offline manuscript revision changes prose, ordering and captions, not recorded scores, table values, model fits or test decisions. The original exporter reads study reports, training logs, freeze and design records, development outputs, audits and score-only analyses; the extension uses a separate report exporter. Design constants and the idealized twin-reference value of 0.5 are identified as such.

\paragraph{Freeze.} The prospective protocol, all code, the data manifest and the design document were frozen as \GhFreezeFiles{} files with SHA-256 checksums (checksum-file digest beginning \texttt{\GhFreezeShaShort}) on \GhFreezeDate; the confirmatory run script refused to start unless every checksum and every data file matched, and wrote to a directory that it refused to overwrite.

\paragraph{Environment.} Python \GhPyVersion{} with PyTorch \GhTorchVersion{} and PyTorch Geometric on CPU (\GhThreads{} threads per process, low scheduling priority) for the GNNs; Node.js for the generator, rules, heuristics and path features.

\paragraph{Addendum v2.} Every addendum number is generated from the addendum report by a separate script into its own macro file; the addendum freeze (text, \GvV{run.freezev2} files, checksum-file digest beginning \texttt{\GvV{freeze.sha}}) was verified by the run script before any new score was computed, together with the original \GvV{run.freezeorig} frozen files. Org-cluster bootstrap with \GvV{boot} replicates (rng seed 913) as in the original run; CPU only, 4 threads, low priority.

\paragraph{Artifact availability.} Code and data (generators, synthetic datasets and run records) are available from the authors on request.

\end{document}